\documentclass{easychair}

\usepackage{doc}

\usepackage{pifont}
\usepackage{xcolor}
\definecolor{second}{HTML}{007000}

\usepackage{comment}

\usepackage{amssymb}
\usepackage{pifont}
\usepackage{booktabs}
\usepackage{longtable}
\usepackage{pgfplotstable}
\usepackage{titlesec}
\usepackage{multirow}
\usepackage{subcaption}
\usepackage{fancyvrb}
\usepackage{cleveref}
\usepackage{threeparttable}
\usepackage{wrapfig}
\usepackage{makecell}
\usepackage{hyperref}
\usepackage{tablefootnote}
\usepackage{float}
\usepackage{pdflscape}

\pgfplotsset{compat=1.16}

\newcommand{\xmark}{\ding{55}}

\pgfkeysifdefined{/pgfplots/table/output empty row/.@cmd}{
    \pgfplotstableset{
        empty header/.style={
          every head row/.style={output empty row},
        }
    }
}{
    \pgfplotstableset{
        empty header/.style={
            typeset cell/.append code={%
                \ifnum\pgfplotstablerow=-1 %
                    \pgfkeyssetvalue{/pgfplots/table/@cell content}{}%
                \fi
            }
        }
    }
}

\titlespacing*{\paragraph}
{0pt}{2pt}{5pt}

\title{The 7th International Verification of Neural Networks Competition (VNN-COMP 2026): Summary and Results}%

\author{ Tobias Ladner$^*$\inst{1}
        \and Konstantin Kaulen$^*$\inst{2}
        \and Stanley Bak\inst{3}
        \and Julien Girard-Satabin\inst{4}
        \and Taylor T. Johnson\inst{5}
        \and Edoardo Manino\inst{6}
        \and ThanhVu H Nguyen\inst{7}
        \and Ben Wooding\inst{5}
        \and Haoze Wu\inst{8, 9}
}

\institute{
        Technical University of Munich, Munich, Germany\\
        \and 
        RWTH Aachen University, Aachen, Germany\\
        \and
        Stony Brook University, Stony Brook, New York, USA\\
        \and Université Paris-Saclay, CEA, List, F-91120, Palaiseau, France\\
        \and
        Vanderbilt University, Nashville, Tennessee, USA\\
        \and
        University of Manchester, Manchester, UK\\
        \and
        George Mason University, Fairfax, Virginia, USA\\
        \and
        Amherst College, Amherst, Massachusetts, USA\\
        \and VMware Research by Broadcom, USA\\
}

\authorrunning{Ladner et al.}
\titlerunning{VNN-COMP 2026 Report}

\date{\phantom{date}}

\begin{document}

\maketitle

\begin{abstract}
This report summarizes the seventh International Verification of Neural Networks Competition (VNN-COMP 2026), held as part of the Symposium on AI Verification (SAIV) co-located with the International Conference on Computer-Aided Verification (CAV) at FLoC 2026 in Lisbon, Portugal.
%
%
The goal of the competition is to provide an objective comparison of the state-of-the-art methods in neural network verification, in terms of scalability and speed.
Along this line, we used standard formats (ONNX for neural networks and VNNLIB for specifications), standard hardware (all tools are run by the organizers on AWS), and tool parameters provided by the tool authors.
This report summarizes the rules, benchmarks, participating tools, results, and lessons learned from this competition.

\end{abstract}

\renewcommand{\thefootnote}{*}
\footnotetext{Equal contribution.}
\renewcommand{\thefootnote}{\arabic{footnote}}



\section{Introduction}
\label{sec:introduction}

Deep-learning-based systems 
are increasingly being deployed in a wide range of domains, including recommendation systems, computer vision, and autonomous driving. While the nominal performance of these methods has increased significantly over the last years, they largely lack formal guarantees on their behavior. 
However, in safety-critical applications, including autonomous systems, robotics, cybersecurity, and cyber-physical systems (CPS), such guarantees are essential for certification and reliability.

While the literature on the verification of traditionally designed systems is wide and successful, neural network verification remains an open problem, despite significant efforts over the last years. In 2020, the International Verification of Neural Networks Competition (VNN-COMP) was established to facilitate comparison between existing approaches, bring researchers working on this problem together, and help shape future directions of the field. VNN-COMP has been held annually since then~\cite{bak2021vnncomp,muller2022vnncomp,brix2023years,brix2023vnncomp,brix2024vnncomp,kaulen20256th}. In 2026, the 7th iteration of the annual VNN-COMP\footnote{\url{http://vnn-comp.github.io/}} was held as a part of the third International Symposium on AI Verification (SAIV)~\cite{saiv2026} that was collocated with the 38th International Conference on Computer-Aided Verification (CAV)~\cite{cav2026}.

This 7th iteration of the VNN-COMP continues last year's trend of increasing standardization and automatization, aiming to enable a fair comparison between the participating tools and to simplify the evaluation of a large number of tools on a variety of (real-world) problems. 
As in the last iteration, VNN-COMP 2026 standardizes (1) neural network and specification formats, ONNX for neural networks and VNN-LIB \cite{vnnlib} for specifications, (2) evaluation hardware, providing participants the choice of a range of cost-equivalent AWS instances with different trade-offs between CPU and GPU performance, and (3) evaluation pipelines, enforcing a uniform interface for the installation and evaluation of tools.


The competition was kicked off with the solicitation for participation in February 2026. Rule discussion started in March and rules were finalized in May 2026 (see an overview in \Cref{sec:rules}). From April to June 2026, benchmarks were proposed and discussed. Meanwhile, the organizing team decided to continue using AWS as the evaluation platform and started to implement an automated submission and testing system for both benchmarks and tools. By mid-July 2026, eleven teams submitted their tools and the organizers evaluated all entrants to obtain the final results, discussed in \Cref{sec:results} and presented at SAIV on July 24, 2026. 
Discussions were structured into three issues on the official GitHub repository\footnote{\url{https://github.com/VNN-COMP/vnncomp2026/issues}}: rules discussion, benchmarks discussion, and tool submission. All submitted benchmarks\footnote{\url{https://github.com/VNN-COMP/vnncomp2026_benchmarks}} and final results\footnote{\url{https://github.com/VNN-COMP/vnncomp2026_results}} were aggregated in separate GitHub repositories. 

The remainder of this report is organized as follows: \Cref{sec:rules} discusses the competition rules, \Cref{sec:participants} lists all participating tools, \Cref{sec:benchmarks} lists all benchmarks,  \Cref{sec:results} summarizes the results, and \Cref{sec:conclusion} concludes the report, discussing potential future improvements.

\newpage

\section{Rules}
\label{sec:rules}

\paragraph*{Terminology}
An \emph{instance} is defined by a property specification (pre- and post-condition), a network, and a timeout. 
For example, one instance might consist of an MNIST classifier with one input image, a given local robustness threshold $\epsilon$, and a specific timeout.
A \emph{benchmark} is defined as a set of related instances.
For example, one benchmark might consist of a specific MNIST classifier with 100 input images, potentially different robustness thresholds $\epsilon$, and one timeout per input.

\paragraph*{Run-time caps}
Run-times are capped on a per-instance basis, i.e., any verification instance will timeout (and be terminated) after at most X seconds, determined by the benchmark proposer. These can be different for each instance. 
The total per-benchmark runtime (sum of all per-instance timeouts) may not exceed 6 hours per benchmark. 
For example, a benchmark proposal could have six instances with a one-hour timeout, or 100 instances with a 3.6-minute timeout, each.
To enable a fair comparison, we measure the startup overhead for each tool by running it on a range of tiny networks and subtract the minimal overhead from the total runtime.

\paragraph*{Hardware}
To allow for comparability of results, all tools were evaluated on equal-cost hardware using  Amazon Web Services (AWS).
Each team could decide between a range of AWS instance types (see \Cref{tab:instances}) providing a CPU, GPU, or mixed focus.


\begin{table}[h]
\centering
\caption{Available AWS instances.}\label{tab:instances}
\renewcommand{\arraystretch}{1.1}
\scalebox{0.985}{
\begin{tabular}{lccc} \toprule
         &vCPUs & RAM [GB] & GPU \\ 
         \midrule
         p3.2xlarge  & 8 & 61 & V100 GPU with 16 GB memory \\
         m5.16xlarge  & 64 & 256 & \xmark \\
         g5.8xlarge  & 32 & 128 & A10G GPU with 24 GB memory \\
         \bottomrule
\end{tabular}
}
\end{table}

\paragraph{Scoring} 
The final score is aggregate as the sum of all benchmark scores.
Each benchmark score is the number of points (sum of instance scores discussed below) achieved by a given tool, normalized by the maximum number of points achieved by any tool on that benchmark. 
Thus, the tool with the highest sum of instance scores for a benchmark will get a benchmark score of 100, ensuring that all benchmarks are weighted equally, regardless of the number of constituting instances.

\paragraph*{Instance score}
\label{sec:scoring}
Each instance is scored is as follows: 
\begin{itemize}\setlength{\itemsep}{0pt}
    \item Correct hold (property proven): 10 points;
    \item Correct violated (counterexample found): 10 points;
    \item Incorrect result: -150 points (penalty increased compared to 2022);
    \item Timeout / Runtime Error / Unknown: 0 points.

\end{itemize}
However, the ground truth for any given instance is generally not known a priori. In the case of disagreement between tools, we, therefore, place the burden of proof on the tool claiming that a specification is violated, i.e., that a counterexample can be found, and deem it correct exactly if it produces a valid counterexample.

%
The provided counterexamples were supposed to define both the input and the resulting output of the networks.
However, for some tools and instances, the output definition was either missing or differed from the network output as computed by the onnxruntime package used to evaluate counterexamples (by performing inference given the inputs).
The competition rules were ambiguous how this would be handled. We decided to discard all outputs in the counterexample files and base the evaluation solely on the given inputs and their respective outputs as computed by the onnxruntime.

\paragraph*{Time bonus}
As in the previous year, no time bonus was awarded.
Instead, all tools that compute the correct result within the time limit receive the same amount of points.

\paragraph{Overhead Correction}
The overhead of tools was measured, but only used to adapt the timeouts. It did not influence the scores, as no time bonus was awarded.
To measure the tool-specific overhead, we created trivial network instances and included those in the measurements. We then observed the minimum verification time over all instances and considered that to be the overhead time for the tool.

\paragraph{Format}
As since 2021, we standardized neural networks to be in \texttt{onnx} format, specifications in \texttt{vnnlib} format, and counterexamples in a format similar to the \texttt{vnnlib} format.
Further, tool authors were required to provide scripts fully automating the installation process of their tool, including the acquisition of any licenses that might be needed. Similar to the previous year, a preparation and execution script had to be provided for running their tool on a specific instance consisting of a network file, specification file, and timeout.
The specifications are interpreted as definitions of counterexamples, meaning that a property is proven ``correct'' if the specification is shown to be unsatisfiable, conversely, the property is shown to be violated if a counterexample fulfilling the specification is found. 
Specifications consisted of disjunctions over conjunctions in both pre- and post-conditions, allowing a wide range of properties from adversarial robustness over multiple hyper-boxes to safety constraints to be encoded.
For example, robustness with respect to inputs in a hyper-box had to be encoded as disjunctive property, where any of the other classes is predicted.

\paragraph{Tracks}
Since VNN-COMP'24, the competition is split into two tracks, both of them scored: A "regular" track, and an "extended" track.
The regular track consists of benchmarks selected by the tool participants based on a voting process, where a benchmark is included in the regular track if at least 50\% of tool participants voted for it to be scored. Benchmarks with at least 1 vote and not included in the regular track were scored in the extended track.
All benchmarks received at least one vote, so no benchmark was unscored.

\newpage{}

\section{Changes compared to previous edition}

This section list the main differences in the competition organization compared to the previous year.  

\subsection{Support for a new benchmark format}

Since the second edition of VNN-COMP, the VNN-LIB format~\cite{vnnlib} was chosen as a standard input format for the competition. VNN-LIB 1.0 was a subset of the SMT-LIB~\cite{BarFT-RR-25} language, widely used in Satisfaction Modulo Theory (SMT) solvers. Its simplicity allowed easy parsing by each prover.

However, VNN-LIB 1.0 lacked a well-defined syntax and a standardised parser, which resulted in each too having to re-implement their own parser and subtle differences in the syntax accepted by each tool.
Furthermore, VNN-LIB 1.0 was insufficiently expressive~\cite{cordeiro2025neural}, lacking for instance the ability to specify neural network types, properties involving linear arithmetic constraints, properties involving multiple applications of the same network or multiple applications of a single network. 

In order to resolve these problems, VNN-LIB 2.0 was developed and released this year~\cite{roy2026vnnlib}. Compared to its predecessor, VNN-LIB 2.0 has a formal grammar, rigorous typing, handles nonlinear operations  and multiple network declarations. Version 2.0 of the standard was accompanied by parsers for various programming languages, including C++, Python, Julia and Ocaml. 
Benchmark proposers were keen to adopt this new version, with all benchmarks in the regular track providing a VNN-LIB 2.0 version of their specification. Some benchmarks in the extended version provided VNN-LIB 2.0-only versions. Tools developers were able to handle most of VNN-LIB 2.0 benchmarks, and fell back to the 1.0 version whenever unknown constructs were met. 

The aim next year is to move fully to VNN-LIB 2.0 as well as simplifying the competition infrastructure by using the standardised command-line format provided by the standard.

\subsection{Hardened counter-example validation}
For VNN-COMP 2026, the validation of counterexamples returned by participating tools was revised to address ambiguities arising from numerical tolerances. Counterexample validation is inherently challenging because verification specifications describe precise constraints, while both verification tools and the reference execution of ONNX networks rely on finite-precision floating-point arithmetic. Previous editions accommodated such numerical discrepancies by allowing violations of the specification constraints within a fixed tolerance. However, applying a tolerance directly to the feasible region changes the verification problem itself: an instance that is \texttt{unsat} under the original specification may become \texttt{sat} after its constraints are relaxed. Consequently, two contradictory answers could in principle both be considered correct, with one verifier proving the original property and another returning a counterexample that only satisfies the tolerance-relaxed specification. 

For VNN-COMP 2026, submitted counterexamples were therefore validated against the original input specification without tolerance. The submitted input was evaluated in a fixed reference environment using ONNX Runtime~1.26.0 and ONNX~1.21.0 under Python~3.12 on CPU of the tool author's chosen evaluation hardware, and the resulting outputs were checked against the original VNN-LIB output constraints.
A counterexample was classified as \texttt{CORRECT} only if the input and the reference network output satisfy the specification without tolerance, and only such counterexamples received the regular score for a correct \texttt{sat} result. At the same time, requiring exact agreement on the output constraints would introduce a different problem: a mathematically valid counterexample close to the decision boundary may be evaluated slightly differently by a verification tool and by the reference ONNX Runtime execution due to floating-point arithmetic, different implementations of network operations, or different evaluation orders. Counterexamples whose inputs satisfy the specification exactly and whose reference outputs satisfy the output constraints only up to an absolute tolerance of $10^{-4}$ are therefore classified separately as \texttt{CORRECT\_WITH\_TOLERANCE}. These results received full points (\emph{i.e.}, were not penalised as \texttt{INCORRECT}), but did not establish a correct \texttt{sat} result that could be used to penalise \texttt{unsat} results given by other tools.

This rule does not completely eliminate the fundamental numerical challenges of counterexample validation. In particular, the output of the fixed reference implementation is itself a finite-precision computation and is not necessarily identical to the real-arithmetic semantics implicitly assumed by many verification algorithms. Counterexamples located extremely close to a property boundary may consequently still change classification depending on numerical details of network execution. Fixing the ONNX Runtime version, execution platform, and numerical tolerance makes this behavior reproducible for the competition, but cannot provide an implementation-independent notion of validity for such borderline cases. The revised rule therefore represents a compromise: numerical tolerances are prevented from enlarging the input domain or turning an originally \texttt{unsat} problem into a valid \texttt{sat} result, while small numerical discrepancies in network evaluation are not penalised.

\subsection{Open-sourcing the evaluation platform}
We published the evaluation platform: \url{https://github.com/VNN-COMP/vnn-eval-platform}.

With more tool authors participating in ever more benchmarks,
this comes with a high financial cost to run all submissions on AWS instances.
Open-sourcing the evaluation platform enables anyone wanting to participate to set up their own instance of the VNN-COMP evaluation platform on their internal lab server.
Any submission to such a local instance then mimics an execution on an AWS instance by running the same benchmark orchestration within a Docker container, thus enabling thorough testing of any tool ahead of the actual submission without stressing our budget.
In the future, this also makes it easier for us to move off running VNN-COMP on AWS.

We are aware that this assumes that anyone who wants to participate has access to some server they can utilize for this; let us know if this is not the case, and, more broadly, do not hesitate to open an issue on the GitHub repository in case you encounter any bugs, have feature requests, or want to share other thoughts!

\newpage

\section{Participants}
\label{sec:participants}
We list the tools and teams that participated in the VNN-COMP 2026
in \Cref{tab:tools}
and reproduce their own descriptions of their tools below.



\begin{table*}[h]
\begin{center}
\begin{minipage}{\linewidth}
\caption{Summary of the key features of participating tools. The hardware column describes the used AWS instance with \texttt{p3} and \texttt{g5} making GPUs available, see \Cref{tab:instances} for more details. Licenses refer to the external licenses required to use the corresponding tool, not the licensing of the tool itself.} 
\label{tab:tools}%
\renewcommand{\arraystretch}{1.25}
\resizebox{\textwidth}{!}{ 
\begin{tabular}{lcm{5.0cm}ccc} \toprule
         Tool & References & Organizations & Place  & Hardware & Licenses\\
         \midrule
         $\alpha$,$\beta$-CROWN & \cite{xu2020automatic,xu2021fast,wang2021betacrown,zhang2022general,shi2024genbab} & UIUC, UCLA, Drexel, Duke, RWTH Aachen & \textbf{1} & g5 & GUROBI\\
%

          ACT & \cite{zhang2025oliva,fukuda2025adaptive,ren2026mining,liu2026detecting,zhang2026tensor} & University of New South Wales & 10 & g5 & GUROBI (optional) \\
         Agate & - & None & 13 & g5 & - \\
         Bright & \cite{DeepPoly:19,xu2021fast, wang2021betacrown} & University of Kent & 7 & g5 & - \\
         CORA &  \cite{althoff_2015,koller_et_al_2025settraining,koller_et_al_2025shadows,kochdumper_et_al_2023,ladner_althoff_2023} & Technical University of Munich & 6 & g5 & MATLAB \\
         
%

%
         NeuralSAT & \cite{duong2023dpll,duong2024harnessing,duong2025neuralsat} & George Mason University & 4 & g5 & GUROBI\\
         nnenum & \cite{bak2020cav,bak2021nnenum} & Stony Brook University & 11 & m5 & - \\
         NNV & \cite{tran2020cav_tool,manzanas2023cav,tumlin2026nnv3} & Vanderbilt University & 5 & g5 & MATLAB\\
         n2v & \cite{sasaki2026n2v} & Vanderbilt University & 8 & g5 & -\\
         PyRAT &  \cite{pyrat2024, sas_pyrat_2025}& Université Paris-Saclay, CEA-List & 3 & g5& closed source CEA license \\

         SobolBox &  \cite{sobolbox} & IIT Gandhinagar & 12 & m5 & - \\
         SuperSimple & - & RWTH Aachen & 9 & m5 & - \\
         Vibecheck & \cite{bak2026vibecheck,bak2026vibechecknn} & Stony Brook University & 2 & g5 & - \\

         \bottomrule
\end{tabular}
}
\end{minipage}
\end{center}
\end{table*}



\subsection{$\alpha,\!\beta$-CROWN}

\paragraph*{Team} Co-leaders: Xiangru Zhong (student lead) and Huan Zhang (faculty advisor); 
Team members:
Duo Zhou,
Lei Huang,
Yifei Li,
Jingyi Wang,
Xuanle Li,
Ashmit Dutta,
Lingyin Zhang.
All members are from the University of Illinois Urbana-Champaign (UIUC).

The team \textbf{acknowledges} Hesun Chen, Haoyu Li, Ruize Gao, Hao Cheng, and Taobo Liao, who were involved in the development of the verifier in the year of 2026 but did not directly work on any benchmarks of the competition. A full list of contributors is available at \url{https://abcrown.org}.

\paragraph*{Description} 
$\alpha,\!\beta$-CROWN (\texttt{alpha-beta-CROWN}) is an efficient neural network verifier based on the linear bound propagation framework and built on a series of works on bound-propagation-based neural network verifiers:  CROWN~\cite{zhang2018efficient}, auto\_LiRPA~\cite{xu2020automatic}, $\alpha$-CROWN~\cite{xu2021fast}, $\beta$-CROWN~\cite{wang2021betacrown}, GCP-CROWN~\cite{zhang2022general}, GenBaB~\cite{shi2024genbab}, BICCOS~\cite{zhou2024scalable}.
The core techniques in $\alpha,\!\beta$-CROWN combine the efficient and GPU-accelerated linear bound propagation method with branch-and-bound methods specialized for neural network verification. 

The linear bound propagation algorithms in $\alpha,\!\beta$-CROWN are based on our \texttt{auto\_LiRPA} library~\cite{xu2020automatic}, which supports general neural network architectures (including convolutional layers, pooling layers, residual connections, recurrent neural networks, Transformers, and general computation graphs) and a wide range of nonlinear functions (e.g., ReLU, tanh, trigonometric functions, sigmoid, max pooling and average pooling), and is efficiently implemented on GPUs with Pytorch and CUDA. We jointly optimize intermediate layer bounds and final layer bounds using gradient ascent (referred to as $\alpha$-CROWN or optimized CROWN/LiRPA~\cite{xu2021fast}). Most importantly, we use branch and bound~\cite{bunelunified2018} (BaB) and incorporate split constraints in BaB into the bound propagation procedure efficiently via the $\beta$-CROWN algorithm~\cite{wang2021betacrown}, use cutting-plane method in GCP-CROWN~\cite{zhang2022general} and BICCOS~\cite{zhou2024scalable} to further tighten the bound, and support general nonlinearities in the branch-and-bound by GenBaB~\cite{shi2024genbab}.
For smaller networks, we also use a mixed integer programming (MIP) formulation~\cite{Tjeng2019EvaluatingRO} combined with tight intermediate layer bounds from $\alpha$-CROWN (referred to as $\alpha$-CROWN + MIP~\cite{zhang2022general}). The combination of efficient, optimizable and GPU-accelerated bound propagation with BaB produces a powerful and scalable neural network verifier.

\paragraph*{Link} \url{https://github.com/Verified-Intelligence/alpha-beta-CROWN} (main version)
\paragraph*{Competition submission} \url{https://github.com/Verified-Intelligence/alpha-beta-CROWN_vnncomp2026} (only for reproducing competition results; please use the main version for other purposes)
\paragraph*{Hardware and licenses} CPU and GPU with 32-bit or 64-bit floating point; Gurobi license required for certain benchmarks.
\paragraph*{Participated benchmarks} All benchmarks.

\subsection{ACT}
\paragraph{Team} Guanqin Zhang, Yulei Sui (University of New South Wales, Sydney).
\paragraph{Description}

ACT (Abstract Constraint Transformation) is an open-source framework for neural network testing and verification, built around a three-tier architecture comprising a front-end, back-end, and pipeline, with native PyTorch support.

The front-end supports model and specification synthesis and translation, converting diverse models and specifications into ACT’s intermediate representation (IR). The back-end performs abstract execution and tensor-based branch-and-bound refinement, with constraint generation decoupled from constraint solving. Its search design draws on several prior works, including eager falsification~\cite{guo2021eager}, stochastic optimization-based falsification~\cite{zhang2025oliva}, MCTS-based exploration~\cite{fukuda2025adaptive}, and verdict boundary mining~\cite{ren2026mining}, among others. ACT’s back-end also checks consistency between sampled concrete executions and the abstract bounds produced by abstract execution to detect unsoundness in verifier implementations~\cite{liu2026detecting}.

Finally, the ACT pipeline connects the front-end and back-end and provides tensor-based testing and verification capabilities, including fuzzing~\cite{zhang2026tensor}. This enables parallel mutation, execution, and property checking across different test inputs, while tensor-based refinement of verification subproblems exploits CPU/GPU parallelism to scale branch-and-bound-based verification.

\paragraph*{Link} \url{https://github.com/SVF-tools/ACT}
\paragraph*{Commit} 492d904d5bc2bb09438a538089f8381d0d75fe4e
\paragraph*{Hardware and licenses} CPU and GPU; Gurobi license (optional).
\paragraph*{Participated Benchmarks} 
\texttt{acasxu\_2023},
\texttt{cersyve},
\texttt{challenging\_certified\_training\_2026},
\texttt{cifar100\_2024},
\texttt{cora\_2024},
\texttt{dist\_shift\_2023},
\texttt{malbeware},
\texttt{relusplitter\_2026},
\texttt{safenlp\_2024},
\texttt{sat\_relu},
\texttt{tinyimagenet\_2024},
\texttt{tllverifybench\_2023},
\texttt{vggnet16\_2022},
\texttt{vit\_2023}.

\subsection{Agate}
\paragraph{Team} David Shriver (unaffiliated).
\paragraph{Description}

Agate is an early implementation of a neural network verification framework. It is still a work in progress. Agate takes in an ONNX model and behavior specification (such as a VNNLIB 2.0 counter example specification) and transforms them into a single model that expresses the satisfiability of the problem. This transformation is similar to the one used by DNNF \cite{shriver2021dnnf}, but combines all subproblems into a single model. The initial implementation submitted encompassed a simple adversarial attack solver and basic interval bound propagation with input branch and bound for complete verification. These approaches were run in parallel and both used the GPU, except for very small models. The only benchmark specific tuning done was to reduce the batch size for benchmarks with large models. The backend was implemented in Jax and supported most common ONNX operators, but notably, could not support operations with input dependent output shapes, such as dynamic slicing, at the time of submission. It uses a custom VNNLIB parser that did not support strict inequality at the time of submission. Finally, a last minute bug in the counter-example post-processor resulted in invalid witnesses, contributing to our poor competition performance. This bug has since been fixed, and our test suite is significantly improved to detect such regressions in the future.

\paragraph*{Link} \url{https://github.com/veridnn/agate-vnncomp2026}
\paragraph*{Commit} a1227f8c46e215e723de35a620e2105a8b4f83aa
\paragraph*{Hardware and licenses} CPU and GPU. Full source available on request.
\paragraph*{Participated Benchmarks} 
\texttt{acasxu\_2023},
\texttt{cersyve},
\texttt{cgan2026},
\texttt{challenging\_certified\_training\_2026},
\texttt{collins\_rul\_cnn\_2022},
\texttt{cora\_2024},
\texttt{dist\_shift\_2023},
\texttt{linearizenn\_2024},
\texttt{lsnc\_relu},
\texttt{malbeware},
\texttt{metaroom\_2023},
\texttt{ml4acopf\_2024},
\texttt{nn4sys},
\texttt{relusplitter\_2026},
\texttt{safenlp\_2024},
\texttt{sat\_relu},
\texttt{tllverifybench\_2023}.

\subsection{Bright}
\paragraph*{Team} Max Peeperkorn and Elena Botoeva (University of Kent)
\paragraph*{Description}

Bright is a composable branch-and-bound neural network verifier written in JAX.
It implements state-of-the-art GPU-accelerated bound propagation techniques \cite{DeepPoly:19,xu2021fast, wang2021betacrown} and is designed as a versatile research tool that emphasises customisation to easily integrate custom branching heuristics and procedures to offload domains to MIP under certain conditions.
The tool is still under development and currently supports fully-connected and convolutional networks with ReLU activations.

\paragraph*{Links} \url{https://git.cs.kent.ac.uk/bright/bright-vnncomp-2026}
\paragraph*{Commit} b2351384b23233b440ee35fd64dc6e3d54d08da6
\paragraph*{Hardware and licenses} CPU and GPU, Gurobi Licence
\paragraph*{Participated benchmarks} acasxu\_2023, cifar100\_2024, cora\_2024, linerizenn\_2024, malbeware, metaroom\_2023, relusplitter\_2026, safenlp\_2024, sat\_relu, tinyimagenet\_2024, tllverifybench\_2023

\subsection{CORA}
\paragraph*{Team} Lukas Koller, Tobias Ladner, Benedikt Kellner, Matthias Althoff \\
\hphantom{Team\ \,} (Technical University of Munich).
\paragraph*{Description} 
CORA~\cite{althoff_2015, Althoff2025manual} enables the formal verification of neural networks, both in open-loop as well as in closed-loop scenarios~\cite{kochdumper_et_al_2023,ladner_althoff_2023}. 
Open-loop verification refers to the task where properties of the output set of a neural network are verified, 
e.g., correctly classified images given noisy input, as also considered at VNN-COMP. 
In closed-loop scenarios, the neural network is used as a controller of a dynamic system, e.g., controlling a car while keeping a safe distance over some time horizon. Moreover, CORA can also train robust neural networks~\cite{koller_et_al_2025settraining,wendl2024training}, 
which uses an efficient batch-wise propagation of zonotopes through a neural network on a GPU. 

For the competition, reachability analysis with zonotopes is used to compute a convex enclosure of the output set of a neural network. In addition to simultaneous input and neuron splitting, a specification-driven input refinement procedure iteratively encloses the set of unsafe inputs~\cite{koller_et_al_2025shadows}. All operations use batch-wise computations to take full advantage of GPU-acceleration. 
Further details on our approach can be found in our competition contribution paper~\cite{koller2026cora_compcontribution}.

\paragraph*{Link} \url{https://github.com/toladnertum/vnncomp-cora-2026}

\paragraph*{Commit} 798551ab7e7cba1c2c8a3a14726478cbfc8690e6

\paragraph*{Hardware and licenses} GPU, MATLAB license.

\paragraph*{Participated Benchmarks} 
\texttt{acasxu}, \texttt{cersyve}, \texttt{cgan2026}, \texttt{challenging\_certified\_training\_2026}, \texttt{cifar100\_2024}, \texttt{collins\_rul\_cnn\_2022}, \texttt{cora\_2024}, \texttt{dist\_shift\_2023}, \texttt{linearizenn\_2024}, \texttt{malbeware}, \texttt{metaroom\_2023}, \texttt{nn4sys}, \texttt{safenlp\_2024}, \texttt{sat\_relu}, \texttt{soundnessbench\_2026}, \texttt{tinyimagenet\_2024}, \texttt{tllverifybench\_2023}, \texttt{isomorphic\_acasxu\_2026}, \texttt{monotonic\_acasxu\_2026}.

\subsection{NeuralSAT}
\paragraph*{Team} Hai Duong and Thanhvu Nguyen (George Mason).

\paragraph*{Description} 
NeuralSAT~\cite{duong2023dpll,duong2024harnessing,duong2025neuralsat,duong2025neuralsat2,duong2026compositional,duong2026generating,duong2026verifying,duong2026verifying2,duong2026verifying3,duong2026clad,duong2026zonogpt} integrates conflict-driven clause learning (CDCL) in SAT/SMT-solving with an DNN abstraction-based theory solver for infeasibility checking.
NeuralSAT implements the DPLL(T) framework used in modern SMT solvers such as  Z3. 
The design of NeuralSAT is inspired by the core algorithms used in SMT solvers such as CDCL components and theory solving. 
The tool is written in Python and uses Gurobi for LP solving. 
Unlike many other modern NNV tools,  NeuralSAT \emph{does not} require parameter tuning and works out of the box, e.g., the tool runs on the wide-range of benchmarks in VNN-COMPs without any tuning.


\paragraph*{Link}
\url{https://github.com/dynaroars/neuralsat}

\paragraph*{Commit}
8c6a493d2b9314b06c6f19cd452f1c7ab5bd2657

\paragraph*{Hardware and licences}
GPU, Gurobi License

\paragraph*{Participated benchmarks} All benchmarks in the regular track.

\subsection{nnenum}
\paragraph*{Team} Doug Wehbe, Eli Pregerson, Stanley Bak (Stony Brook University)

\paragraph*{Description} 
The nnenum tool~\cite{bak2021nnenum} uses multiple levels of abstraction to achieve high-performance verification of ReLU networks without sacrificing completeness~\cite{bak2020vnn}. 
The core verification method is based on reachability analysis using star sets~\cite{tran2019fm}, combined with the ImageStar method~\cite{tran2020cav} to propagate sets through all linear layers supported by the ONNX runtime, such as convolutional layers with arbitrary parameters.  
The tool is written in Python 3 and uses GLPK for LP solving.
This year we improved handling of sparse transformations like convolution layers to use less memory, as well as added some support for residual connections.

\paragraph*{Link}
\url{https://github.com/wehbedoug/nnenum-2026}

\paragraph*{Commit}
fa2a14d736973bfb728856afaffa35ed6cfae0ef

\paragraph*{Hardware and licences}
CPU, Gurobi license (optional)

\paragraph*{Participated benchmarks}
\texttt{acasxu}, \texttt{cgan}, \texttt{collins-rul-cnn}, \texttt{cora}, \texttt{linearizenn}, \texttt{metaroom}, \texttt{nn4sys}, \texttt{relusplitter}, \texttt{safeNLP}, \texttt{sat-relu},  \texttt{tllverifybench}, \texttt{vggnet16}.

\subsection{NNV}
\paragraph*{Team} Samuel Sasaki, Anne M. Tumlin, Ben Wooding, Taylor T. Johnson (Vanderbilt University)
\paragraph*{Description} The Neural Network Verification (NNV) Tool~\cite{tran2020cav_tool,manzanas2023cav,lopez2025saiv_cc,tumlin2026nnv3} is a formal verification software tool for deep learning models and cyber-physical systems with neural network components written in MATLAB and available at \url{https://github.com/verivital/nnv}. NNV uses a star-set state-space representation and reachability algorithm that allows for a layer-by-layer computation of exact or overapproximate reachable sets for feed-forward~\cite{tran2019fm}, convolutional~\cite{tran2020cav}, semantic segmentation (SSNN)~\cite{tran2021cav}, and recurrent (RNN)\cite{tran2023hscc} neural networks, as well as neural network control systems (NNCS)~\cite{tran2019emsoft,tran2020cav_tool} and neural ordinary differential equations (Neural ODEs)~\cite{manzanas2022formats}. NNV3, the latest version~\cite{tumlin2026nnv3}, considers weight perturbations~\cite{zubair2025verification,zubair2026modelstar}, graph neural networks (GNNs)~\cite{tumlin2026reachability}, videos~\cite{sasaki2025videoclass}, and fairness~\cite{tumlin2024fairnnv}.
The star-set based algorithm is naturally parallelizable, which allows NNV to be designed to perform efficiently on multi-core platforms. Additionally, if a particular safety property is violated, NNV can be used to construct and visualize the complete set of counterexample inputs for a neural network (exact-analysis). 
%
For this competition, updated from last year's, we tailor the solver approach depending on the benchmark at hand, although all follow a similar flow. First, we perform a simulation-guided search for counterexamples for a fixed number of samples. If no counterexamples are found (i.e., demonstrate that the property is SAT), then we utilize an iterative refinement approach using reachability analysis to verify the property (UNSAT). This consists of performing reachability analysis using a relax-approximation method~\cite{tran2021cav}, if not verified, then a less conservative approximation based on zonotope pre-filtering approach~\cite{tran2021fac}, and finally using the exact analysis when possible~\cite{tran2020cav} until the specification is verified or there is a timeout. Based on the benchmark to evaluate, the initial reachability analysis may be any of the overapproximation methods or the exact method, based on the complexity of the benchmarks (size of network, input, etc). 
In addition to the star set-based approaches, we implement another approach based on conformal prediction that can scale better to larger networks and input sizes, although only probabilistic guarantees can be provided~\cite{hashemi2025neurips}. This is the only method that uses GPU, all others run solely on CPU.

\paragraph*{Link} \url{https://github.com/verivital/nnv}

\paragraph*{Commit}
376568b449535a779f751dd2def774a0ed94ead8

\paragraph*{Hardware and licenses} CPU and GPU, MATLAB license.

\paragraph*{Participated Benchmarks} 
Regular track (all) and Extended track (vggnet16, ml4acopf, reluspliter).

\subsection{n2v}
\paragraph*{Team} Samuel Sasaki, Ben Wooding, Anne M. Tumlin, Taylor T. Johnson (Vanderbilt University)
\paragraph*{Description} 
\textsf{n2v}~\cite{sasaki2026n2v} is an open-source Python toolbox for neural network verification, entering the competition this year. \textsf{n2v} is a successor to the MATLAB-based NNV tool~\cite{tran2020cav_tool,manzanas2023cav,tumlin2026nnv3}, reimplemented in Python with native PyTorch integration to lower the barrier to adoption for the deep learning and verification communities. \textsf{n2v} supports exact and over-approximate reachability using star sets, zonotopes, and hyperrectangles. In addition, \textsf{n2v} contributes probabilistic verification via conformal inference, PGD-based falsification, and a batched HiGHS-backed linear-programming pipeline.

\paragraph*{Link} \url{https://github.com/sammsaski/n2v}

\paragraph*{Commit}
2b9ddcb39ca61780353e4416fd34d9d181538478

\paragraph*{Hardware and licenses} CPU and GPU, fully open-source.

\paragraph*{Participated Benchmarks} 
All benchmarks.

\subsection{PyRAT}

\paragraph*{Team} Augustin Lemesle, Sasha Cuau, Serge Durand, Tristan Le Gall (CEA-List), Julien Lehmann (OCamlPro)
\paragraph*{Description} 
PyRAT (Python Reachability Assessment Tool) \cite{pyrat2024,lemesle2025saiv_cc} is an abstract interpretation based tool to verify the safety and robustness of neural networks. PyRAT relies on multiple zonotopic domains to efficiently compute the reachable states for various architectures of neural networks (fully-connected, convolutionnal, recurrent, transformers, ...). It supports multiple non-linear activation functions (ReLU, Sigmoid, Softmax, Floor, ...) with precise abstractions while maintaining correctness and floating-point soundness on CPU. PyRAT can also leverage GPU-parallelization to scale to large networks or to batch analyses.
For smaller networks and problems, PyRAT can leverage branch and bound strategies on the inputs with heuristics like ReCIPH~\cite{durand2022reciph}. While on larger ReLU networks, PyRAT will use branch and bound strategies on the ReLU neurons in conjunction with fast GPU computation. 

For this year competition, PyRAT extended its property support to VNN-LIB 2.0 supporting its new paradigms like multiple inputs, networks and more complex properties. Relying on our relational zonotopic domains allowed PyRAT to address these new and complex relations between inputs and networks.

\paragraph*{Link} \href{https://git.frama-c.com/pub/pyrat/}{https://git.frama-c.com/pub/pyrat}
\paragraph*{Commit} 2dc848d64f987e283b6b85d739c6f5d6dfc6dd91
\paragraph*{Hardware and licenses} CPU and GPU, closed source CEA licence (free academic licence available for research purposes).
\paragraph*{Participated benchmarks} All benchmarks.

\subsection{SobolBox}
\paragraph*{Team}
Sarthak Das (Indian Institute of Technology Gandhinagar)

\paragraph*{Description}
SobolBox is a spiritual successor to the unsubmitted INNVerS \cite{innvers} project, developed by Shubhajit Roy (IIT Gandhinagar) and Avishek Lahiri (IACS Kolkata) for VNNCOMP-2021, under the supervision of Dr. Rajarshi Ray (IACS Kolkata).

SobolBox \cite{sobolbox,das2025saiv_cc} is a Python-based black-box falsification tool for detecting safety violations in neural networks. It accepts ONNX-format models and VNNLIB 1.0 safety specifications, treating neural networks as multi-input, multi-output, non-convex black boxes without making architectural assumptions. Input bounds are extracted from VNNLIB specifications using the Z3 theorem prover, after which low-discrepancy candidate points are generated using Sobol sequence sampling \cite{sobol}.

The core premise of SobolBox is that safety violations, if any, are most likely to occur near the optima. A preliminary pass over Sobol samples is used to quickly detect counterexamples before optimization, if they exist. For the remaining instances, promising regions are selected from the extrema of the sampled network outputs and refined using a combination of global PSO \cite{488968} and local L-BFGS-B optimization \cite{lbfgsb}. Candidate optima are checked for safety violations, with Sign-OPT searches \cite{cheng2020signoptqueryefficienthardlabeladversarial} additionally performed around the optima to detect nearby counterexamples.

The resulting output-bound under-approximations are checked against the safety property using Z3. SobolBox reports \texttt{sat} with a counterexample when a violation is found, \texttt{unsat} when the computed output bounds exclude a violation, or \texttt{unknown} when the analysis is inconclusive. Since the computed output bounds are under-approximations, \texttt{unsat} results are high-confidence but unsound. Networks whose effective input dimension exceeds the 21,201-dimensional limit of the employed Joe-Kuo Sobol construction \cite{doi:10.1137/070709359} are reported as \texttt{unknown}. SobolBox also employs memoization of black-box evaluations, parallelization, and caching of Sobol samples and computed output bounds to reduce computational overhead across runs.

For benchmarks with binary inputs and outputs, SobolBox provides an alternate discrete-search mode using rounded Sobol samples and Nevergrad's \texttt{PortfolioDiscreteOnePlusOne} optimizer; this mode does not report \texttt{unsat} since discrete search being stochastic can't mathematically prove the property holds everywhere, even within the underapproximated output bounds.

\paragraph*{Links}
\url{https://codeberg.org/tech-absurdist/SobolBox}

\paragraph{Commit}
64826496ec80bdc9f9925aefd9789d880e94272b

\paragraph*{Hardware and licenses}
CPU, no license required.

\paragraph*{Participated benchmarks}
\texttt{acasxu}, 
\texttt{cersyve}, 
\texttt{cgan2026},
\texttt{collins\_rul\_cnn\_2022},
\texttt{dist\_shift\_2023},
\texttt{linearizenn\_2024},
\texttt{metaroom\_2023},
\texttt{ml4acopf\_2024},
\texttt{safenlp\_2024},
\texttt{sat\_relu},
\texttt{tllverifybench\_2023}.

\subsection{SuperSimple}
\paragraph*{Team} Konstantin Kaulen (RWTH Aachen University)
\paragraph*{Description}
SuperSimple is an intentionally minimal and well-documented neural network verification tool developed as an educational and competition baseline for VNN-COMP.
It implements a deliberately simple verification pipeline: first, it searches for counterexamples using projected gradient descent (PGD); subsequently, it attempts to prove the property using incomplete bound propagation.
For the latter, SuperSimple builds on \texttt{auto\_LiRPA}~\cite{xu2020automatic} and successively applies CROWN~\cite{zhang2018efficient} and optimized CROWN ($\alpha$-CROWN)~\cite{xu2021fast}.
It returns \texttt{unsat} whenever the resulting bounds suffice to prove that the given property holds, and \texttt{unknown} otherwise.

This submission of a simple tool serves several purposes.
First, in light of the increasing computational demands of complete verification~\cite{KauEtAl25} and the complexity of configuring state-of-the-art verification systems~\cite{KauHoo26}, SuperSimple provides a lightweight and easy-to-use baseline for assessing how many VNN-COMP instances can already be solved using comparatively simple techniques and modest computational resources.
This provides an indication of the verification performance accessible to non-expert practitioners without highly specialized verification pipelines or large-scale compute environments.
Second, SuperSimple can reveal benchmarks for which such baseline techniques are already sufficient and may, in some cases, even outperform more sophisticated verification systems in terms of running time.
Such cases can highlight unnecessary algorithmic overhead, suboptimal configuration, or opportunities for improving the efficiency of state-of-the-art verifiers.
Finally, SuperSimple helps identify benchmarks that have become too easy to meaningfully discriminate between modern verification approaches: if almost all instances can be solved using this simple pipeline, the benchmark may be regarded as largely solved and of limited value for measuring further progress at the state of the art.

\paragraph*{Links}\url{https://github.com/kkaulen/simple_vnn_comp_verifier}
\paragraph*{Commit}780317d108e0de8aa9b29b4773079a9c4607213e
\paragraph*{Hardware and licenses}CPU or GPU, no license required.
\paragraph*{Participated benchmarks} 
All benchmarks (scored and unscored).

\subsection{vibecheck}
\paragraph*{Team} Doug Wehbe, Stanley Bak (Stony Brook University)
\paragraph*{Description}
The vibecheck formal verification tool is a high-performance, vibe-coded decision procedure for neural networks.
It was created using Claude Code, mostly Opus 4.8, over the course of two months.
It combines a variety of analysis techniques: bounds propagation, zonotope analysis, star set abstract domain, MILP solving, $\alpha$-CROWN-style slope optimization for ReLU relaxations, and a (maybe new) high performance GPU-based branch and bound solver to handle splitting.
A blog post is available~\cite{bak2026vibecheck} detailing the development process and experience.

\paragraph*{Links}\url{https://github.com/stanleybak/vibecheck-nn}
\paragraph*{Commit} 341c56bd5ca719d0935b9b6659c8610158974197
\paragraph*{Hardware and licenses} GPU, Gurobi
\paragraph*{Participated benchmarks} 
All benchmarks (scored and unscored).

\newpage
\section{Benchmarks}
\label{sec:benchmarks}

In this section, we provide an overview of all scored benchmarks, reproducing the benchmark proposers' descriptions.
Artifacts for all benchmarks are available in the repository\footnote{\url{https://github.com/VNN-COMP/vnncomp2026_benchmarks}}.
Table~\ref{tab:benchmarks} presents a summary of all benchmarks. 
%
Table~\ref{tab:benchmark_stats} presents statistics across all benchmarks.

      
{\small
\begin{longtable}{@{}llrrrr@{}}
\caption{Benchmark statistics per track. \textbf{VNNLIB} lists the specification-format versions each benchmark ships; \textbf{Budget} is the sum of the per-instance timeouts in hours.}\label{tab:benchmark_stats}\\
\toprule
\textbf{Benchmark} & \textbf{VNNLIB} & \textbf{Networks} & \textbf{Specs} & \textbf{Instances} & \textbf{Budget (h)} \\
\midrule
\endhead
\multicolumn{6}{@{}l}{\emph{Regular track}} \\
\midrule
acasxu\_2023 & 1.0, 2.0 & 45 & 10 & 186 & 6.0 \\
cersyve & 1.0, 2.0 & 12 & 3 & 12 & 0.3 \\
cgan2026 & 1.0, 2.0 & 8 & 21 & 21 & 6.0 \\
challenging\_certified\_training\_2026 & 1.0, 2.0 & 6 & 60 & 60 & 6.0 \\
cifar100\_2024 & 1.0, 2.0 & 2 & 200 & 200 & 5.6 \\
collins\_rul\_cnn\_2022 & 1.0, 2.0 & 3 & 41 & 62 & 6.0 \\
cora\_2024 & 1.0, 2.0 & 9 & 60 & 180 & 1.5 \\
dist\_shift\_2023 & 1.0, 2.0 & 1 & 72 & 72 & 6.0 \\
linearizenn\_2024 & 1.0, 2.0 & 10 & 60 & 60 & 15.0 \\
lsnc\_relu & 1.0, 2.0 & 1 & 80 & 80 & 0.6 \\
malbeware & 1.0, 2.0 & 3 & 50 & 150 & 4.2 \\
metaroom\_2023 & 1.0, 2.0 & 100 & 100 & 100 & 5.8 \\
ml4acopf\_2024 & 1.0, 2.0 & 9 & 23 & 69 & 11.5 \\
nn4sys & 1.0, 2.0 & 11 & 182 & 194 & 6.0 \\
relusplitter\_2026 & 1.0, 2.0 & 104 & 20 & 120 & 6.0 \\
safenlp\_2024 & 1.0, 2.0 & 1 & 1054 & 1080 & 6.0 \\
sat\_relu & 1.0, 2.0 & 100 & 100 & 100 & 2.8 \\
soundnessbench\_2026 & 1.0, 2.0 & 2 & 60 & 60 & 2.5 \\
tinyimagenet\_2024 & 1.0, 2.0 & 1 & 200 & 200 & 5.6 \\
tllverifybench\_2023 & 1.0, 2.0 & 32 & 32 & 32 & 5.3 \\
traffic\_signs\_recognition\_2023 & 1.0, 2.0 & 3 & 45 & 45 & 6.0 \\
vggnet16\_2022 & 1.0, 2.0 & 1 & 18 & 18 & 6.0 \\
vit\_2023 & 1.0, 2.0 & 2 & 200 & 200 & 5.6 \\
yolo\_2023 & 1.0, 2.0 & 1 & 72 & 72 & 6.0 \\
\midrule
\textbf{Regular total} & & \textbf{467} & \textbf{2763} & \textbf{3373} & \textbf{132.2} \\
\midrule
\midrule
\multicolumn{6}{@{}l}{\emph{Extended track}} \\
\midrule
adaptive\_cruise\_control\_non\_linear\_2026 & 2.0 & 1 & 50 & 50 & 1.4 \\
cctsdb\_yolo\_2023 & 1.0, 2.0 & 2 & 39 & 39 & 3.8 \\
collins\_aerospace\_benchmark & 1.0, 2.0 & 1 & 6 & 6 & 6.0 \\
isomorphic\_acasxu\_2026 & 2.0 & 50 & 50 & 50 & 1.4 \\
monotonic\_acasxu\_2026 & 2.0 & 31 & 50 & 50 & 1.4 \\
smart\_turn\_multimodal\_2026 & 2.0 & 1 & 50 & 50 & 1.4 \\
\midrule
\textbf{Extended total} & & \textbf{86} & \textbf{245} & \textbf{245} & \textbf{15.3} \\
\midrule
\midrule
\textbf{Overall} & & \textbf{553} & \textbf{3008} & \textbf{3618} & \textbf{147.5} \\
\bottomrule
\end{longtable}
}

\begin{table}[h]
    \centering
    \caption{Overview of all scored benchmarks.}
    \label{tab:benchmarks}
    \resizebox{\textwidth}{!}{
    \renewcommand{\arraystretch}{1.4}
    \begin{tabular}{ccccccc}
    \toprule
    Category &
    Benchmark &
    Application &
    Network Types &
    \# Params &
    Effective Input Dim &
    Track
    \\
    \midrule
    \multirow{8}{*}{Complex} 
    & cGAN & \makecell{Image Generation \\ \& Image Prediction} & Conv. + Vision Transformer & 500k - 68M & 5 & regular \\
    & NN4Sys & \makecell{Learned Index, Cardinality Estimation \\ \& Adaptive Bitrate Selection} & FC + Conv1D, ReLU + Sigmoid & 33k - 37M & 1-308 & regular \\
    & LinearizeNN & NN controller approximation & FC. + Conv. + Vision Transformer + Residual + ReLU & 203k & 4 & regular \\
    & ml4acopf & Power System & Complex (ReLU + Trigonometric + Sigmoid) & 4k-680k & 22 - 402 & extended \\
    & ViT & Vision & Conv. + Residual + Softmax + BatchNorm & 68k - 76k & 3072 & extended \\
    & Collins Aerospace & - & FC + Conv. + Residual, LeakyReLU + MaxPool + Square & 1.8M & 1.2M & extended \\
    & LSNC-ReLU & Lyapunov stability of NN controllers & FC, ReLU & 275 & 6 & extended \\
    & CCTSDB & - & FC + Conv. + Residual, ReLU +  MaxPool + Clip & 100k & 2 & extended \\
    \cmidrule(lr){1-7}
    \multirow{7}{*}{\makecell{CNN \\ \& ResNet}} 
    & Collins RUL CNN & Condition Based Maintenance & Conv. + ReLU, Dropout  & 60k - 262k   & 400 - 800 & regular \\
    & VGGNet16 & Image Classification & Conv. + ReLU + MaxPool    & 138M & 150k & extended \\
    & Traffic Signs Recognition & Image Classification & Conv. + Sign + MakPool + BatchNorm & 905k - 1.7M & 2.7k - 12k & extended \\
    & cifar100 & Image Classification & FC + Conv. + Residual, ReLU + BatchNorm & 2.5M - 3.8M & 3072 & regular \\
    & tinyimagenet & Image Classification & FC + Conv. + Residual, ReLU + BatchNorm & 3.6M & 9408 & regular \\
    & Metaroom & - & Conv. + FC, ReLU & 466k - 7.4M & 5376 & regular \\
    & Yolo & - & FC + Conv. + Residual, ReLU + Sigmoid & 22k - 37M & 1 - 308 & extended \\
    & SoundnessBench & Verifier Test & FC + Conv. + ReLU & 1.7M & 12k & regular \\
    & Relusplitter & Meta-Benchmark & FC + Conv. + ReLU & 13k - 625k & 5 - 3k & extended \\
    & MalBeWare & Malware Classification & FC + Conv. + ReLU & 102k - 1.5M & 4k & regular \\
    & cersyve & Neural Certificates for Control Systems & FC + Residual + ReLU & 1k - 4k & 2 - 6 & regular \\
    \cmidrule(lr){1-7} %
    \multirow{5.5}{*}{\makecell{FC}}
    & TLL Verify Bench & Two-Level Lattice NN & \makecell{Two-Level Lattice NN \\(FC. + ReLU)}  & 17k - 67M & 2 & regular \\
    & Acas XU & Collision Detection & FC. + ReLU & 13k & 5 & regular \\
    & Dist Shift & Distribution Shift Detection & FC. + ReLU + Sigmoid & 342k - 855k & 792 & regular \\
    & safeNLP & Sentence classification & FC. + ReLU & 4k & 30 & regular \\
    & CORA & Image Classification & FC. + ReLU & 575k, 1.1M & 784, 3072 & regular \\
    & SAT ReLU & Verifier Test & FC. + ReLU & 100-35k & 2 - 100 & regular \\
    \bottomrule
    \end{tabular}
    }
\end{table}

\subsection{cGAN}
\paragraph*{Proposed by} Feiyang Cai, Ali Arjomandbigdeli, Stanley Bak (Stony Brook University)
\paragraph*{Motivation}
While existing neural network verification benchmarks focus on discriminative models, the exploration of practical and widely used generative networks remains neglected in terms of robustness assessment.
This benchmark introduces a set of image generation networks specifically designed for verifying the robustness of the generative networks.
\paragraph*{Networks}
The generative networks are trained using conditional generative adversarial networks (cGAN), whose objective is to generate camera images that contain a vehicle obstacle located at a specific distance in front of the ego vehicle, where the distance is controlled by the input distance condition.
The network to be verified is the concatenation of a generator and a discriminator.  The generator takes two inputs: 1) a distance condition (1D scalar) and 2) a noise vector controlling the environment (4D vector). The output of the generator is the generated image. The discriminator takes the generated image as input and outputs two values: 1) a real/fake score (1D scalar) and 2) a predicted distance (1D scalar).
Several different models with varying architectures (CNN and vision transformer) and image sizes (32x32, 64x64) are provided for different difficulty levels.
\paragraph*{Specifications}
The verification task is to check whether the generated image aligns with the input distance condition, or in other words, verify whether the input distance condition matches the predicted distance of the generated image.
In each specification, the inputs (condition distance and latent variables) are constrained in small ranges, and the output is the predicted distance with the same center as the condition distance but with slightly larger range.
\paragraph*{Link} \url{https://github.com/feiyang-cai/cgan_benchmark2023}

\subsection{NN4Sys}
\paragraph*{Proposed by} the $\alpha,\!\beta$-CROWN team with collaborations with Shuyi Lin and Cheng Tan at Northeastern University.
\paragraph*{Application}
The benchmark contains networks for database learned index, video streaming learned adaptive bitrate, and learned cardinality
estimation which map inputs from various dimensions to 1-dimension outputs.

\begin{itemize}

\item \textit{Background}: learned index, learned cardinality, and learned
    adaptive bitrate are all instances in neural networks for computer systems
        (NN4Sys), which are neural network based methods performing system
        operations. These classes of methods show great potential but have one
        drawback---the outputs of an NN4Sys model (a neural network) can be
        arbitrary, which may lead to unexpected issues in systems.

\item \textit{What to verify}: our benchmark provides multiple pairs of (1) trained NN4Sys model
and (2) corresponding specifications. We design these pairs with different parameters such
that they cover a variety of user needs and have varied difficulties for verifiers. 
We describe benchmark details in our NN4SysBench paper~\cite{lin2024nn4sysbench}.

\item \textit{Translating NN4Sys applications to a VNN benchmark}: 
the original NN4Sys applications have some sophisticated structures that are hard to verify.
We tailored the neural networks and their specifications to be suitable for VNN-COMP.
For example, learned index~\cite{kraska18case} contains multiple NNs in a tree structure that together serve one purpose.
However, this cascading structure is inconvenient/unsupported to verify
because there is a ``switch" operation---choosing one NN in the second stage
based on the prediction of the first stage's NN.
To convert learned indexes to a standard form, we train a monolithic (larger) NN.

\item \textit{A note on broader impact}: using NNs for systems is a broad topic, but many existing works
lack strict safety guarantees. We believe that NN Verification can help system developers gain confidence
to apply NNs to critical systems. We hope our benchmark can be an early step toward this vision.

\end{itemize}

\paragraph*{Networks}
This benchmark has eleven networks with different parameters: two for learned
indexes, four for learned cardinality estimation and five for learned adaptive bitrate.
The learned index uses fully-connected feed-forward neural networks. The other
two---the learned cardinality and the learned adaptive bitrate---have a
relatively sophisticated internal structure. Please see our NN4SysBench
paper~\cite{lin2024nn4sysbench} for details.

\paragraph*{Specifications}
For learned indexes,
the specification aims to check if the prediction error is bounded.
The specification is a collection of pairs of input and output intervals such that
any input in the input interval should be mapped to the corresponding output interval.
For learned cardinality estimation and learned adaptive bitrate,
the specifications check the prediction error bounds (similar to the learned indexes)
and monotonicity of the networks.
By monotonicity specifications, we mean that for two inputs, the network should produce a larger
output for the larger input, which is required by cardinality estimation or adaptive bitrate.

\paragraph{Link:} \url{https://github.com/Khoury-srg/VNNComp26_NN4Sys}

\subsection{LinearizeNN}
\paragraph*{Proposed by}  Ali Arjomandbigdeli, Stanley Bak (Stony Brook University).
\paragraph*{Motivation}
Assuming having a neural network controller approximation with a piecewise linear model in the form of a set of linear models with added noise to account for local linearization error. The objective of this benchmark is to investigate the neural network output falls within the range we obtain from our linear model output plus some uncertainty. The idea of this benchmark came from one of our recent paper~\cite{ArjomandBigdeli2024} in which we approximated the NN controller with a piecewise linear model, and we wanted to check if the neural network output falls within the range we obtained from our linear model output plus some uncertainty.

\paragraph*{Networks} The neural network controller we used in this benchmark is an image-based controller for an Autonomous Aircraft Taxiing System whose goal is to control an aircraft's taxiing at a steady speed on a taxiway. This network was introduced  in the paper "Verification of Image-based Neural Network Controllers Using Generative Models"~\cite{katz2021veri}. The neural network integrates a concatenation of the cGAN (conditional GAN) and controller, resulting in a unified neural network controller with low-dimensional state inputs. In this problem, the inputs to the neural network consist of two state variables and two latent variables. The aircraft's state is determined by its crosstrack position (p) and heading angle error ($\theta$) with respect to the taxiway center line. Two latent variables with a range of -0.8 to 0.8 are introduced to account for environmental changes. Because in this case the output spec depends on both the input and output and considering the VNN-LIB limitation, we added a skip-connection layer to the neural network to have the input values present in the output space. We also added one linear layer after that to create a linear equation for each local model.

\paragraph*{Specifications} As mentioned earlier, the aim of this benchmark is to examine whether the neural network output stays within the range defined by the linear model's output, including a margin for uncertainty.Given input $x \in X$ and output $Y = f_{NN}(x)$, the query is of the form: $A_{mat}\times X + b + U_{lb} \leq Y \leq A_{mat}\times X + b + U_{ub}$ for each linear model in its abstraction region.

\paragraph*{Link} \url{https://github.com/aliabigdeli/LinearizeNN_benchmark2024}






\subsection{ml4acopf}
\paragraph*{Proposed by} Haoruo Zhao, Michael Klamkin, Mathieu Tanneau, Wenbo Chen, and Pascal Van Hentenryck (Georgia Institute of Technology), and Hassan Hijazi, Juston Moore, and Haydn Jones (Los Alamos National Laboratory).

\paragraph*{Motivation}
Machine learning models are utilized to predict solutions for an optimization model known as AC Optimal Power Flow (ACOPF) in the power system. Since the solutions are continuous, a regression model is employed. The objective is to evaluate the quality of these machine learning model predictions, specifically by determining whether they satisfy the constraints of the optimization model. Given the challenges in meeting some constraints, the goal is to verify whether the worst-case violations of these constraints are within an acceptable tolerance level.

\paragraph*{Networks}
The neural network designed comprises two components. The first component predicts the solutions of the optimization model, while the second evaluates the violation of each constraint that needs checking. The first component consists solely of general matrix multiplication (GEMM) and rectified linear unit (ReLU) operators. However, the second component has a more complex structure, as it involves evaluating the violation of AC constraints using nonlinear functions, including sigmoid, quadratic, and trigonometric functions such as sine and cosine. This complex evaluation component is incorporated into the network due to a limitation of the VNNLIB format, which does not support trigonometric functions. Therefore, these constraints violation evaluation are included in the neural network.

\paragraph*{Specifications}
In this benchmark, four different properties are checked, each corresponding to a type of constraint violation:
\begin{enumerate}
    \item Power balance constraints: the net power at each bus node is equal to the sum of the power flows in the branches connected to that node.
    \item Thermal limit constraints: power flow on a transmission line is within its maximum and minimum limits.
    \item Generation bounds: a generator's active and reactive power output is within its maximum and minimum limits.
    \item Voltage magnitude bounds: a voltage's magnitude output is within its maximum and minimum limits.
\end{enumerate}

The input to the model is the active and reactive load. The chosen input point for perturbation is a load profile for which a corresponding feasible solution to the ACOPF problem is known to exist. For the feasibility check, the input load undergoes perturbation. Although this perturbation does not exactly match physical laws, the objective is to ascertain whether a machine learning-predicted solution with the perturbation can produce a solution that does not significantly violate the constraints.

The scale of the perturbation and the violation threshold are altered by testing whether an adversarial example can be easily found using projected gradient descent with the given perturbation. The benchmark, provided with a fixed random seed, is robust against the simple projected gradient descent that is implemented.

\paragraph*{Link} \url{https://github.com/AI4OPT/ml4acopf_benchmark}

\subsection{ViT}
\paragraph*{Proposed by} the $\alpha,\!\beta$-CROWN team.
\paragraph*{Motivation}
Transformers~\cite{vaswani2017attention} based on the self-attention mechanism have much more complicated architectures and contain more kinds of nonlinerities, compared to simple feedforward networks with relatively simple activation functions. 
It makes verifying Transformers challenging. We aim to encourage the development of verification techniques for Transformer-based models, and we also aim to benchmark neural network verifiers on relatively complicated neural network architectures and more general nonlinearities. Therefore, we propose a new benchmark with Vision Transformers (ViTs)~\cite{dosovitskiy2020image}. This benchmark is developed based on our work on neural network verification for models with general nonlinearities~\cite{shi2024genbab}.

\paragraph*{Networks}
The benchmark contains two ViTs, as shown in \Cref{tab:vits}.
Considering the difficulty of verifying ViTs, we modify the ViTs and make the models relatively shallow and narrow, with significantly reduced number of layers and attention heads.
Following \cite{shi2019robustness}, we also replace the layer normalization with batch normalization.
The models are mainly trained with PGD training~\cite{madry2017towards}, and we also add a weighted IBP~\cite{gowal2018effectiveness,shi2021fast} loss for one of the models as a regularization.

\begin{table}[ht]
\centering
\caption{Networks in the ViT benchmark.}
\label{tab:vits}
\begin{tabular}{ccc}
\toprule 
Model & \texttt{PGD\_2\_3\_16} & \texttt{IBP\_3\_3\_8} \\
\midrule
Layers & 2 & 3\\
Attention heads & 3 & 3\\
Patch size & 16 & 8\\
Weight of IBP loss & 0 & 0.01\\
Training $\epsilon$ & $\frac{2}{255}$ & $\frac{1}{255}$\\
Clean accuracy & 59.78\% & 62.21\%\\
\bottomrule
\end{tabular}
\end{table}

\paragraph*{Specifications} 
The specifications are generated from the robustness verification problem with $\ell_\infty$ perturbation. 
We use the CIFAR-10 dataset with perturbation size $\epsilon=\frac{1}{255}$ at test time.
We have filtered the CIFAR-10 test set to exclude instances where either adversarial examples can be found (by PGD attack~\cite{madry2017towards} with 100 steps and 1000 restarts) or the vanilla CROWN-like method~\cite{zhang2018efficient,shi2019robustness} can already easily verify. 
We randomly keep 100 instances for each model, with a timeout threshold of 100 seconds. 
Note that since instances with adversarial examples have mostly been excluded during the filtering process, this version of the benchmark may not be able to reflect soundness issues in verifiers, and we refer readers to \cite{zhou2024testing} for discussions on testing soundness with models including ViT.

\paragraph*{Link} \url{https://github.com/shizhouxing/ViT_vnncomp2023}

\subsection{LSNC-ReLU}
\paragraph*{Proposed by} the $\alpha,\!\beta$-CROWN team.
\paragraph*{Motivation}
We develop a benchmark for the problem of verifying the Lyapunov stability of NN controllers in nonlinear dynamical systems within a region-of-intrest and a region-of-attraction. This is important for providing stability guarantees that are essential for safety-critical applications with NN controllers. It is also a useful application of neural network verification as recently demonstrated in \cite{yang2024lyapunov,shi2024certified}, and we refer readers to those works for more details on the problem.

\paragraph*{Networks and Specifications}
The NN controller and Lyapunov function are trained using the technique presented in CT-BaB~\cite{shi2024certified}. Compared to a general nonlinear dynamics as in \cite{yang2024lyapunov}, we utilized a neural-network dynamics. More specifically, the dynamics of this model is given by a shallow neural network with ReLU activations that are trained to approximate the 2D quadrotor dynamics. In this way, only ReLU nonlinearity is presented in the network (no trig functions, leakyrelu, etc.). The verification goal is to certify the Lyapunov stability of the NN controller in a nonlinear dynamical system within a certain Lyapunov sublevel set $(V < c)$. Specifications for the benchmark are randomly generated and consist of random rings of the form $x: c_1 < V(x) < c_2$.

\paragraph*{Link} \url{https://github.com/xiangruzh/LSNC-ReLU}

\subsection{Collins-RUL-CNN}
\paragraph*{Proposed by} Collins Aerospace, Applied Research \& Technology (\href{https://www.collinsaerospace.com/what-we-do/capabilities/technology-and-innovation/applied-research-and-technology}{website}).

\paragraph*{Motivation} Machine Learning (ML) is a disruptive technology for the aviation industry. This particularly concerns safety-critical aircraft functions, where high-assurance design and verification methods have to be used in order to obtain approval from certification authorities for the new ML-based products. Assessment of correctness and robustness of trained models, such as neural networks, is a crucial step for demonstrating the absence of unintended functionalities~\cite{ForMuLA, kirov2023formal}. The key motivation for providing this benchmark is to strengthen the interaction between the VNN community and the aerospace industry by providing a realistic use case for neural networks in future avionics systems~\cite{kirov2023benchmark}.

\paragraph*{Application} Remaining Useful Life (RUL) is a widely used metric in Prognostics and Health Management (PHM) that manifests the remaining lifetime of a component (e.g., mechanical bearing, hydraulic pump, aircraft engine). RUL is used for Condition-Based Maintenance (CBM) to support aircraft maintenance and flight preparation. It contributes to such tasks as augmented manual inspection of components and scheduling of maintenance cycles for components, such as repair or replacement, thus moving from preventive maintenance to \emph{predictive} maintenance (do maintenance only when needed, based on component’s current condition and estimated future condition). This could allow to eliminate or extend service operations and inspection periods, optimize component servicing (e.g., lubricant replacement), generate inspection and maintenance schedules, and obtain significant cost savings. Finally, RUL function can also be used in airborne (in-flight) applications to dynamically inform pilots on the health state of aircraft components during flight. Multivariate time series data is often used as RUL function input, for example, measurements from a set of sensors monitoring the component state, taken at several subsequent time steps (within a time window). Additional inputs may include information about the current flight phase, mission, and environment. Such highly multi-dimensional input space motivates the use of Deep Learning (DL) solutions with their capabilities of performing automatic feature extraction from raw data.

\paragraph*{Networks} The benchmark includes 3 convolutional neural networks (CNNs) of different complexity: different numbers of filters and different sizes of the input space. All networks contain only convolutional and fully connected layers with ReLU activations. All CNNs perform the regression function. They have been trained on the same dataset (time series data for mechanical component degradation during flight).

\paragraph*{Specifications} We propose 3 properties for the NN-based RUL estimation function. First, two properties (robustness and monotonicity) are local, i.e., defined around a given point. We provide a script with an adjustable random seed that can generate these properties around input points randomly picked from a test dataset. For robustness properties, the input perturbation (delta) is varied between 5\% and 40\%, while the number of perturbed inputs varies between 2 and 16. For monotonicity properties, monotonic shifts between 5\% and 20\% from a given point are considered. Properties of the last type ("if-then") require the output (RUL) to be in an expected value range given certain input ranges. Several if-then properties of different complexity are provided (depending on range widths).

\paragraph*{Link} \url{https://github.com/loonwerks/vnncomp2022}

\paragraph*{Paper} Available in~\cite{kirov2023benchmark} or on request.
\subsection{VGGNET16}
\paragraph*{Proposed by} Stanley Bak, Stony Brook University

\paragraph*{Motivation} This benchmark tries to scale up the size of networks being analyzed by using the well-studied VGGNET-16 architecture~\cite{simonyan2014very} that runs on ImageNet. Input-output properties are proposed on pixel-level perturbations that can lead to image misclassification. 

\paragraph*{Networks} All properties are run on the same network, which includes 138 million parameters. The network features convolution layers, ReLU activation functions, as well as max pooling layers.

\paragraph*{Specifications} Properties analyzed ranged from single-pixel perturbations to perturbations on all 150528 pixles (L-infinity perturbations). A subset of the images was used to create the specifications, one from each category, which was randomly chosen to attack. Pixels to perturb were also randomly selected according to a random seed.

\paragraph*{Link} \url{https://github.com/stanleybak/vggnet16_benchmark2022/}

\subsection{Traffic Signs Recognition}
\paragraph*{Proposed by} M\u{a}d\u{a}lina Era\c{s}cu and Andreea Postovan (West University of Timisoara, Romania)
\paragraph*{Motivation} Traffic signs play a crucial role in ensuring road safety and managing traffic flow in both city and highway driving. The recognition of these signs, a vital component of autonomous driving vision systems, faces challenges such as susceptibility to adversarial examples~\cite{szegedy2013intriguing} and occlusions~\cite{zhang2020lightweight}, stemming from diverse traffic scene conditions.

\paragraph*{Networks} Binary neural networks (BNNs) show promise in computationally limited and energy-constrained environments within the realm of autonomous driving~\cite{hubara2016binarized}. BNNs, where weights and/or activations are binarized to $\pm 1$, offer reduced model size and simplified convolution operations for image recognition compared to traditional neural networks (NNs).

We trained and tested various BNN architectures using the German Traffic Sign Recognition Benchmark (GTSRB) dataset~\cite{GTSRB}. This multi-class dataset, containing images of German road signs across 43 classes, poses challenges for both humans and models due to factors like perspective change, shade, color degradation, and lighting conditions. The dataset was also tested using the Belgian Traffic Signs \cite{BelgianTrafficSignDatabase} and Chinese Traffic Signs \cite{ChineseTrafficSignDatabase} datasets. The Belgium Traffic Signs dataset, with 62 classes, had 23 overlapping classes with GTSRB. The Chinese Traffic Signs dataset, with 58 classes, shared 15 classes with GTSRB. Pre-processing steps involved relabeling classes in the Belgium and Chinese datasets to match those in GTSRB and eliminating non-overlapping classes (see \cite{postovan2023architecturing} for details).

\begin{figure}[h]
  \centering
    \includegraphics[width=0.7\textwidth]{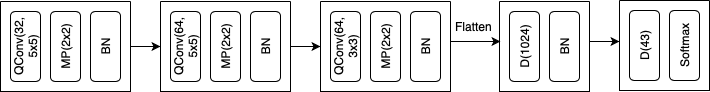}
    \caption{Accuracy Efficient Architecture for GTSRB and Belgium dataset}
    \label{fig:Acc-Efficient-Arch-GTSRB-Belgium}
\end{figure}

\begin{figure}[h]
  \centering
    \includegraphics[width=0.7\textwidth]{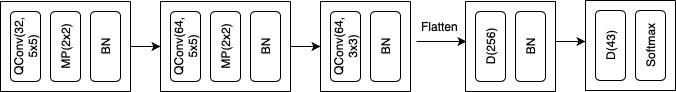}
    \caption{Accuracy Efficient Architecture for Chinese dataset}
    \label{fig:Acc-Efficient-Arch-Chinese}
\end{figure}

\begin{figure}[h]
  \centering
    \includegraphics[width=0.3\textwidth]{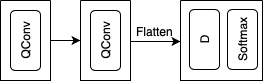}
    \caption{XNOR(QConv) architecture}
    \label{fig:XNOR(QConv)-arch}
\end{figure}

We provide three models with the structure in Figures \ref{fig:Acc-Efficient-Arch-GTSRB-Belgium}, \ref{fig:Acc-Efficient-Arch-Chinese}, and \ref{fig:XNOR(QConv)-arch}. They contain QConv, Batch Normalization (BN), Max Pooling (ML), Fully Connected/Dense (D) layers.  Note that the QConv layer binarizes the corresponding convolutional layer. All models were trained for 30 epochs. The model from Figure \ref{fig:Acc-Efficient-Arch-GTSRB-Belgium} was trained with images having the dimension 64px x 64 px, the one from Figure \ref{fig:Acc-Efficient-Arch-Chinese} with 48px x 48 px and the one from Figure \ref{fig:XNOR(QConv)-arch} with 30px x 30 px. The two models involving Batch Normalization layers introduce real valued parameters besides the binary ones, while the third one contains only binary parameters (see Table \ref{tab:stats}) for statistics.

\begin{table}[h]
\caption{Training and Testing Statistics}
\label{tab:stats}
\centering
\scriptsize
\begin{tabular}{|c|c|ccc|ccc|}
\hline
\multirow{2}{*}{\textbf{Input size}} & \multirow{2}{*}{\textbf{Model name}} & \multicolumn{3}{c|}{\textbf{Accuracy}}                                      & \multicolumn{3}{c|}{\textbf{\#Params}}                                      \\ \cline{3-8} 
                            &                             & \multicolumn{1}{c|}{\textbf{German}} & \multicolumn{1}{c|}{\textbf{China}} & \textbf{Belgium} & \multicolumn{1}{c|}{\textbf{Binary}}  & \multicolumn{1}{c|}{\textbf{Real}} & \textbf{Total}   \\ \hline
64px $\times$ 64px          & Figure \ref{fig:Acc-Efficient-Arch-GTSRB-Belgium}                  & \multicolumn{1}{c|}{96.45}  & \multicolumn{1}{c|}{81.50} & 88.17   & \multicolumn{1}{c|}{1772896} & \multicolumn{1}{c|}{2368} & 1775264 \\ \hline
48px $\times$ 48px          & Figure \ref{fig:Acc-Efficient-Arch-Chinese}                  & \multicolumn{1}{c|}{95.28}  & \multicolumn{1}{c|}{83.90} & 87.78   & \multicolumn{1}{c|}{904288}  & \multicolumn{1}{c|}{832}  & 905120  \\ \hline
30px $\times$ 30px          & Figure \ref{fig:XNOR(QConv)-arch}                  & \multicolumn{1}{c|}{81.54}  & \multicolumn{1}{c|}{N/A}   & N/A     & \multicolumn{1}{c|}{1005584} & \multicolumn{1}{c|}{0}    & 1005584 \\ \hline
\end{tabular}
\end{table}
\paragraph*{Specifications} To evaluate the \emph{adversarial robustness} of the networks above, we assessed perturbations within the infinity norm around zero, with the radius denoted as $\epsilon = \{1, 3, 5, 10, 15\}$. This involved randomly selecting three distinct images from the GTSRB dataset's test set for each model and generating \textsc{VNNLIB} files for each epsilon in the set. In total, we created 45 \textsc{VNNLIB} files. Due to a 6-hour total timeout constraint for solving all instances, each instance had a maximum timeout of 480 seconds. To review the generated \textsc{VNNLIB} specification files submitted to VNNCOMP 2023, as well as to generate new ones, please refer to \url{https://github.com/apostovan21/vnncomp2023}.

\paragraph*{Link} \url{https://github.com/apostovan21/vnncomp2023}

\subsection{CIFAR100}

\paragraph*{Proposed by} the $\alpha,\!\beta$-CROWN team.
\paragraph*{Motivation} This benchmark is reused from VNN-COMP 2022 with a reduced complexity (only two out of the four models with medium sizes are retained). 
See details in Section 4.5 of the report of VNN-COMP 2022~\cite{muller2022vnncomp}.

\paragraph*{Networks} We provide two ResNet models on CIFAR-100 with different model widths and depths (input dimension $32 \times 32 \times 3$, 100 classes):
\begin{itemize}
    \item \texttt{CIFAR100-ResNet-medium}: 8 residual blocks, 17 convolutional layers + 2 linear layers
    \item \texttt{CIFAR100-ResNet-large}: 8 residual blocks, 19 convolutional layers + 2 linear layers (almost identical to standard ResNet-18 architecture)
\end{itemize}

\paragraph*{Specifications} 
We randomly select 100 images from the CIFAR-100 test set with a verification timeout of 100 seconds for each of the two models. 
We filtered out the samples which can be verified by vanilla CROWN (which is used during training) to make the benchmark more challenging. The filtering process is done offline on a machine with a GPU due to the large sizes of these models. 
A small proportion of instances (around 18\%) with adversarial examples have been retained for potentially identifying unsound results. 

\paragraph*{Link} \url{https://github.com/huanzhang12/vnncomp2024_cifar100_benchmark}

\subsection{TinyImagenet}

\paragraph*{Proposed by} the $\alpha,\!\beta$-CROWN team.
\paragraph*{Motivation} This benchmark is reused from VNN-COMP 2022. See details in Section 4.5 of the report of VNN-COMP 2022~\cite{muller2022vnncomp}.

\paragraph*{Networks} We provide a ResNet for TinyImageNet (input dimension $64 \times 64 \times 3$, 200 classes):
\begin{itemize}
    \item \texttt{TinyImageNet-ResNet-medium}: 8 residual blocks, 17 convolutional layers + 2 linear layers
\end{itemize}

\paragraph*{Specifications} 
We randomly select 200 images from the TinyImageNet test set with a verification timeout of 100 seconds for each of the two models. A filtering procedure has been adopted similar to the CIFAR100 benchmark.

\paragraph*{Link} \url{https://github.com/huanzhang12/vnncomp2024_tinyimagenet_benchmark}

\subsection{Yolo}
\paragraph*{Proposed by} the $\alpha,\!\beta$-CROWN team.

\paragraph*{Motivation} Object detection is central to many safety-critical applications, yet verification research has had limited benchmarks that capture its unique challenges. Even in a simplified form, YOLO highlights key aspects of detection tasks: multiple outputs, spatial reasoning over bounding boxes, and the need to ensure reliable perception under perturbations. A YOLO benchmark therefore provides a concrete step toward advancing verification methods for complex perception tasks such as object detection.

\paragraph*{Networks} The model is a modified version of YOLOv2. For simplicity of verification, the backbone of the original YOLOv2 model is replaced with a much smaller network. The network takes in an image ($3\times 52\times 52$) and outputs a tensor of a shape $125\times 13\times 13$, which contains the confidence scores, the classification result, and the positions of 5 bounding boxes for each of the $13 \times 13$ grid cells. The network was trained using adversarial training, incorporating an $\ell_{1}$ regularization term in the loss.

\paragraph*{Specifications} Robustness properties are defined on the inputs and raw output tensors of the backbone network. For each instance, the property is verified if there is at least one bounding box successfully predicting the same object (with a confidence score greater than a preset threshold) given any bounded random perturbations on the input image.

\paragraph*{Link} \url{https://github.com/xiangruzh/Yolo-Benchmark}

\subsection{SoundnessBench}
\paragraph*{Proposed by} the $\alpha,\!\beta$-CROWN team.
\paragraph*{Motivation}
We introduce SoundnessBench, a novel benchmark designed to evaluate the  soundness of neural network (NN) verifiers, inspired by the work~\cite{zhou2024testing}. While recent years have seen significant progress in developing both NN verifiers and benchmarks, existing benchmarks typically lack ground truth for challenging instances. As a result, it remains difficult to determine whether a verification claim is sound when no verifier can falsify the property under consideration.

\paragraph*{Networks} SoundnessBench includes a simple feedforward neural network and a residual network with minimal architectures and standard operators widely supported by mainstream verifiers. These operators include fully connected (linear) layers, 2D convolutional layers, ReLU activations, as well as reshaping and flattening operations between 2D and 1D formats.

\paragraph*{Specifications} The verification tasks are synthetic, with some specifications intentionally designed to embed counterexamples that are difficult for standard falsification methods—such as projected gradient descent (PGD)—to discover. Specifications are expressed as conjunctions of clauses.

\paragraph*{Link} \url{https://github.com/Kevinyang555/SoundnessBench_vnncomp2026}

\subsection{Relusplitter}

\paragraph{Proposed by} Linhan Li and ThanhVu Nguyen (George Mason University)
\paragraph{Motivation}
Many existing DNN verification benchmarks are becoming ineffective at evaluating modern verifiers because their instances are often too easy or too hard and lack architectural diversity. To address these issues and allow for the reuse of established benchmarks, we propose Relusplitter \cite{Li2025destabilizing}. It takes a DNN verification instance (network and property) and generates a semantically equivalent network on which the property is harder to verify. Relusplitter achieves this by systematically destabilizing stable neurons in the network.

\paragraph{Description}
This benchmark uses selected instances from benchmarks used in previous VNNCOMP iterations as seed inputs to generate more challenging instances. The networks include both fully connected (FC) and convolutional (Conv) architectures. All networks use a standard feed-forward structure. The timeout for the instances was reduced based on results from prior iterations, and the generated instances have 3x timeout. 
Benchmarks used as seed inputs include (1) ACAS Xu, (2) MNIST\_FC, (3) Oval21, and (4) Cifar Biasfield.

\paragraph{Networks} The seed instances use the original network, and the generated instances use a destabilized network generated from the original network.
\paragraph{Specifications} The original specification from the seed instance was used for each generated instance. 
\paragraph*{Link} \url{https://github.com/dynaroars/relusplitter/tree/VNNCOMP-25}

\subsection{TLL Verify Bench}
\paragraph*{Proposed by} James Ferlez (University of California, Irvine)

\paragraph*{Motivation} This benchmark consists of Two-Level Lattice (TLL) NNs, which have been shown to be amenable to fast verification algorithms (e.g. \cite{FerlezKS22}). Thus, this benchmark was proposed as a means of comparing TLL-specific verification algorithms with general-purpose NN verification algorithms (i.e. algorithms that can verify arbitrary deep, fully-connected ReLU NNs).

\paragraph*{Networks}  The networks in this benchmark are a subset of the ones used in \cite[Experiment 3]{FerlezKS22}. Each of these TLL NNs has $n=2$ inputs and $m=1$ output. The architecture of a TLL NN is further specified by two parameters: $N$, the number of local linear functions, and $M$, the number of selector sets. This benchmark contains TLLs of sizes $N = M = 8, 16, 24, 32, 40, 48, 56, 64$, with $30$ randomly generated examples of each (the generation procedure is described in \cite[Section 6.1.1]{FerlezKS22}). At runtime, the specified verification timeout determines how many of these networks are included in the benchmark so as to achieve an overall 6-hour run time; this selection process is deterministic. Finally, a TLL NN has a natural representation using multiple computation paths \cite[Figure 1]{FerlezKS22}, but many tools are only compatible with fully-connected networks. Hence, the ONNX models in this benchmark implement TLL NNs by ``stacking'' these computation paths to make a fully connected NN (leading to sparse weight matrices: i.e. with many zero weights and biases). The \texttt{TLLnet} class (\url{https://github.com/jferlez/TLLnet}) contains the code necessary to generate these implementations via the \texttt{exportONNX} method.

\paragraph*{Specifications}  All specifications have as input constraints the hypercube $[-2,2]^2$. Since all networks have only a single output, the output properties consist of a randomly generated real number and a randomly generated inequality direction. Random output samples from the network are used to roughly ensure that the real number property has an equal likelihood of being within the output range of the NN and being outside of it (either above or below all NN outputs on the input constraint set). The inequality direction is generated independently and with each direction having an equal probability. This scheme biases the benchmark towards verification problems for which counterexamples exist. 

\paragraph*{Link} \url{https://github.com/jferlez/TLLVerifyBench}
\paragraph*{Commit}
199d2c26d0ec456e62906366b694a875a21ff7ef

\subsection{ACAS Xu}
\paragraph{Networks} The ACASXu benchmark consists of ten properties defined over 45 neural networks used to issue turn advisories to aircraft to avoid collisions. The neural networks have 300 neurons arranged in 6 layers, with ReLU activation functions. There are five inputs corresponding to the aircraft states, and five network outputs, where the minimum output is used as the turn advisory the system ultimately produces.

\paragraph{Specifications} We use the original 10 properties~\cite{katz2017reluplex}, where properties 1-4 are checked on all 45 networks as was done in later work by the original authors~\cite{katz2019marabou}. Properties 5-10 are checked on a single network. The total number of benchmarks is therefore 186. The original verification times ranged from seconds to days---including some benchmark instances that did not finish. This year we used a timeout of around two minutes (116 seconds) for each property, in order to fit within a total maximum runtime of six hours.





\subsection{Real-world distribution shifts}
\paragraph*{Proposed by} the Marabou team.
\paragraph*{Motivation}
While robustness against handcrafted perturbations (e.g., norm-bounded) for perception networks are more commonly investigated, robustness against real-world distribution shifts~\cite{wu2022toward} are less studied but of practical interests. This benchmark set contains queries for verifying the latter type of robustness.  
\paragraph*{Networks} The network is a concatenation of a generative model and a MNIST classifier. The generative model is trained to take in an unperturbed image and an embedding of a particular type of distribution shifts in latent space, and produce a perturbed image. The distribution shift captured in this case is the "shear" perturbation. 
\paragraph*{Specifications} The verification task is to certify that a classifier correctly classifies all images in a perturbation set, which is a set of images generated by the generative model given a fixed image and a ball centering the mean perturbations on this image (in the latent space). This mean perturbation is computed by a prior network.
\paragraph*{Link} \url{https://github.com/wu-haoze/dist-shift-vnn-comp}

\subsection{safeNLP}

\begin{figure}[htbp]
    \centering
    \includegraphics[width=0.9\columnwidth]{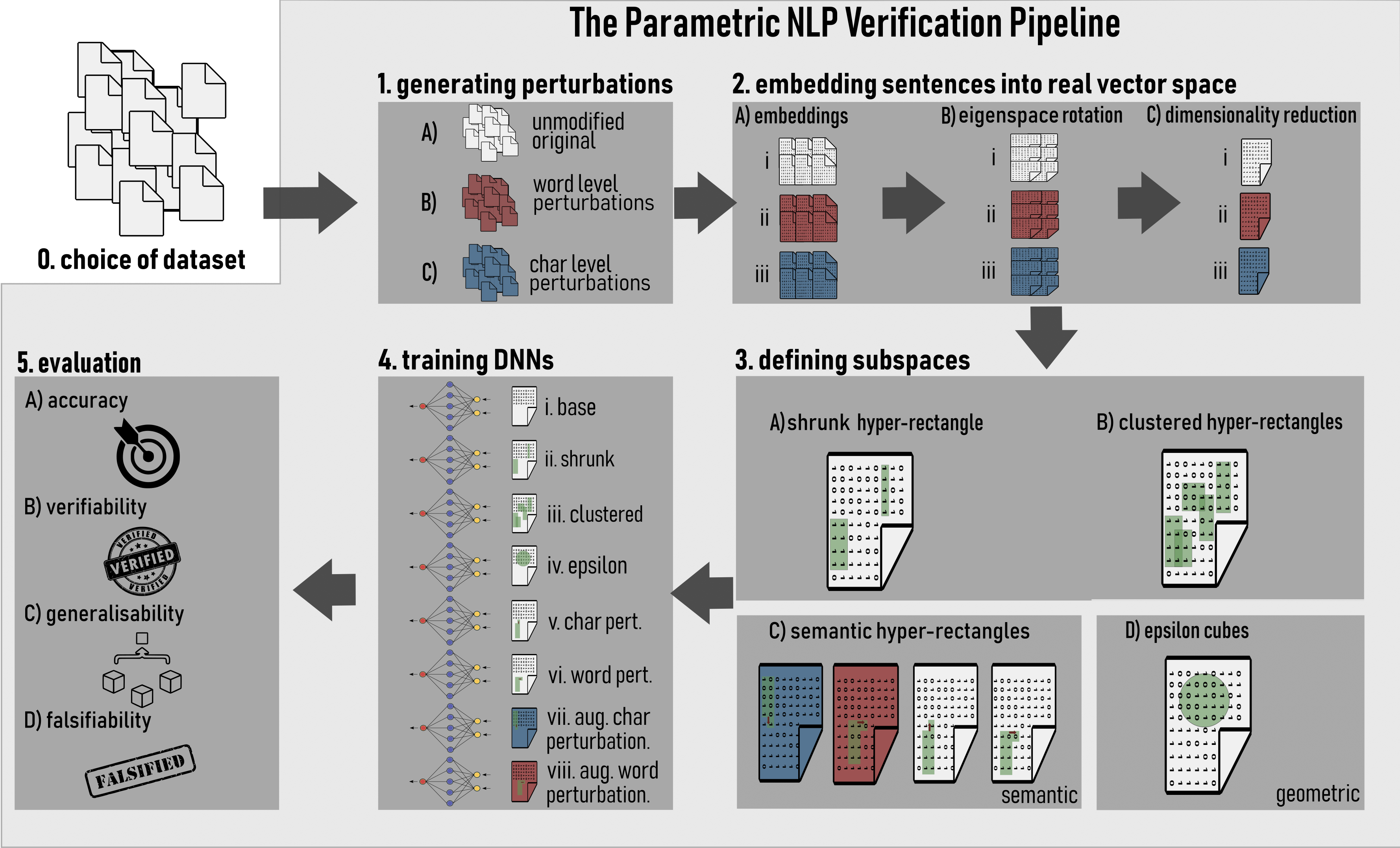}
    \caption{\small\emph{Generic approach to generating the NLP verification pipelines~\cite{casadio2023antonio,casadio2024nlp} deployed to obtain the safeNLP benchmark.}}
   \label{fig:antonio}
\end{figure}

\paragraph*{Proposed by} Marco Casadio, Ekaterina Komendantskaya, Luca Arnaboldi, Tanvi Dinkar.

\paragraph*{Motivation}
While considerable research has been dedicated to the verification of DNN-based systems in domains such as computer vision, there has been a notable lack of focus on the verification of natural language processing (NLP) systems. This is particularly critical given the rise of conversational agents across various domains, where inaccurate or misleading responses can cause real-world harm. For example, recent EU legislation~\cite{EUlaw} requires chatbots to disclose their non-human nature when queried, and developers of the chatbots should provide firm, and if possible, formal, guarantees that such disclosure will be given in an accurate manner. Medical assistants give another example where formal guarantees about the conversational agent responses are needed in order to safeguard against chatbots  generating harmful medical advice~\cite{bickmore2018patient}. While some initial work has been done in this area of NLP verification~\cite{jia2019certified,huang2019achieving,welbl2020towards,zhang2021certified,wang2023robustness,ko2019popqorn,du2021cert,shi2020robustness,bonaert2021fast}, no agreement on commonly  accepted benchmarks has been reached in this domain. To address this gap, we introduce safeNLP, the first such benchmark. 

\paragraph*{Application}
In~\cite{casadio2024nlp}, we have undertaken a large-scale study of the existing literature on  NLP verification, and distilled common patterns among the existing approaches. Usually, given a dataset consisting of sentences divided into classes, Large Language Models (LLMs) are used to embed these sentences into real-vector spaces, after which smaller neural networks are trained to classify the embedded vectors (relative to the originally given classes). For verification, one can generate meaning-preserving sentence perturbations, again embed them into vector spaces, and verify that subspaces that contain the (embeddings of) the perturbed sentences are classified correctly. Also, in line with classical verification pipelines~\cite{CasadioKDKKAR22}, one can use these input subspaces to train the neural network to be robust on them.  The problem was that each of the existing approaches~\cite{jia2019certified,huang2019achieving,welbl2020towards,zhang2021certified,wang2023robustness,ko2019popqorn,du2021cert,shi2020robustness,bonaert2021fast} used parts of this pipeline in different ways, which made it difficult to compare or audit the results. In~\cite{casadio2023antonio,casadio2024nlp}, we made a generic implementation of this pipeline, 
where each of the components of the pipeline is implemented in a modular and transparent way. For example, we can choose  and vary embedding functions, training modes, algorithms for sentence perturbations and algorithms for robust training, independently and modularly; as shown in Figure~\ref{fig:antonio}. This implementation was used to generate the presented VNNCOMP benchmark.

\begin{itemize}
    \item \emph{Datasets:} Although there was no clear consensus in~\cite{jia2019certified,huang2019achieving,welbl2020towards,zhang2021certified,wang2023robustness,ko2019popqorn,du2021cert,shi2020robustness,bonaert2021fast}, the most frequently used dataset in prior works was the IMDB dataset containing film reviews. Its disadvantage is unclear relation to safety critical domains that usually motivate verification efforts. On the other hand, none of the previously used datasets concerned safety-critical applications of NLP. We decided to address this problem, and therefore applied our generic NLP verification pipeline on two safety-critical datasets:  R-U-A-Robot~\cite{gros2021ruarobot}, which focuses on the chatbot disclosure problem, and Medical~\cite{abercrombie2022risk}, which addresses the issue of harmful advice provided by medical chatbots. Both datasets are pre-processed into two classes, positive and negative, to simplify the verification task. For further details on the pre-processing steps and datasets, see~\cite{casadio2024nlp} and the \href{https://github.com/ANTONIONLP/safeNLP}{benchmark GitHub repository}.
    \item \emph{Input Space:} In both datasets, sentences are transformed into fixed-size vector representations, i.e. embeddings, which serve as the inputs to the neural networks. For this VNNCOMP benchmark, we used Sentence-BERT~\cite{reimers-gurevych-2019-sentence}. 
    \item \emph{What to Verify:} For each dataset, we generated meaning-preseving sentence perturbations at character and word level as in Moradi et al.~\cite{moradi2021evaluating} and at sentence level with Vicuna~\cite{vicuna2023}. For each positive sentence in the dataset, the smallest hypercube containing the embeddings of all of its obtained perturbations formed one input subspace for verification. Such subspaces were obtained for all positive sentences from the given data set, and were subject to VNCCOMP verification challenge. 
    \item \emph{A note on broader impact:} Verified models can serve as filters for larger NLP systems: e.g. to screen inputs to ensure they meet safety criteria before being passed on to more complex models.
\end{itemize}

\paragraph*{Networks} The safeNLP benchmark includes two neural networks, each corresponding to a different dataset (R-U-A-Robot and Medical). Both networks share the same architecture, consisting of two fully-connected layers. The hidden layer has 128 units with a ReLU activation function, while the output layer has 2 units representing the two classification classes (positive/negative). To enhance the robustness of the networks to the specified safety requirements, they are trained using a custom PGD (Projected Gradient Descent)~\cite{madry2018towards} adversarial training technique. In particular, the PGD attack explores the above-mentioned subspaces of the input space (cf. also Figure~\ref{fig:antonio}). 

\paragraph*{Specifications} The benchmark uses hyper-rectangles in the 30-dimensional embedding space as the subspaces of choice, offering a computationally efficient way to define more precise and adaptable regions compared to the traditional $\epsilon$-cubes. 
The specifications require verifying that, for a given network and hyper-rectangle, every point within the hyper-rectangle is classified as the positive class by the network. To meet time constraints, we randomly select 1,080 such specifications, each linked to one of the two networks and a corresponding hyper-rectangle, with a timeout of 20 seconds per specification.

\paragraph*{Link} \url{https://github.com/ANTONIONLP/safeNLP}

\subsection{CORA Benchmark}
\paragraph*{Proposed by} the CORA team.
\paragraph*{Motivation}
The verification of neural networks can be quite slow, i.e., the verification of a single instance can take multiple days -- which is often hard to justify, particularly in safety-critical scenarios. To encourage the fast verification of neural networks, our benchmark focuses on the verification time by setting a small timeout and testing three different (adversarial) training techniques that aim to ease the verifiability.
\paragraph*{Networks} The benchmark consists of one ReLU-neural network architecture (7x250 + ReLU), which was trained on three datasets, (MNIST, SVHN, and CIFAR10), using three different (adversarial) training methods, i.e., standard (point), interval-bound propagation, and set-based. Both interval-bound propagation and set-based training are training methods that improve the robustness of the trained neural network and aim to ease later verification. The neural networks are taken from the first evaluation run of~\cite{koller_et_al_2025settraining}; please refer to~\cite{koller_et_al_2025settraining} for the training details.
\paragraph*{Specifications} All networks are trained on classification tasks. The goal is to verify that no image within a given input set is incorrectly classified.
\paragraph*{Link} \url{https://github.com/kollerlukas/cora-vnncomp2024-benchmark}

\subsection{SAT ReLU}

\paragraph{Proposed by} Edoardo Manino (The University of Manchester)

\paragraph{Motivation} Neural network verifiers are fundamentally not different from any other piece of software in that they may contain bugs. As such, it is necessary to test their behavior for correctness. To this end, the SAT ReLU benchmark is constructed with the additional requirement of having its verification verdicts (safe or unsafe) known a priori. With it, we can test all neural network verifiers for correctness.

\paragraph{Description} The SAT ReLU benchmark is entirely synthetic. Each instance is built by generating a random 3-SAT boolean formula with known satisfiability verdict. In total, we generate $50$ satisfiable formulae and $50$ unsatisfiable ones in CNF format. We vary the number of variables in $v=[2,100]$ and the number of clauses in $c=[v,5v]$. Then, we encode the formulae as neural network verification problems.

\paragraph{Networks} We encode each CNF formula as a fully-connected ReLU network with one hidden layer. To do so, we take inspiration from the ReLU gadgets introduced by \cite{katz2017reluplex}. First, we assume that each element of the input $x\in\mathbb{R}^v$ represents a boolean variable $x_i\in\{0,1\}$. Half of the network encodes the boolean arithmetic necessary to compute the CNF formula under this assumption. Second, we let each variable vary in $x_i\in[0,1]$ and use the other half of the network to encode the necessary binarization constraints. Finally, we merge the two halves into a single network. The final network has $v$ inputs, $c+2v$ hidden activations, and $2$ outputs.

\paragraph{Specifications} Given the nature of the networks we construct, the original CNF formula is satisfiable if and only if there exists some input $x\in\{0,1\}^n$ that produces an output of $y_1\geq1$. Non-binary inputs cause the second output to be positive, and thus are rejected by setting a binarization constraint of $y_2\leq0$.

\paragraph*{Link} \url{https://github.com/emanino/neurocodebench}

\subsection{MetaRoom}
\paragraph*{Proposed by} Hanjiang Hu (CMU), Zuxin Liu (CMU), Linyi Li (UIUC), Jiacheng Zhu (CMU), Ding Zhao (CMU).

\paragraph*{Motivation} MetaRoom Benchmark is based on the MetaRoom dataset, which aims at the robustness certification and verification of deep learning based vision models against camera motion perturbation in the indoor robotics application. Specifically, this benchmark focuses on the classification of 20 labels against camera moves within tiny perturbation radii of translation along the z-axis and rotation along the y-axis (e.g. 1e-5 m, 2.5e-4 degree). The MetaRoom dataset is a realistic indoor object recognition dataset with dense point cloud maps, which enables image projections from different camera poses and simulates object-centric semantic robustness in real-world robotic applications. More details about the dataset can be found in \cite{hu2022robustness}.

\paragraph*{Networks} The models of \texttt{cnn\_4layer} and \texttt{cnn\_6layer} are 4-layer and 6-layer convolutional neural networks with a last layer of a fully connected layer for multi-class classification. The input shape is \texttt{1*3*32*56}, which is projected from  one-dimensional relative camera rotation or translation around 0 (e.g. [-1e-5, 1e-5]) from a very dense point cloud based on direct Z-buffering. And the output is for the classification of 20 labels. The pretrained models give correct predictions at camera motion movement origins. 

\paragraph*{Specifications} Simplified from the 1-dim semantic perturbation studied in \cite{hu2023robustness,hu2024pixel}, the benchmark for VNN-COMP adopts the pre-projected image specifications from a very dense point cloud, where the lower and upper bounds are all pixels of the image during the projection within the camera perturbation radius of translation along the z-axis or rotation along the y-axis. The output specifications will be the classification labels at camera movement origins.

\paragraph*{Link} \url{https://github.com/HanjiangHu/metaroom_vnn_comp2023}

\subsection{cersyve}
\paragraph*{Proposed by} Yujie Yang (THU, CMU), Hanjiang Hu (CMU), Tianhao Wei (CMU), Shengbo Eben Li (THU), Changliu Liu (CMU).

\paragraph*{Motivation} Neural safety certificate verification is a special type of verification problems that require neural networks to satisfy certain properties everywhere in the state space, i.e., the input set of the verification problem is the entire state space. This is different from most existing benchmarks, which only consider verification in either a small disturbance set around data samples or part of the state space, such as MNIST, CIFAR, and ACAS Xu. Another distinct feature of safety certificate verification is that it not only involves a single safety certificate network but also needs system dynamics and control policies for verification, and certain conversions are required before these components can be formulated as a standard verification problem.

\paragraph*{Description} While this version of cersyve only contains necessary networks and specifications for verification, our full version also includes a set of neural safety certificate synthesis tools, including pre-training, adversarial training, and verification-guided training modules, as well as evaluation tools for synthesized certificates. For a detailed description of cersyve, please refer to our paper~\cite{yang2025scalable}.

\paragraph*{Networks} The cersyve benchmark contains nine commonly used control tasks with state dimensions ranging from two to six. These tasks include both linear and nonlinear dynamics and safety constraints. We provide two ONNX models for each task. One is a pre-trained safety certificate, and the other is fine-tuned. The models are already integrated with all necessary elements for verification, such as system dynamics, constraints, and control policies. Users can view each of them as a single neural network for standard verification.

\paragraph*{Specifications} The cersyve benchmark requires to verify that within a bounded state space, whether the constraint satisfaction and forward invariance properties of a neural safety certificate is satisfied for all states. For a detailed description of the verification problem formulation, please refer to our paper~\cite{yang2025scalable}. The input specifications are lower and upper bounds on each dimension of the state. The output specifications are always two inequalities corresponding to the two properties.

\paragraph*{Link} \url{https://github.com/intelligent-control-lab/cersyve.jl/tree/VNN-COMP-2025}

\subsection{Challenging Certified Training}
\paragraph*{Proposed by} Konstantin Kaulen (RWTH Aachen University)
\paragraph*{Motivation} Using recent advances in certified training, we obtained state-of-the-art certifiably robust image classifiers using the CTRAIN library for certified training~\cite{kaulen2025ctrain,kaulen2026rethinking}. Their standard and certified accuracies are summarized in Table~\ref{tab_challenging_certified_training}. Despite this progress, we observed that complete verification of these models is highly computationally demanding. When evaluating all test-set properties using $\alpha,\beta$-CROWN with a cutoff of 1,000 seconds, 2,994 out of 60,000 verification tasks timed out, with a timeout rate of up to 14.33\%.
Thus, complete verification is becoming a major computational bottleneck for further progress in certifiably robust neural networks: improvements in certified training performance are of limited practical value if the resulting guarantees cannot be established within reasonable computational budgets. The Challenging Certified Training benchmark therefore provides difficult and practically relevant instances for measuring progress in complete neural network verification.

\begin{table}[t]
\centering
\caption{Performance and complete-verification difficulty of the models underlying the Challenging Certified Training benchmark. Verification results were obtained with $\alpha,\beta$-CROWN using a 1,000-second cutoff on 10,000 test-set properties per model.}
\label{tab_challenging_certified_training}
\begin{tabular}{llrrr}
\toprule
Dataset / $\epsilon$ & Network & Standard acc. & Certified acc. & Timeouts \\
\midrule
CIFAR-10, $2/255$ & CNN-7      & 83.54\% & 66.04\% & 658 (6.58\%) \\
CIFAR-10, $2/255$ & Wide CNN-7 & 85.37\% & 60.28\% & 1,433 (14.33\%) \\
CIFAR-10, $8/255$ & CNN-7      & 57.19\% & 35.49\% & 134 (1.34\%) \\
CIFAR-10, $8/255$ & Wide CNN-7 & 57.74\% & 34.60\% & 132 (1.32\%) \\
TinyImageNet, $1/255$ & CNN-7      & 41.21\% & 27.81\% & 236 (2.36\%) \\
TinyImageNet, $1/255$ & Wide CNN-7 & 41.89\% & 28.35\% & 401 (4.01\%) \\
\midrule
\multicolumn{4}{r}{Total timeouts} & 2,994 (4.99\%) \\
\bottomrule
\end{tabular}
\end{table}

\paragraph*{Description}
The benchmark contains 60 local-robustness instances derived from the six models shown in Table~\ref{tab_challenging_certified_training}. To obtain instances spanning a broad range of verification difficulty, all 10,000 test-set properties for each model were first evaluated using $\alpha,\beta$-CROWN with its standard configuration and a cutoff time of 1,000 seconds. The resulting instances were stratified according to their observed verification runtime into four categories: $[0,10]$ seconds, $[10,100]$ seconds, $[100,1000]$ seconds, and timeouts. For each model, we sample two instances from each of the first two categories and three instances from each of the latter two categories, yielding ten properties per model and 60 benchmark instances overall. This construction combines moderately difficult properties with instances that remain unsolved by a state-of-the-art verifier within a substantial computational budget.

\paragraph*{Networks}
The benchmark comprises six convolutional neural networks trained using MTL-IBP~\cite{DePalmaConexCombinations}. For CIFAR-10, it contains CNN-7 architectures for $\ell_\infty$ perturbation radii of $\epsilon=2/255$ and $\epsilon=8/255$. For TinyImageNet, it contains corresponding standard and width-scaled CNN-7 architectures for $\epsilon=1/255$. Their standard and certified accuracies are reported in Table~\ref{tab_challenging_certified_training}.

\paragraph*{Specifications} Each specification encodes standard $\ell_\infty$ local robustness of a CIFAR-10 or TinyImageNet test image at the perturbation radius corresponding to the respective model.
\paragraph*{Link}\url{https://github.com/kkaulen/vnncomp_challenging_certified_training}
\subsection{Additional Benchmarks}

We have not yet obtained benchmark descriptions for other benchmark sets reported in Table \ref{tab:benchmarks}. We will update the report when these descriptions are available.
Artifacts of the benchmarks are available in the repository\footnote{\url{https://github.com/VNN-COMP/vnncomp2026_benchmarks}}. 

\newpage
\section{Results}
\label{sec:results}

Each tool was run on each of the benchmarks and produced a \texttt{csv} result file, that was provided as feedback to the tool authors using the online execution platform.
The final \texttt{csv} files for each tool as well as scoring scripts are available online: \url{https://github.com/VNN-COMP/vnncomp2026_results}.
The results were automatically analyzed to compute scores and create the statistics presented in this section, where a tool is omitted from individual benchmark results if it timed out on all instances.  

\subsection{Regular Track}
The regular track contained all the benchmarks that were voted for by at least half of all participants.


\begin{table}[h]
\begin{center}
\caption{Overall Score} \label{tab:score_regular}
{\setlength{\tabcolsep}{2pt}
\begin{tabular}[h]{@{}lll@{}}
\toprule
\textbf{\# ~} & \textbf{Tool} & \textbf{Score}\\
\midrule
1 & $\alpha$-$\beta$-CROWN & 2344.0 \\
2 & Vibecheck & 2278.6 \\
3 & PyRAT & 1813.6 \\
4 & NeuralSAT & 1725.5 \\
5 & NNV & 1149.5 \\
6 & CORA & 1055.5 \\
7 & Bright & 1000.5 \\
8 & N2v & 984.5 \\
9 & SuperSimple & 665.3 \\
10 & Act & 650.3 \\
11 & nnenum & 489.3 \\
12 & SobolBox & 450.8 \\
13 & Agate & 0 \\
\bottomrule
\end{tabular}
}
\end{center}
\end{table}

\begin{figure}[h]
\centerline{\includegraphics[width=\textwidth]{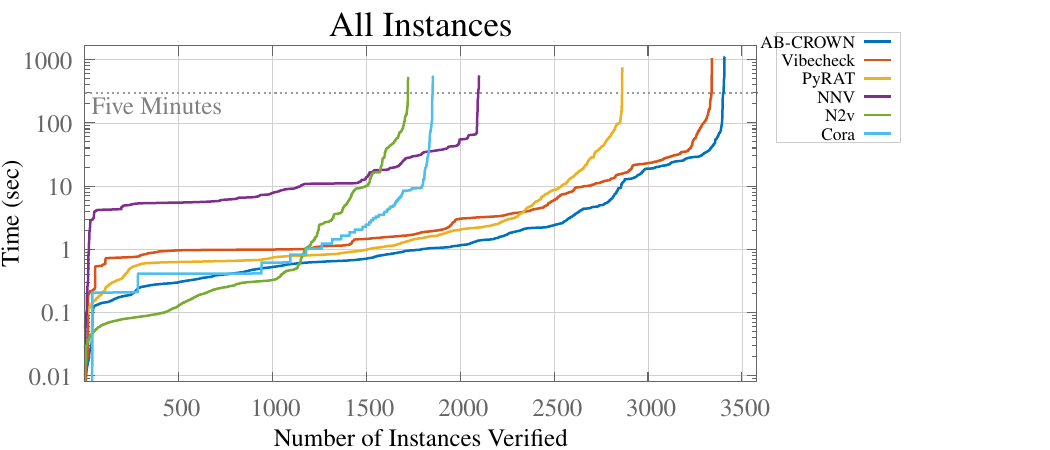}}
\caption{Cactus Plot for All Instances.}
\label{fig:quantPic}
\end{figure}

\begin{figure}[h]
\centerline{\includegraphics[width=\textwidth]{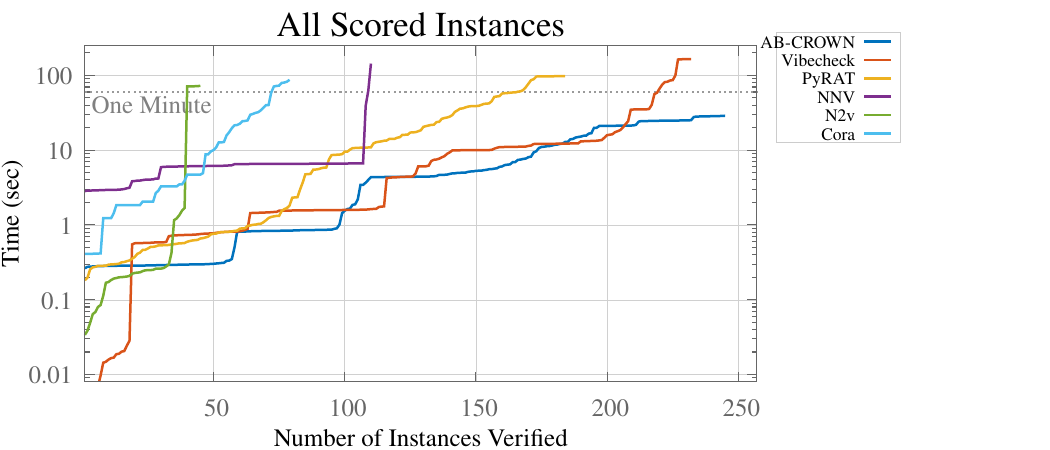}}
\caption{Cactus Plot for All Scored Instances.}
\label{fig:quantPic}
\end{figure}

\subsection{Extended Track}
All benchmarks that were voted for by at least one team and did not make it into the regular track were part of the extended track.
Every benchmark was voted for at least once, so no benchmark was unscored.


\begin{table}[h]
\begin{center}
\caption{Overall Score} \label{tab:score_extended}
{\setlength{\tabcolsep}{2pt}
\begin{tabular}[h]{@{}lll@{}}
\toprule
\textbf{\# ~} & \textbf{Tool} & \textbf{Score}\\
\midrule
1 & $\alpha$-$\beta$-CROWN & 600.0 \\
2 & Vibecheck & 574.0 \\
3 & PyRAT & 478.0 \\
4 & NNV & 271.3 \\
5 & N2v & 178.0 \\
6 & CORA & 158.0 \\
\bottomrule
\end{tabular}
}
\end{center}
\end{table}

\clearpage

\subsection{Other Stats}

This section presents other statistics related to the measurements that are interesting but did not play a direct role in scoring this year.
The results reflect the tool performance on the regular track.



\begin{table}[h]
\begin{center}
\caption{Overhead} \label{tab:overhead}
{\setlength{\tabcolsep}{2pt}
\begin{tabular}[h]{@{}llr@{}}
\toprule
\textbf{\# ~} & \textbf{Tool} & \textbf{Seconds}\\
\midrule
1 & nnenum & 0.9 \\
2 & CORA & 1.0 \\
3 & Vibecheck & 1.7 \\
4 & Agate & 2.7 \\
5 & Bright & 2.9 \\
6 & NeuralSAT & 3.2 \\
7 & SuperSimple & 3.6 \\
8 & N2v & 4.1 \\
9 & $\alpha$-$\beta$-CROWN & 5.6 \\
10 & SobolBox & 7.4 \\
11 & PyRAT & 7.5 \\
12 & NNV & 8.2 \\
13 & Act & 10.9 \\
\bottomrule
\end{tabular}
}
\end{center}
\end{table}


\begin{table}[h]
\begin{center}
\caption{Num Benchmarks Participated} \label{tab:stats0}
{\setlength{\tabcolsep}{2pt}
\begin{tabular}[h]{@{}llr@{}}
\toprule
\textbf{\# ~} & \textbf{Tool} & \textbf{Count}\\
\midrule
1 & Vibecheck & 25 \\
2 & PyRAT & 24 \\
3 & $\alpha$-$\beta$-CROWN & 24 \\
4 & NeuralSAT & 22 \\
5 & N2v & 21 \\
6 & NNV & 19 \\
7 & Agate & 18 \\
8 & SuperSimple & 17 \\
9 & nnenum & 13 \\
10 & CORA & 13 \\
11 & Act & 12 \\
12 & SobolBox & 11 \\
13 & Bright & 11 \\
\bottomrule
\end{tabular}
}
\end{center}
\end{table}


\begin{table}[h]
\begin{center}
\caption{Num Instances Verified} \label{tab:stats1}
{\setlength{\tabcolsep}{2pt}
\begin{tabular}[h]{@{}llr@{}}
\toprule
\textbf{\# ~} & \textbf{Tool} & \textbf{Count}\\
\midrule
1 & $\alpha$-$\beta$-CROWN & 3164 \\
2 & Vibecheck & 3108 \\
3 & PyRAT & 2679 \\
4 & NeuralSAT & 2637 \\
5 & Bright & 2048 \\
6 & NNV & 1991 \\
7 & CORA & 1776 \\
8 & N2v & 1677 \\
9 & Act & 1558 \\
10 & SuperSimple & 1551 \\
11 & nnenum & 1378 \\
12 & SobolBox & 559 \\
13 & Agate & 254 \\
\bottomrule
\end{tabular}
}
\end{center}
\end{table}


\begin{table}[h]
\begin{center}
\caption{Num SAT} \label{tab:stats2}
{\setlength{\tabcolsep}{2pt}
\begin{tabular}[h]{@{}llr@{}}
\toprule
\textbf{\# ~} & \textbf{Tool} & \textbf{Count}\\
\midrule
1 & $\alpha$-$\beta$-CROWN & 1182 \\
2 & Vibecheck & 1161 \\
3 & PyRAT & 1093 \\
4 & N2v & 1045 \\
5 & Bright & 1022 \\
6 & SuperSimple & 1007 \\
7 & NeuralSAT & 998 \\
8 & NNV & 963 \\
9 & CORA & 898 \\
10 & Act & 759 \\
11 & nnenum & 695 \\
12 & SobolBox & 355 \\
\bottomrule
\end{tabular}
}
\end{center}
\end{table}


\begin{table}[h]
\begin{center}
\caption{Num UNSAT} \label{tab:stats3}
{\setlength{\tabcolsep}{2pt}
\begin{tabular}[h]{@{}llr@{}}
\toprule
\textbf{\# ~} & \textbf{Tool} & \textbf{Count}\\
\midrule
1 & $\alpha$-$\beta$-CROWN & 1982 \\
2 & Vibecheck & 1947 \\
3 & NeuralSAT & 1639 \\
4 & PyRAT & 1586 \\
5 & NNV & 1028 \\
6 & Bright & 1026 \\
7 & CORA & 878 \\
8 & Act & 799 \\
9 & nnenum & 683 \\
10 & N2v & 632 \\
11 & SuperSimple & 544 \\
12 & Agate & 254 \\
13 & SobolBox & 204 \\
\bottomrule
\end{tabular}
}
\end{center}
\end{table}


\begin{table}[h]
\begin{center}
\caption{Incorrect Results (or Missing CE)} \label{tab:stats4}
{\setlength{\tabcolsep}{2pt}
\begin{tabular}[h]{@{}llr@{}}
\toprule
\textbf{\# ~} & \textbf{Tool} & \textbf{Count}\\
\midrule
1 & Agate & 955 \\
2 & NeuralSAT & 94 \\
3 & nnenum & 58 \\
4 & SuperSimple & 14 \\
5 & Act & 10 \\
6 & SobolBox & 2 \\
7 & CORA & 1 \\
8 & $\alpha$-$\beta$-CROWN & 1 \\
\bottomrule
\end{tabular}
}
\end{center}
\end{table}

\clearpage

\section{Conclusion and Ideas for Future Competitions}
\label{sec:conclusion}

This report summarizes the 7$^\text{th}$ Verification of Neural Networks Competition (VNN-COMP), held in 2026.
As per last year, the best-performing tools are GPU-enabled linear bound propagation methods using a branch-and-bound framework.
The new VNNLIB 2.0 input format was widely adopted by tool proposers, which sends a positive signal. It shows that the community is eager to extend the reach of verification to more complex properties; we hope this will encourage difficult benchmarks propositions for the next iteration.

VNN-COMP 2026 successfully implemented a range of improvement opportunities identified during the previous iteration. These included requiring witnesses of found counter-examples to disambiguate tool disagreement, increasing automatization to enable a smoother final evaluation, and making a broader range of AWS instances available to allow for a better fit with tools' requirements. Additionally, we introduced the Competition Contribution papers included in the SAIV proceedings~\cite{saiv2025} for participants to summarize their approaches.
In addition, in cases where the answer to a benchmark is known a priori, we will explore using the known correct results as the ground truth.
Further ideas for future competitions include the use of scored benchmarks specifically designed for year-on-year progress tracking, the reduction of tool tuning, a batch-processing mode, and more rigorous soundness evaluation. Finally, as research on proof generation for neural network verification grows~\cite{isac2022neural,elboher2025abstraction,duong2026generating}, a longer term goal is to require solvers to produce UNSAT proofs.

\section*{Acknowledgements}
%

The 2026 competition was supported by SafeIntelligence and MathWorks, which provided funds for computation costs.

Additionally, this research was supported in part by the Air Force Research Laboratory Information Directorate, through the Air Force Office of Scientific Research Summer Faculty Fellowship
Program, Contract Numbers FA8750-15-3-6003, FA9550-15-0001 and FA9550-20-F-0005.
This material is based upon work supported by the Air Force Office of Scientific Research under award numbers FA9550-19-1-0288, FA9550-21-1-0121, and FA9550-22-1-0019, the National Science Foundation (NSF) under grant numbers 2107035, 2220401, 2220418, 2220426, and 2238133, and the Defense Advanced Research Projects Agency (DARPA) Assured Neuro Symbolic Learning and Reasoning (ANSR)
programs through contract number and FA8750-23-C-0518.
Any opinions, findings, and conclusions or recommendations expressed in this material are those of the author(s) and do not necessarily reflect the views of the United States Air Force, DARPA, nor NSF.
This research was additionally supported by a
grant from the project AL 1185/33-1 funded
by the German Research Foundation (Deutsche Forschungsgemeinschaft, DFG).

Tool and benchmark authors listed in \Cref{sec:participants} and \Cref{sec:benchmarks} participated in the preparation and review of this report. 
Special thanks go to Volkan Şahin for his significant contributions to the development and maintenance of the evaluation system.

\clearpage
\label{sect:bib}
\bibliographystyle{plain}
\bibliography{bib/nnv, bib/nnenum, bib/peregriNN, bib/verinet, bib/oval,bib/venus,bib/MIPVerify, bib/mnbab, bib/alpha-beta-CROWN, bib/collins,bib/dnnf,bib/nvjl,bib/nn4sys,bib/Marabou,bib/RPM,bib/AVeriNN,bib/VeRAPAk,bib/general,bib/traffic-signs-recognition,bib/neuralsat, bib/pyrat, bib/vit, bib/cora, bib/never2, bib/safeNLP, bib/SobolBox, bib/relusplitter,bib/metaroom,bib/cersyve,bib/supersimple,bib/act,bib/n2v, bib/agate}


\appendix

\clearpage
\section{Benchmark Results}
In this section, we provide more fine-grained results.

\subsection{Regular Track}
\label{sec:benchmark_results_regular}

\clearpage

\begin{table}[h]
\begin{center}
\caption{Benchmark \texttt{2026-acasxu-2023}} \label{tab:cat_2026_acasxu_2023}
{\setlength{\tabcolsep}{2pt}
\begin{tabular}[h]{@{}llllllrrr@{}}
\toprule
\textbf{\# ~} & \textbf{Tool} & \textbf{Verified} & \textbf{Falsified} & \textbf{Fastest} & \textbf{Penalty} & \textbf{Points} & \textbf{Score} & \textbf{Solved}\\
\midrule
1 & Vibecheck & 139 & 47 & 0 & 0 & 1860 & 100.0 & 100.0\% \\
2 & PyRAT & 139 & 47 & 0 & 0 & 1860 & 100.0 & 100.0\% \\
3 & $\alpha$-$\beta$-CROWN & 139 & 47 & 0 & 0 & 1860 & 100.0 & 100.0\% \\
4 & CORA & 137 & 47 & 0 & 0 & 1840 & 98.9 & 98.9\% \\
5 & NeuralSAT & 139 & 44 & 0 & 0 & 1830 & 98.4 & 98.4\% \\
6 & Bright & 133 & 46 & 0 & 0 & 1790 & 96.2 & 96.2\% \\
7 & NNV & 130 & 47 & 0 & 0 & 1770 & 95.2 & 95.2\% \\
8 & nnenum & 139 & 44 & 0 & 1 & 1680 & 90.3 & 98.4\% \\
9 & SobolBox & 115 & 42 & 0 & 0 & 1570 & 84.4 & 84.4\% \\
10 & Act & 132 & 46 & 0 & 3 & 1330 & 71.5 & 95.7\% \\
11 & N2v & 80 & 38 & 0 & 0 & 1180 & 63.4 & 63.4\% \\
12 & SuperSimple & 54 & 38 & 0 & 0 & 920 & 49.5 & 49.5\% \\
13 & Agate & 85 & 0 & 0 & 46 & -6050 & 0 & 45.7\% \\
\bottomrule
\end{tabular}
}
\end{center}
\end{table}

\begin{figure}[h]
\centerline{\includegraphics[width=\textwidth]{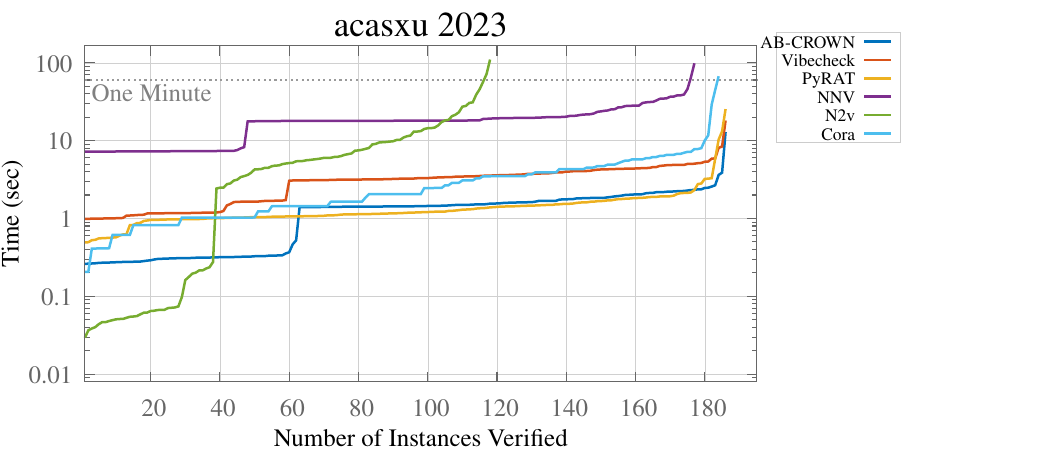}}
\caption{Cactus Plot for acasxu 2023.}
\label{fig:quantPic}
\end{figure}

\clearpage

\begin{table}[h]
\begin{center}
\caption{Benchmark \texttt{2026-cersyve}} \label{tab:cat_2026_cersyve}
{\setlength{\tabcolsep}{2pt}
\begin{tabular}[h]{@{}llllllrrr@{}}
\toprule
\textbf{\# ~} & \textbf{Tool} & \textbf{Verified} & \textbf{Falsified} & \textbf{Fastest} & \textbf{Penalty} & \textbf{Points} & \textbf{Score} & \textbf{Solved}\\
\midrule
1 & Vibecheck & 6 & 6 & 0 & 0 & 120 & 100.0 & 100.0\% \\
2 & NeuralSAT & 6 & 6 & 0 & 0 & 120 & 100.0 & 100.0\% \\
3 & $\alpha$-$\beta$-CROWN & 6 & 6 & 0 & 0 & 120 & 100.0 & 100.0\% \\
4 & PyRAT & 5 & 6 & 0 & 0 & 110 & 91.7 & 91.7\% \\
5 & CORA & 5 & 6 & 0 & 0 & 110 & 91.7 & 91.7\% \\
6 & Act & 3 & 6 & 0 & 0 & 90 & 75.0 & 75.0\% \\
7 & SuperSimple & 0 & 6 & 0 & 0 & 60 & 50.0 & 50.0\% \\
8 & SobolBox & 0 & 6 & 0 & 0 & 60 & 50.0 & 50.0\% \\
9 & N2v & 0 & 6 & 0 & 0 & 60 & 50.0 & 50.0\% \\
10 & NNV & 1 & 4 & 0 & 0 & 50 & 41.7 & 41.7\% \\
11 & nnenum & 0 & 3 & 0 & 1 & -120 & 0 & 25.0\% \\
12 & Agate & 0 & 0 & 0 & 6 & -900 & 0 & 0.0\% \\
\bottomrule
\end{tabular}
}
\end{center}
\end{table}

\begin{figure}[h]
\centerline{\includegraphics[width=\textwidth]{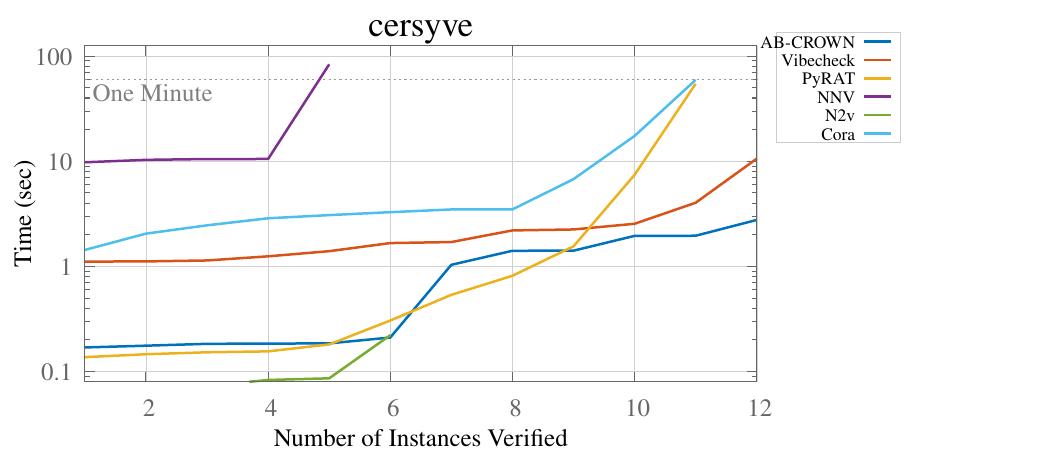}}
\caption{Cactus Plot for cersyve.}
\label{fig:quantPic}
\end{figure}

\clearpage

\begin{table}[h]
\begin{center}
\caption{Benchmark \texttt{2026-cgan2026}} \label{tab:cat_2026_cgan2026}
{\setlength{\tabcolsep}{2pt}
\begin{tabular}[h]{@{}llllllrrr@{}}
\toprule
\textbf{\# ~} & \textbf{Tool} & \textbf{Verified} & \textbf{Falsified} & \textbf{Fastest} & \textbf{Penalty} & \textbf{Points} & \textbf{Score} & \textbf{Solved}\\
\midrule
1 & Vibecheck & 8 & 12 & 0 & 0 & 200 & 100.0 & 95.2\% \\
2 & $\alpha$-$\beta$-CROWN & 8 & 12 & 0 & 0 & 200 & 100.0 & 95.2\% \\
3 & PyRAT & 8 & 11 & 0 & 0 & 190 & 95.0 & 90.5\% \\
4 & NeuralSAT & 4 & 12 & 0 & 0 & 160 & 80.0 & 76.2\% \\
5 & NNV & 0 & 11 & 0 & 0 & 110 & 55.0 & 52.4\% \\
6 & CORA & 0 & 7 & 0 & 0 & 70 & 35.0 & 33.3\% \\
7 & N2v & 0 & 6 & 0 & 0 & 60 & 30.0 & 28.6\% \\
8 & nnenum & 0 & 1 & 0 & 0 & 10 & 5.0 & 4.8\% \\
9 & SobolBox & 9 & 10 & 0 & 2 & -110 & 0 & 90.5\% \\
10 & Agate & 0 & 0 & 0 & 18 & -2700 & 0 & 0.0\% \\
\bottomrule
\end{tabular}
}
\end{center}
\end{table}

\begin{figure}[h]
\centerline{\includegraphics[width=\textwidth]{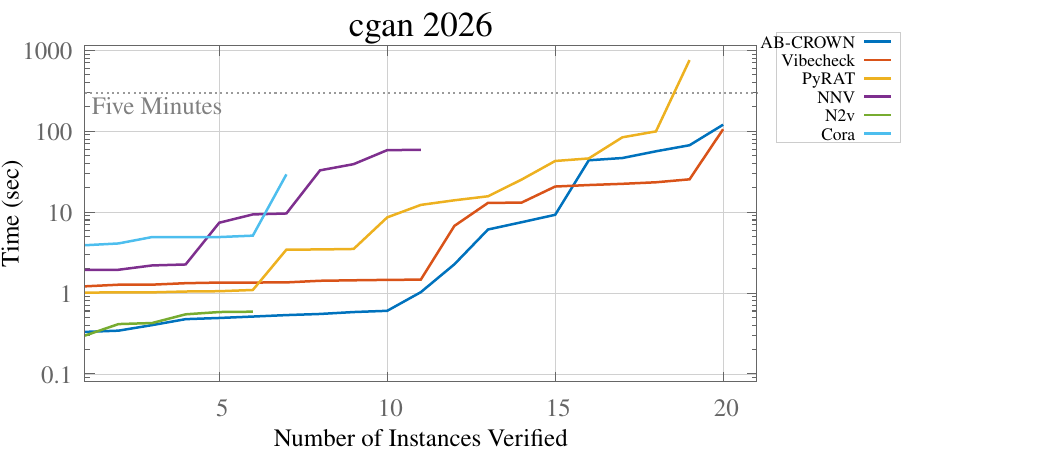}}
\caption{Cactus Plot for cgan 2026.}
\label{fig:quantPic}
\end{figure}

\clearpage

\begin{table}[h]
\begin{center}
\caption{Benchmark \texttt{2026-challenging-certified-training-2026}} \label{tab:cat_2026_challenging_certified_training_2026}
{\setlength{\tabcolsep}{2pt}
\begin{tabular}[h]{@{}llllllrrr@{}}
\toprule
\textbf{\# ~} & \textbf{Tool} & \textbf{Verified} & \textbf{Falsified} & \textbf{Fastest} & \textbf{Penalty} & \textbf{Points} & \textbf{Score} & \textbf{Solved}\\
\midrule
1 & $\alpha$-$\beta$-CROWN & 27 & 11 & 0 & 0 & 380 & 100.0 & 63.3\% \\
2 & Vibecheck & 22 & 11 & 0 & 0 & 330 & 86.8 & 55.0\% \\
3 & NeuralSAT & 16 & 4 & 0 & 0 & 200 & 52.6 & 33.3\% \\
4 & PyRAT & 6 & 10 & 0 & 0 & 160 & 42.1 & 26.7\% \\
5 & SuperSimple & 0 & 11 & 0 & 0 & 110 & 28.9 & 18.3\% \\
6 & N2v & 0 & 4 & 0 & 0 & 40 & 10.5 & 6.7\% \\
7 & NNV & 0 & 3 & 0 & 0 & 30 & 7.9 & 5.0\% \\
8 & Agate & 0 & 0 & 0 & 3 & -450 & 0 & 0.0\% \\
\bottomrule
\end{tabular}
}
\end{center}
\end{table}

\begin{figure}[h]
\centerline{\includegraphics[width=\textwidth]{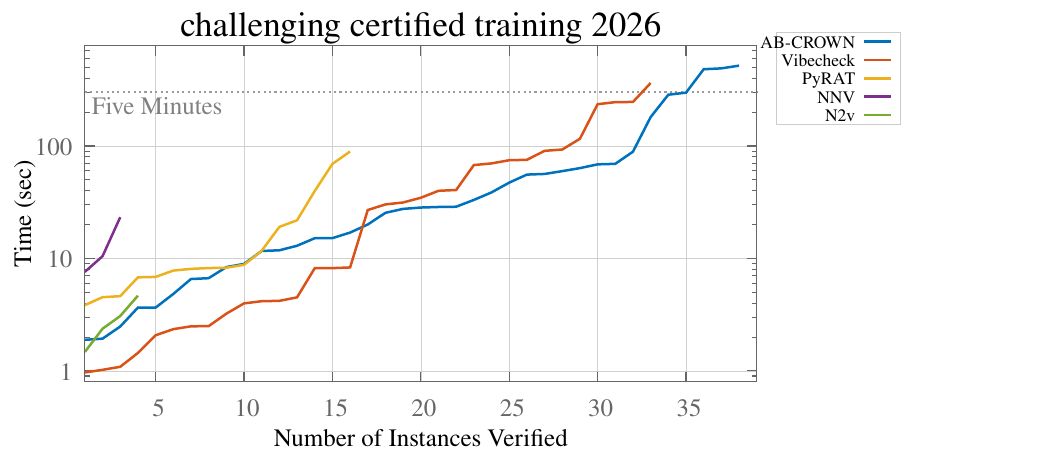}}
\caption{Cactus Plot for challenging certified training 2026.}
\label{fig:quantPic}
\end{figure}

\clearpage

\begin{table}[h]
\begin{center}
\caption{Benchmark \texttt{2026-cifar100-2024}} \label{tab:cat_2026_cifar100_2024}
{\setlength{\tabcolsep}{2pt}
\begin{tabular}[h]{@{}llllllrrr@{}}
\toprule
\textbf{\# ~} & \textbf{Tool} & \textbf{Verified} & \textbf{Falsified} & \textbf{Fastest} & \textbf{Penalty} & \textbf{Points} & \textbf{Score} & \textbf{Solved}\\
\midrule
1 & $\alpha$-$\beta$-CROWN & 127 & 33 & 0 & 0 & 1600 & 100.0 & 80.0\% \\
2 & Vibecheck & 125 & 33 & 0 & 0 & 1580 & 98.8 & 79.0\% \\
3 & NeuralSAT & 104 & 33 & 0 & 0 & 1370 & 85.6 & 68.5\% \\
4 & Bright & 89 & 33 & 0 & 0 & 1220 & 76.2 & 61.0\% \\
5 & PyRAT & 88 & 31 & 0 & 0 & 1190 & 74.4 & 59.5\% \\
6 & NNV & 93 & 12 & 0 & 0 & 1050 & 65.6 & 52.5\% \\
7 & Act & 35 & 14 & 0 & 0 & 490 & 30.6 & 24.5\% \\
8 & SuperSimple & 0 & 33 & 0 & 0 & 330 & 20.6 & 16.5\% \\
9 & N2v & 0 & 32 & 0 & 0 & 320 & 20.0 & 16.0\% \\
\bottomrule
\end{tabular}
}
\end{center}
\end{table}

\begin{figure}[h]
\centerline{\includegraphics[width=\textwidth]{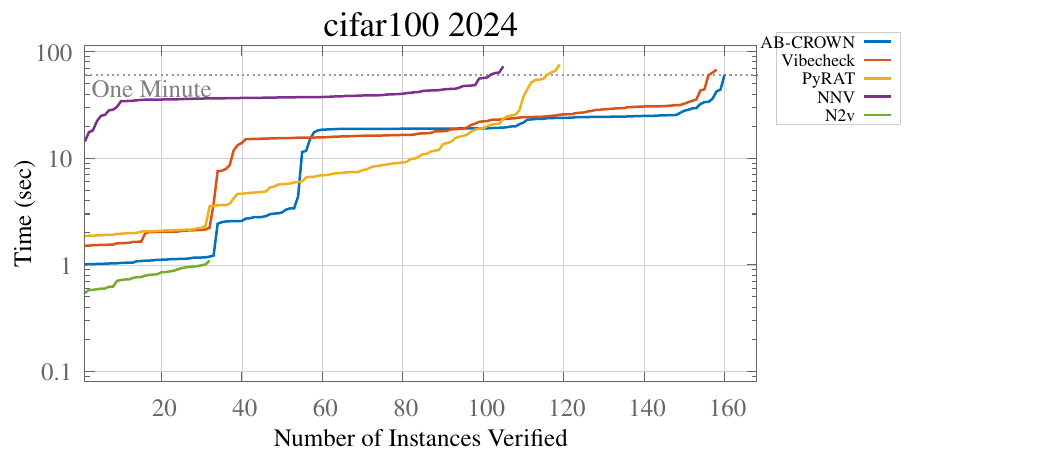}}
\caption{Cactus Plot for cifar100 2024.}
\label{fig:quantPic}
\end{figure}

\clearpage

\begin{table}[h]
\begin{center}
\caption{Benchmark \texttt{2026-collins-rul-cnn-2022}} \label{tab:cat_2026_collins_rul_cnn_2022}
{\setlength{\tabcolsep}{2pt}
\begin{tabular}[h]{@{}llllllrrr@{}}
\toprule
\textbf{\# ~} & \textbf{Tool} & \textbf{Verified} & \textbf{Falsified} & \textbf{Fastest} & \textbf{Penalty} & \textbf{Points} & \textbf{Score} & \textbf{Solved}\\
\midrule
1 & nnenum & 45 & 17 & 0 & 0 & 620 & 100.0 & 100.0\% \\
2 & Vibecheck & 45 & 17 & 0 & 0 & 620 & 100.0 & 100.0\% \\
3 & PyRAT & 45 & 17 & 0 & 0 & 620 & 100.0 & 100.0\% \\
4 & NeuralSAT & 45 & 17 & 0 & 0 & 620 & 100.0 & 100.0\% \\
5 & NNV & 45 & 17 & 0 & 0 & 620 & 100.0 & 100.0\% \\
6 & N2v & 45 & 17 & 0 & 0 & 620 & 100.0 & 100.0\% \\
7 & $\alpha$-$\beta$-CROWN & 45 & 17 & 0 & 0 & 620 & 100.0 & 100.0\% \\
8 & CORA & 45 & 0 & 0 & 0 & 450 & 72.6 & 72.6\% \\
9 & SobolBox & 1 & 17 & 0 & 0 & 180 & 29.0 & 29.0\% \\
10 & Agate & 0 & 0 & 0 & 17 & -2550 & 0 & 0.0\% \\
\bottomrule
\end{tabular}
}
\end{center}
\end{table}

\begin{figure}[h]
\centerline{\includegraphics[width=\textwidth]{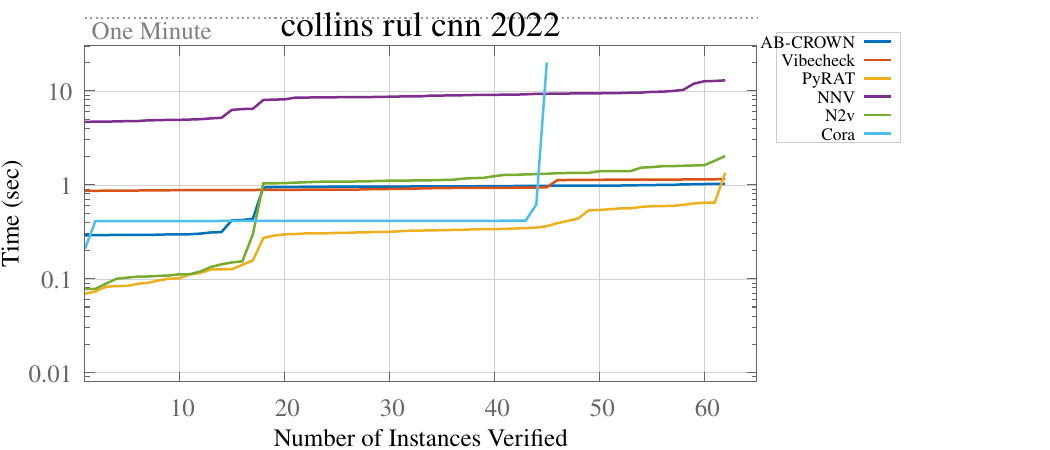}}
\caption{Cactus Plot for collins rul cnn 2022.}
\label{fig:quantPic}
\end{figure}

\clearpage

\begin{table}[h]
\begin{center}
\caption{Benchmark \texttt{2026-cora-2024}} \label{tab:cat_2026_cora_2024}
{\setlength{\tabcolsep}{2pt}
\begin{tabular}[h]{@{}llllllrrr@{}}
\toprule
\textbf{\# ~} & \textbf{Tool} & \textbf{Verified} & \textbf{Falsified} & \textbf{Fastest} & \textbf{Penalty} & \textbf{Points} & \textbf{Score} & \textbf{Solved}\\
\midrule
1 & Vibecheck & 25 & 127 & 0 & 0 & 1520 & 100.0 & 84.4\% \\
2 & $\alpha$-$\beta$-CROWN & 25 & 127 & 0 & 0 & 1520 & 100.0 & 84.4\% \\
3 & NeuralSAT & 24 & 127 & 0 & 0 & 1510 & 99.3 & 83.9\% \\
4 & PyRAT & 24 & 126 & 0 & 0 & 1500 & 98.7 & 83.3\% \\
5 & Bright & 23 & 126 & 0 & 0 & 1490 & 98.0 & 82.8\% \\
6 & SuperSimple & 21 & 126 & 0 & 0 & 1470 & 96.7 & 81.7\% \\
7 & N2v & 19 & 126 & 0 & 0 & 1450 & 95.4 & 80.6\% \\
8 & CORA & 21 & 119 & 0 & 0 & 1400 & 92.1 & 77.8\% \\
9 & NNV & 0 & 104 & 0 & 0 & 1040 & 68.4 & 57.8\% \\
10 & nnenum & 21 & 1 & 0 & 0 & 220 & 14.5 & 12.2\% \\
11 & Act & 2 & 4 & 0 & 0 & 60 & 3.9 & 3.3\% \\
12 & Agate & 0 & 0 & 0 & 89 & -13350 & 0 & 0.0\% \\
\bottomrule
\end{tabular}
}
\end{center}
\end{table}

\begin{figure}[h]
\centerline{\includegraphics[width=\textwidth]{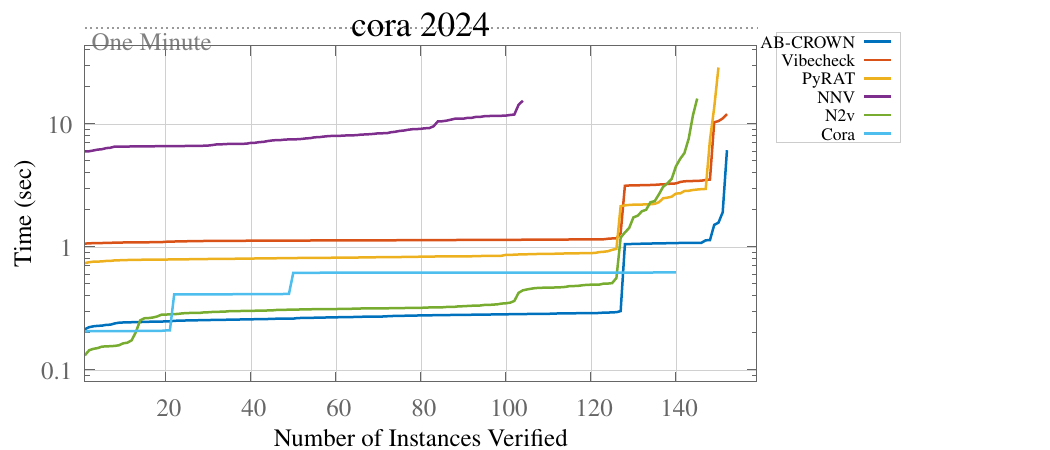}}
\caption{Cactus Plot for cora 2024.}
\label{fig:quantPic}
\end{figure}

\clearpage

\begin{table}[h]
\begin{center}
\caption{Benchmark \texttt{2026-dist-shift-2023}} \label{tab:cat_2026_dist_shift_2023}
{\setlength{\tabcolsep}{2pt}
\begin{tabular}[h]{@{}llllllrrr@{}}
\toprule
\textbf{\# ~} & \textbf{Tool} & \textbf{Verified} & \textbf{Falsified} & \textbf{Fastest} & \textbf{Penalty} & \textbf{Points} & \textbf{Score} & \textbf{Solved}\\
\midrule
1 & Vibecheck & 64 & 8 & 0 & 0 & 720 & 100.0 & 100.0\% \\
2 & PyRAT & 64 & 8 & 0 & 0 & 720 & 100.0 & 100.0\% \\
3 & NeuralSAT & 64 & 8 & 0 & 0 & 720 & 100.0 & 100.0\% \\
4 & N2v & 64 & 8 & 0 & 0 & 720 & 100.0 & 100.0\% \\
5 & CORA & 64 & 8 & 0 & 0 & 720 & 100.0 & 100.0\% \\
6 & $\alpha$-$\beta$-CROWN & 64 & 8 & 0 & 0 & 720 & 100.0 & 100.0\% \\
7 & SuperSimple & 53 & 8 & 0 & 0 & 610 & 84.7 & 84.7\% \\
8 & NNV & 49 & 8 & 0 & 0 & 570 & 79.2 & 79.2\% \\
9 & SobolBox & 0 & 8 & 0 & 0 & 80 & 11.1 & 11.1\% \\
10 & Act & 64 & 1 & 0 & 7 & -400 & 0 & 90.3\% \\
11 & Agate & 0 & 0 & 0 & 8 & -1200 & 0 & 0.0\% \\
\bottomrule
\end{tabular}
}
\end{center}
\end{table}

\begin{figure}[h]
\centerline{\includegraphics[width=\textwidth]{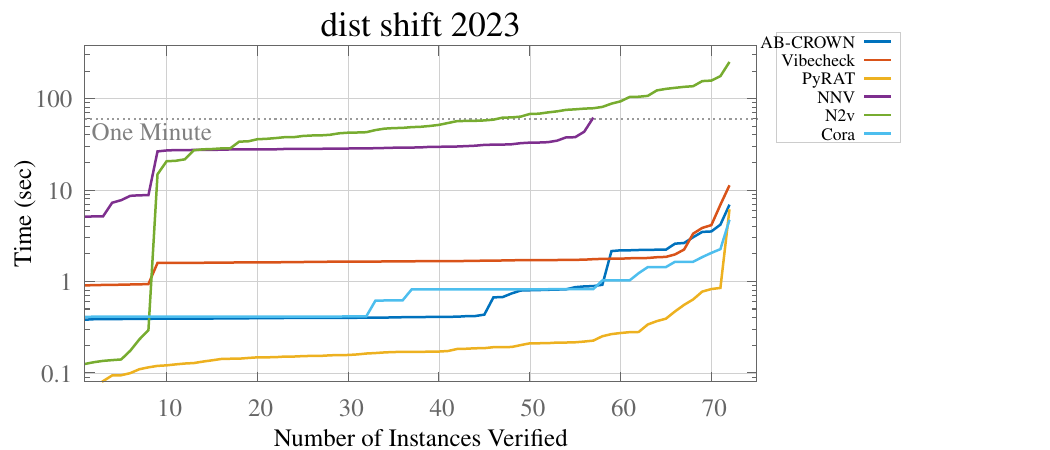}}
\caption{Cactus Plot for dist shift 2023.}
\label{fig:quantPic}
\end{figure}

\clearpage

\begin{table}[h]
\begin{center}
\caption{Benchmark \texttt{2026-linearizenn-2024}} \label{tab:cat_2026_linearizenn_2024}
{\setlength{\tabcolsep}{2pt}
\begin{tabular}[h]{@{}llllllrrr@{}}
\toprule
\textbf{\# ~} & \textbf{Tool} & \textbf{Verified} & \textbf{Falsified} & \textbf{Fastest} & \textbf{Penalty} & \textbf{Points} & \textbf{Score} & \textbf{Solved}\\
\midrule
1 & nnenum & 59 & 1 & 0 & 0 & 600 & 100.0 & 100.0\% \\
2 & Vibecheck & 59 & 1 & 0 & 0 & 600 & 100.0 & 100.0\% \\
3 & SobolBox & 59 & 1 & 0 & 0 & 600 & 100.0 & 100.0\% \\
4 & PyRAT & 59 & 1 & 0 & 0 & 600 & 100.0 & 100.0\% \\
5 & NeuralSAT & 59 & 1 & 0 & 0 & 600 & 100.0 & 100.0\% \\
6 & NNV & 59 & 1 & 0 & 0 & 600 & 100.0 & 100.0\% \\
7 & CORA & 59 & 1 & 0 & 0 & 600 & 100.0 & 100.0\% \\
8 & Bright & 59 & 1 & 0 & 0 & 600 & 100.0 & 100.0\% \\
9 & $\alpha$-$\beta$-CROWN & 59 & 1 & 0 & 0 & 600 & 100.0 & 100.0\% \\
10 & N2v & 53 & 1 & 0 & 0 & 540 & 90.0 & 90.0\% \\
11 & SuperSimple & 14 & 1 & 0 & 0 & 150 & 25.0 & 25.0\% \\
12 & Agate & 0 & 0 & 0 & 1 & -150 & 0 & 0.0\% \\
\bottomrule
\end{tabular}
}
\end{center}
\end{table}

\begin{figure}[h]
\centerline{\includegraphics[width=\textwidth]{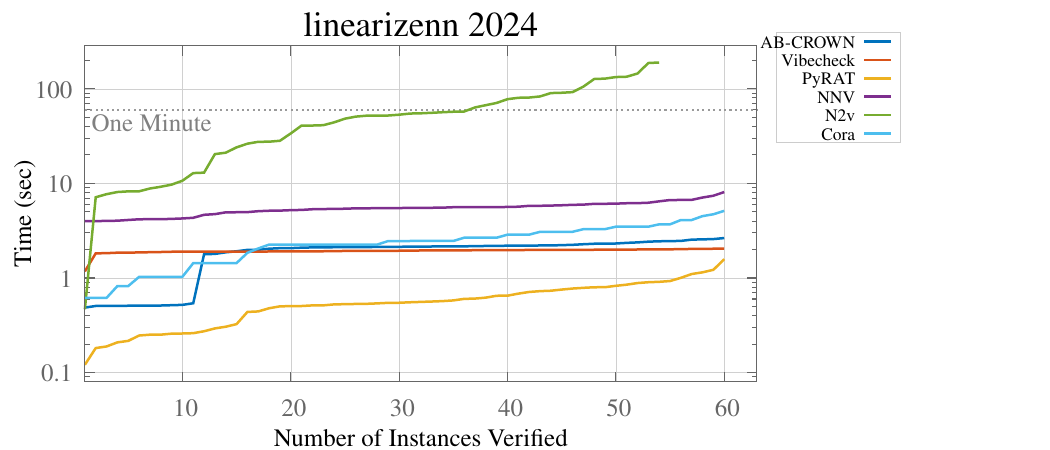}}
\caption{Cactus Plot for linearizenn 2024.}
\label{fig:quantPic}
\end{figure}

\clearpage

\begin{table}[h]
\begin{center}
\caption{Benchmark \texttt{2026-lsnc-relu}} \label{tab:cat_2026_lsnc_relu}
{\setlength{\tabcolsep}{2pt}
\begin{tabular}[h]{@{}llllllrrr@{}}
\toprule
\textbf{\# ~} & \textbf{Tool} & \textbf{Verified} & \textbf{Falsified} & \textbf{Fastest} & \textbf{Penalty} & \textbf{Points} & \textbf{Score} & \textbf{Solved}\\
\midrule
1 & Vibecheck & 68 & 12 & 0 & 0 & 800 & 100.0 & 100.0\% \\
2 & $\alpha$-$\beta$-CROWN & 68 & 12 & 0 & 0 & 800 & 100.0 & 100.0\% \\
3 & NNV & 0 & 9 & 0 & 0 & 90 & 11.2 & 11.2\% \\
4 & PyRAT & 0 & 8 & 0 & 0 & 80 & 10.0 & 10.0\% \\
5 & Agate & 0 & 0 & 0 & 8 & -1200 & 0 & 0.0\% \\
\bottomrule
\end{tabular}
}
\end{center}
\end{table}

\begin{figure}[h]
\centerline{\includegraphics[width=\textwidth]{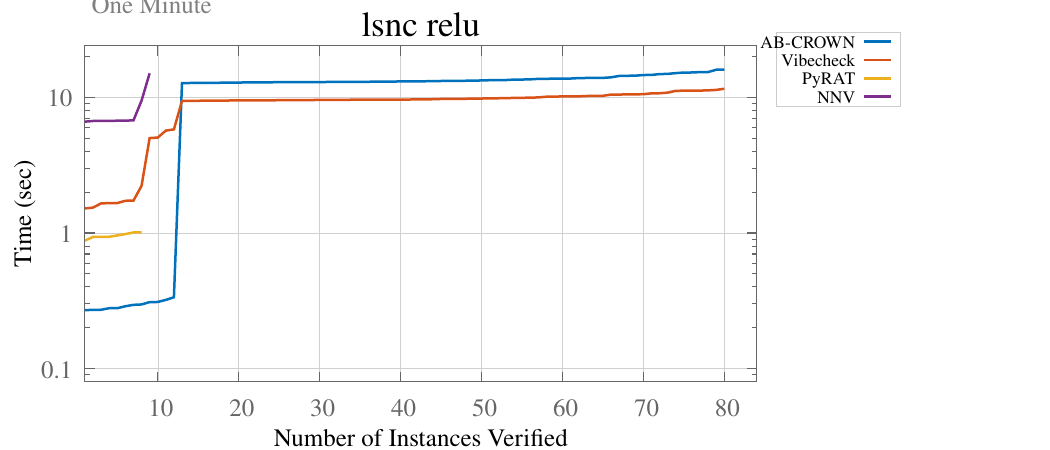}}
\caption{Cactus Plot for lsnc relu.}
\label{fig:quantPic}
\end{figure}

\clearpage

\begin{table}[h]
\begin{center}
\caption{Benchmark \texttt{2026-malbeware}} \label{tab:cat_2026_malbeware}
{\setlength{\tabcolsep}{2pt}
\begin{tabular}[h]{@{}llllllrrr@{}}
\toprule
\textbf{\# ~} & \textbf{Tool} & \textbf{Verified} & \textbf{Falsified} & \textbf{Fastest} & \textbf{Penalty} & \textbf{Points} & \textbf{Score} & \textbf{Solved}\\
\midrule
1 & $\alpha$-$\beta$-CROWN & 121 & 29 & 0 & 0 & 1500 & 100.0 & 100.0\% \\
2 & SuperSimple & 120 & 29 & 0 & 0 & 1490 & 99.3 & 99.3\% \\
3 & NeuralSAT & 120 & 29 & 0 & 0 & 1490 & 99.3 & 99.3\% \\
4 & Bright & 120 & 29 & 0 & 0 & 1490 & 99.3 & 99.3\% \\
5 & Vibecheck & 121 & 27 & 0 & 0 & 1480 & 98.7 & 98.7\% \\
6 & Act & 120 & 28 & 0 & 0 & 1480 & 98.7 & 98.7\% \\
7 & PyRAT & 118 & 28 & 0 & 0 & 1460 & 97.3 & 97.3\% \\
8 & NNV & 119 & 17 & 0 & 0 & 1360 & 90.7 & 90.7\% \\
9 & CORA & 116 & 10 & 0 & 0 & 1260 & 84.0 & 84.0\% \\
10 & N2v & 40 & 20 & 0 & 0 & 600 & 40.0 & 40.0\% \\
11 & nnenum & 79 & 8 & 0 & 2 & 570 & 38.0 & 58.0\% \\
12 & Agate & 0 & 0 & 0 & 6 & -900 & 0 & 0.0\% \\
\bottomrule
\end{tabular}
}
\end{center}
\end{table}

\begin{figure}[h]
\centerline{\includegraphics[width=\textwidth]{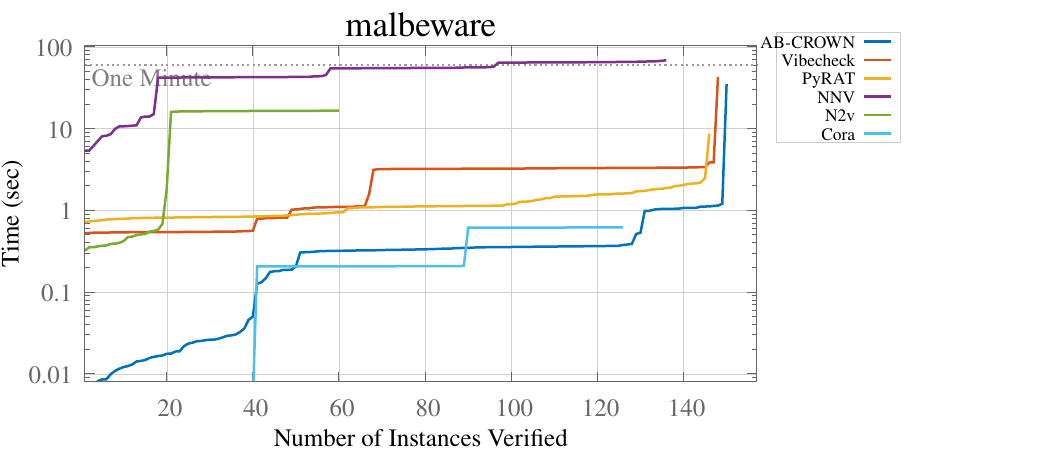}}
\caption{Cactus Plot for malbeware.}
\label{fig:quantPic}
\end{figure}

\clearpage

\begin{table}[h]
\begin{center}
\caption{Benchmark \texttt{2026-metaroom-2023}} \label{tab:cat_2026_metaroom_2023}
{\setlength{\tabcolsep}{2pt}
\begin{tabular}[h]{@{}llllllrrr@{}}
\toprule
\textbf{\# ~} & \textbf{Tool} & \textbf{Verified} & \textbf{Falsified} & \textbf{Fastest} & \textbf{Penalty} & \textbf{Points} & \textbf{Score} & \textbf{Solved}\\
\midrule
1 & Vibecheck & 93 & 6 & 0 & 0 & 990 & 100.0 & 99.0\% \\
2 & Bright & 93 & 6 & 0 & 0 & 990 & 100.0 & 99.0\% \\
3 & $\alpha$-$\beta$-CROWN & 93 & 6 & 0 & 0 & 990 & 100.0 & 99.0\% \\
4 & PyRAT & 93 & 5 & 0 & 0 & 980 & 99.0 & 98.0\% \\
5 & NeuralSAT & 92 & 6 & 0 & 0 & 980 & 99.0 & 98.0\% \\
6 & NNV & 92 & 4 & 0 & 0 & 960 & 97.0 & 96.0\% \\
7 & N2v & 91 & 5 & 0 & 0 & 960 & 97.0 & 96.0\% \\
8 & CORA & 90 & 0 & 0 & 0 & 900 & 90.9 & 90.0\% \\
9 & nnenum & 50 & 2 & 0 & 0 & 520 & 52.5 & 52.0\% \\
10 & SobolBox & 0 & 4 & 0 & 0 & 40 & 4.0 & 4.0\% \\
11 & Agate & 0 & 0 & 0 & 4 & -600 & 0 & 0.0\% \\
\bottomrule
\end{tabular}
}
\end{center}
\end{table}

\begin{figure}[h]
\centerline{\includegraphics[width=\textwidth]{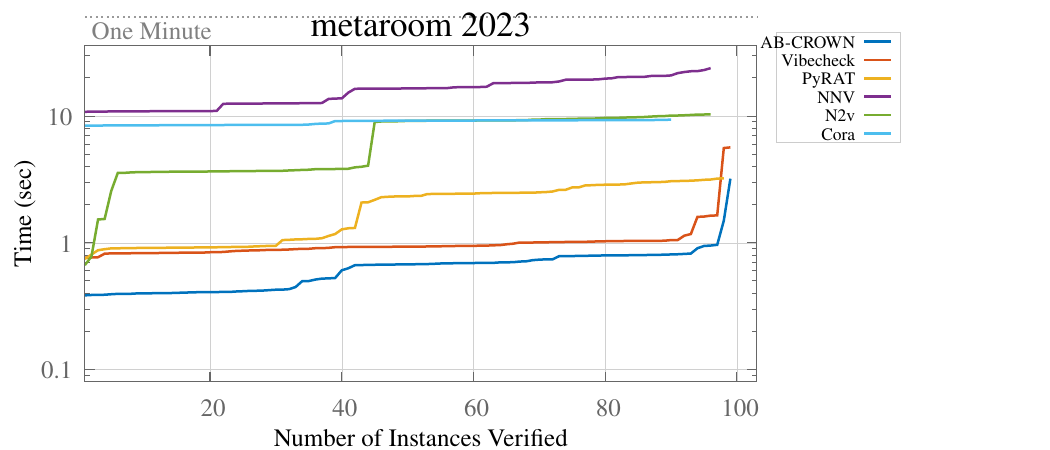}}
\caption{Cactus Plot for metaroom 2023.}
\label{fig:quantPic}
\end{figure}

\clearpage

\begin{table}[h]
\begin{center}
\caption{Benchmark \texttt{2026-ml4acopf-2024}} \label{tab:cat_2026_ml4acopf_2024}
{\setlength{\tabcolsep}{2pt}
\begin{tabular}[h]{@{}llllllrrr@{}}
\toprule
\textbf{\# ~} & \textbf{Tool} & \textbf{Verified} & \textbf{Falsified} & \textbf{Fastest} & \textbf{Penalty} & \textbf{Points} & \textbf{Score} & \textbf{Solved}\\
\midrule
1 & $\alpha$-$\beta$-CROWN & 62 & 6 & 0 & 0 & 680 & 100.0 & 98.6\% \\
2 & Vibecheck & 60 & 6 & 0 & 0 & 660 & 97.1 & 95.7\% \\
3 & NeuralSAT & 45 & 3 & 0 & 0 & 480 & 70.6 & 69.6\% \\
4 & PyRAT & 34 & 6 & 0 & 0 & 400 & 58.8 & 58.0\% \\
5 & NNV & 16 & 3 & 0 & 0 & 190 & 27.9 & 27.5\% \\
6 & SuperSimple & 17 & 0 & 0 & 0 & 170 & 25.0 & 24.6\% \\
7 & N2v & 13 & 3 & 0 & 0 & 160 & 23.5 & 23.2\% \\
8 & SobolBox & 0 & 3 & 0 & 0 & 30 & 4.4 & 4.3\% \\
9 & Agate & 0 & 0 & 0 & 15 & -2250 & 0 & 0.0\% \\
\bottomrule
\end{tabular}
}
\end{center}
\end{table}

\begin{figure}[h]
\centerline{\includegraphics[width=\textwidth]{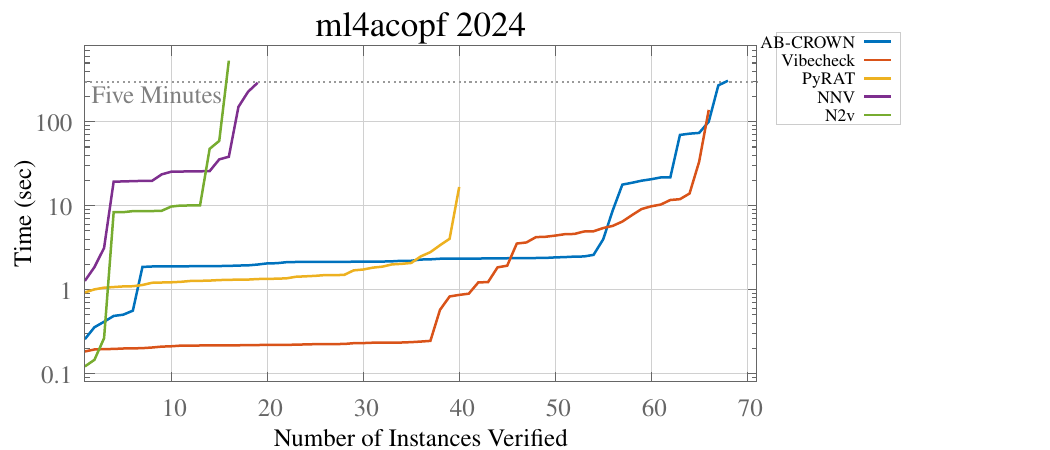}}
\caption{Cactus Plot for ml4acopf 2024.}
\label{fig:quantPic}
\end{figure}

\clearpage

\begin{table}[h]
\begin{center}
\caption{Benchmark \texttt{2026-nn4sys}} \label{tab:cat_2026_nn4sys}
{\setlength{\tabcolsep}{2pt}
\begin{tabular}[h]{@{}llllllrrr@{}}
\toprule
\textbf{\# ~} & \textbf{Tool} & \textbf{Verified} & \textbf{Falsified} & \textbf{Fastest} & \textbf{Penalty} & \textbf{Points} & \textbf{Score} & \textbf{Solved}\\
\midrule
1 & $\alpha$-$\beta$-CROWN & 194 & 0 & 0 & 0 & 1940 & 100.0 & 100.0\% \\
2 & Vibecheck & 184 & 0 & 0 & 0 & 1840 & 94.8 & 94.8\% \\
3 & PyRAT & 163 & 0 & 0 & 0 & 1630 & 84.0 & 84.0\% \\
4 & NeuralSAT & 121 & 0 & 0 & 0 & 1210 & 62.4 & 62.4\% \\
5 & N2v & 37 & 0 & 0 & 0 & 370 & 19.1 & 19.1\% \\
6 & NNV & 31 & 0 & 0 & 0 & 310 & 16.0 & 16.0\% \\
7 & SuperSimple & 26 & 0 & 0 & 0 & 260 & 13.4 & 13.4\% \\
8 & nnenum & 22 & 0 & 0 & 0 & 220 & 11.3 & 11.3\% \\
9 & Agate & 0 & 0 & 0 & 1 & -150 & 0 & 0.0\% \\
\bottomrule
\end{tabular}
}
\end{center}
\end{table}

\begin{figure}[h]
\centerline{\includegraphics[width=\textwidth]{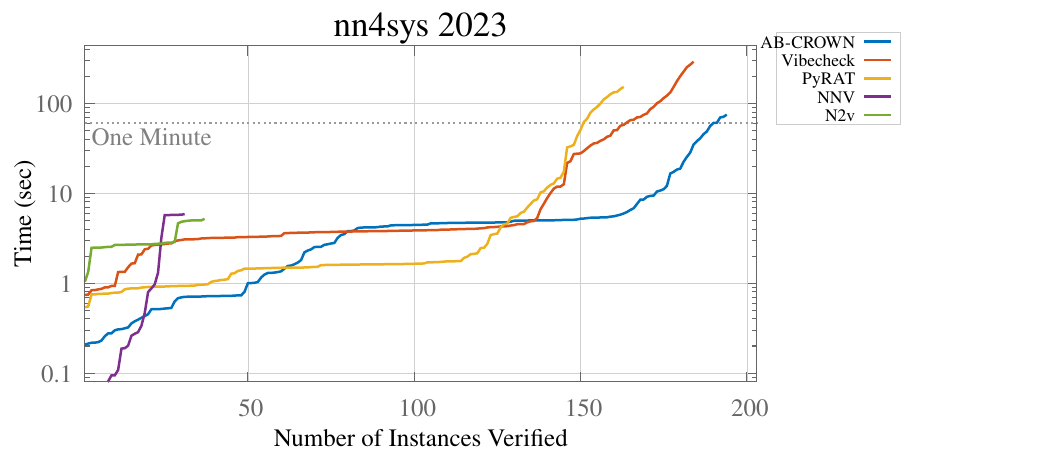}}
\caption{Cactus Plot for nn4sys 2023.}
\label{fig:quantPic}
\end{figure}

\clearpage

\begin{table}[h]
\begin{center}
\caption{Benchmark \texttt{2026-relusplitter-2026}} \label{tab:cat_2026_relusplitter_2026}
{\setlength{\tabcolsep}{2pt}
\begin{tabular}[h]{@{}llllllrrr@{}}
\toprule
\textbf{\# ~} & \textbf{Tool} & \textbf{Verified} & \textbf{Falsified} & \textbf{Fastest} & \textbf{Penalty} & \textbf{Points} & \textbf{Score} & \textbf{Solved}\\
\midrule
1 & Vibecheck & 96 & 24 & 0 & 0 & 1200 & 100.0 & 100.0\% \\
2 & $\alpha$-$\beta$-CROWN & 96 & 24 & 0 & 0 & 1200 & 100.0 & 100.0\% \\
3 & NeuralSAT & 84 & 24 & 0 & 0 & 1080 & 90.0 & 90.0\% \\
4 & Act & 90 & 18 & 0 & 0 & 1080 & 90.0 & 90.0\% \\
5 & PyRAT & 70 & 24 & 0 & 0 & 940 & 78.3 & 78.3\% \\
6 & Bright & 58 & 24 & 0 & 0 & 820 & 68.3 & 68.3\% \\
7 & NNV & 51 & 6 & 0 & 0 & 570 & 47.5 & 47.5\% \\
8 & SuperSimple & 28 & 24 & 0 & 0 & 520 & 43.3 & 43.3\% \\
9 & N2v & 9 & 24 & 0 & 0 & 330 & 27.5 & 27.5\% \\
10 & Agate & 0 & 0 & 0 & 18 & -2700 & 0 & 0.0\% \\
\bottomrule
\end{tabular}
}
\end{center}
\end{table}

\begin{figure}[h]
\centerline{\includegraphics[width=\textwidth]{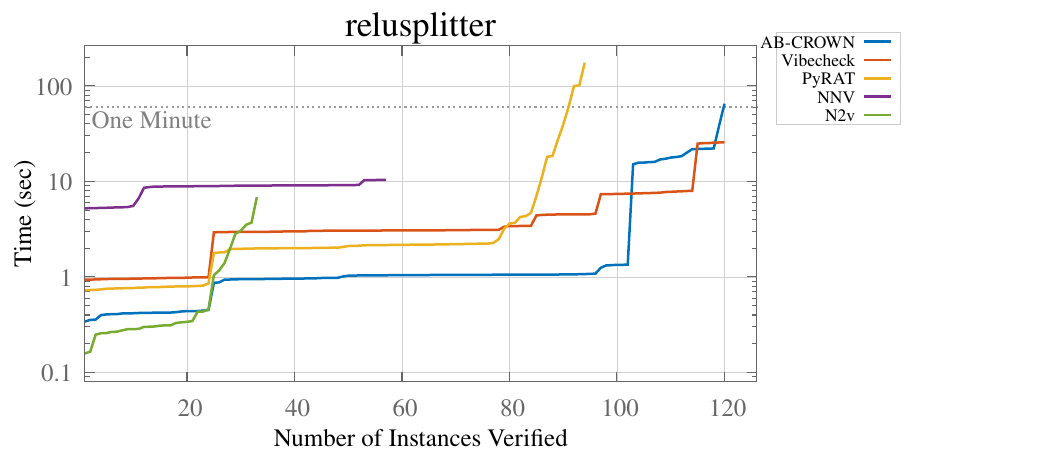}}
\caption{Cactus Plot for relusplitter.}
\label{fig:quantPic}
\end{figure}

\clearpage

\begin{table}[h]
\begin{center}
\caption{Benchmark \texttt{2026-safenlp-2024}} \label{tab:cat_2026_safenlp_2024}
{\setlength{\tabcolsep}{2pt}
\begin{tabular}[h]{@{}llllllrrr@{}}
\toprule
\textbf{\# ~} & \textbf{Tool} & \textbf{Verified} & \textbf{Falsified} & \textbf{Fastest} & \textbf{Penalty} & \textbf{Points} & \textbf{Score} & \textbf{Solved}\\
\midrule
1 & Vibecheck & 418 & 662 & 0 & 0 & 10800 & 100.0 & 100.0\% \\
2 & $\alpha$-$\beta$-CROWN & 418 & 662 & 0 & 0 & 10800 & 100.0 & 100.0\% \\
3 & PyRAT & 387 & 660 & 0 & 0 & 10470 & 96.9 & 96.9\% \\
4 & Bright & 278 & 659 & 0 & 0 & 9370 & 86.8 & 86.8\% \\
5 & NNV & 254 & 645 & 0 & 0 & 8990 & 83.2 & 83.2\% \\
6 & CORA & 258 & 639 & 0 & 0 & 8970 & 83.1 & 83.1\% \\
7 & SuperSimple & 161 & 662 & 0 & 0 & 8230 & 76.2 & 76.2\% \\
8 & N2v & 161 & 644 & 0 & 0 & 8050 & 74.5 & 74.5\% \\
9 & Act & 139 & 585 & 0 & 0 & 7240 & 67.0 & 67.0\% \\
10 & SobolBox & 0 & 203 & 0 & 0 & 2030 & 18.8 & 18.8\% \\
11 & nnenum & 262 & 579 & 0 & 54 & 310 & 2.9 & 77.9\% \\
12 & NeuralSAT & 390 & 573 & 0 & 86 & -3270 & 0 & 89.2\% \\
13 & Agate & 162 & 0 & 0 & 661 & -97530 & 0 & 15.0\% \\
\bottomrule
\end{tabular}
}
\end{center}
\end{table}

\begin{figure}[h]
\centerline{\includegraphics[width=\textwidth]{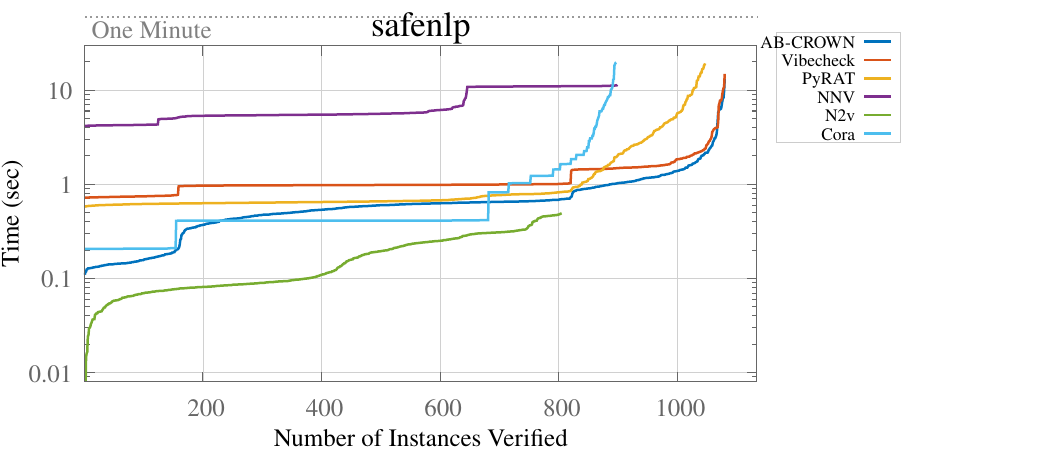}}
\caption{Cactus Plot for safenlp.}
\label{fig:quantPic}
\end{figure}

\clearpage

\begin{table}[h]
\begin{center}
\caption{Benchmark \texttt{2026-sat-relu}} \label{tab:cat_2026_sat_relu}
{\setlength{\tabcolsep}{2pt}
\begin{tabular}[h]{@{}llllllrrr@{}}
\toprule
\textbf{\# ~} & \textbf{Tool} & \textbf{Verified} & \textbf{Falsified} & \textbf{Fastest} & \textbf{Penalty} & \textbf{Points} & \textbf{Score} & \textbf{Solved}\\
\midrule
1 & Vibecheck & 50 & 50 & 0 & 0 & 1000 & 100.0 & 100.0\% \\
2 & NeuralSAT & 50 & 50 & 0 & 0 & 1000 & 100.0 & 100.0\% \\
3 & Bright & 50 & 50 & 0 & 0 & 1000 & 100.0 & 100.0\% \\
4 & $\alpha$-$\beta$-CROWN & 50 & 50 & 0 & 0 & 1000 & 100.0 & 100.0\% \\
5 & PyRAT & 39 & 50 & 0 & 0 & 890 & 89.0 & 89.0\% \\
6 & CORA & 50 & 49 & 0 & 1 & 840 & 84.0 & 99.0\% \\
7 & Act & 1 & 50 & 0 & 0 & 510 & 51.0 & 51.0\% \\
8 & NNV & 0 & 50 & 0 & 0 & 500 & 50.0 & 50.0\% \\
9 & N2v & 0 & 50 & 0 & 0 & 500 & 50.0 & 50.0\% \\
10 & SobolBox & 0 & 49 & 0 & 0 & 490 & 49.0 & 49.0\% \\
11 & nnenum & 3 & 28 & 0 & 0 & 310 & 31.0 & 31.0\% \\
12 & SuperSimple & 31 & 34 & 0 & 14 & -1450 & 0 & 65.0\% \\
13 & Agate & 1 & 0 & 0 & 42 & -6290 & 0 & 1.0\% \\
\bottomrule
\end{tabular}
}
\end{center}
\end{table}

\begin{figure}[h]
\centerline{\includegraphics[width=\textwidth]{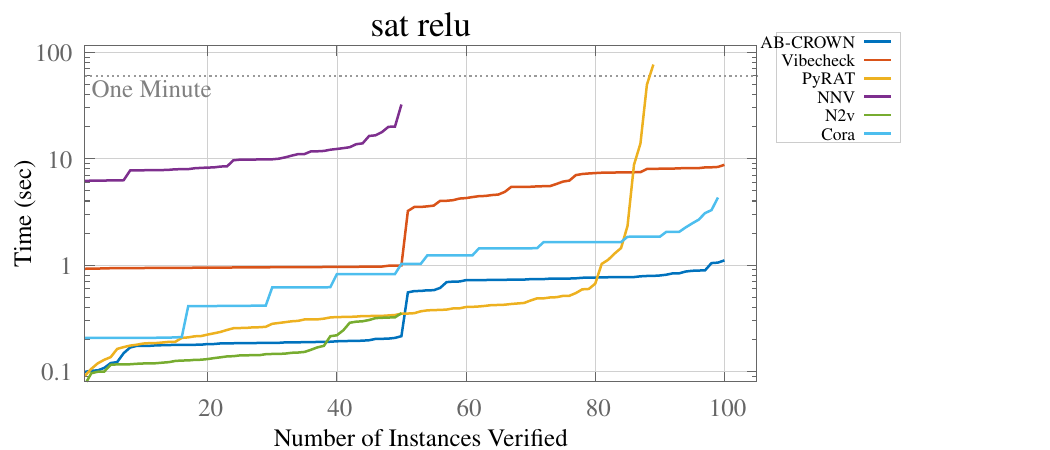}}
\caption{Cactus Plot for sat relu.}
\label{fig:quantPic}
\end{figure}

\clearpage

\begin{table}[h]
\begin{center}
\caption{Benchmark \texttt{2026-soundnessbench-2026}} \label{tab:cat_2026_soundnessbench_2026}
{\setlength{\tabcolsep}{2pt}
\begin{tabular}[h]{@{}llllllrrr@{}}
\toprule
\textbf{\# ~} & \textbf{Tool} & \textbf{Verified} & \textbf{Falsified} & \textbf{Fastest} & \textbf{Penalty} & \textbf{Points} & \textbf{Score} & \textbf{Solved}\\
\midrule
1 & Vibecheck & 14 & 39 & 0 & 0 & 530 & 100.0 & 88.3\% \\
2 & $\alpha$-$\beta$-CROWN & 14 & 45 & 0 & 1 & 440 & 83.0 & 98.3\% \\
3 & PyRAT & 14 & 7 & 0 & 0 & 210 & 39.6 & 35.0\% \\
4 & N2v & 0 & 14 & 0 & 0 & 140 & 26.4 & 23.3\% \\
5 & CORA & 14 & 0 & 0 & 0 & 140 & 26.4 & 23.3\% \\
6 & SuperSimple & 11 & 0 & 0 & 0 & 110 & 20.8 & 18.3\% \\
7 & NeuralSAT & 14 & 13 & 0 & 8 & -930 & 0 & 45.0\% \\
\bottomrule
\end{tabular}
}
\end{center}
\end{table}

\begin{figure}[h]
\centerline{\includegraphics[width=\textwidth]{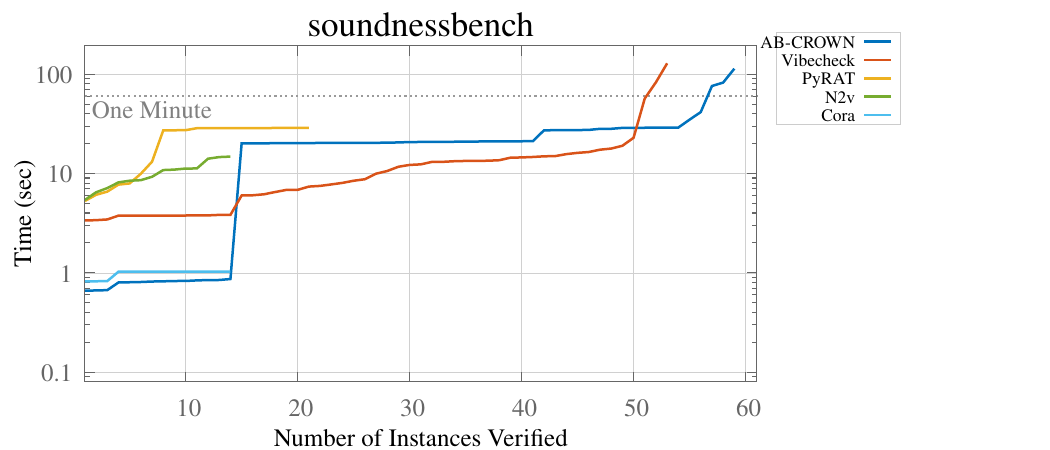}}
\caption{Cactus Plot for soundnessbench.}
\label{fig:quantPic}
\end{figure}

\clearpage

\begin{table}[h]
\begin{center}
\caption{Benchmark \texttt{2026-tinyimagenet-2024}} \label{tab:cat_2026_tinyimagenet_2024}
{\setlength{\tabcolsep}{2pt}
\begin{tabular}[h]{@{}llllllrrr@{}}
\toprule
\textbf{\# ~} & \textbf{Tool} & \textbf{Verified} & \textbf{Falsified} & \textbf{Fastest} & \textbf{Penalty} & \textbf{Points} & \textbf{Score} & \textbf{Solved}\\
\midrule
1 & Vibecheck & 139 & 39 & 0 & 0 & 1780 & 100.0 & 89.0\% \\
2 & $\alpha$-$\beta$-CROWN & 139 & 38 & 0 & 0 & 1770 & 99.4 & 88.5\% \\
3 & Bright & 104 & 36 & 0 & 0 & 1400 & 78.7 & 70.0\% \\
4 & NeuralSAT & 100 & 36 & 0 & 0 & 1360 & 76.4 & 68.0\% \\
5 & PyRAT & 75 & 35 & 0 & 0 & 1100 & 61.8 & 55.0\% \\
6 & NNV & 80 & 10 & 0 & 0 & 900 & 50.6 & 45.0\% \\
7 & SuperSimple & 0 & 35 & 0 & 0 & 350 & 19.7 & 17.5\% \\
8 & N2v & 0 & 35 & 0 & 0 & 350 & 19.7 & 17.5\% \\
\bottomrule
\end{tabular}
}
\end{center}
\end{table}

\begin{figure}[h]
\centerline{\includegraphics[width=\textwidth]{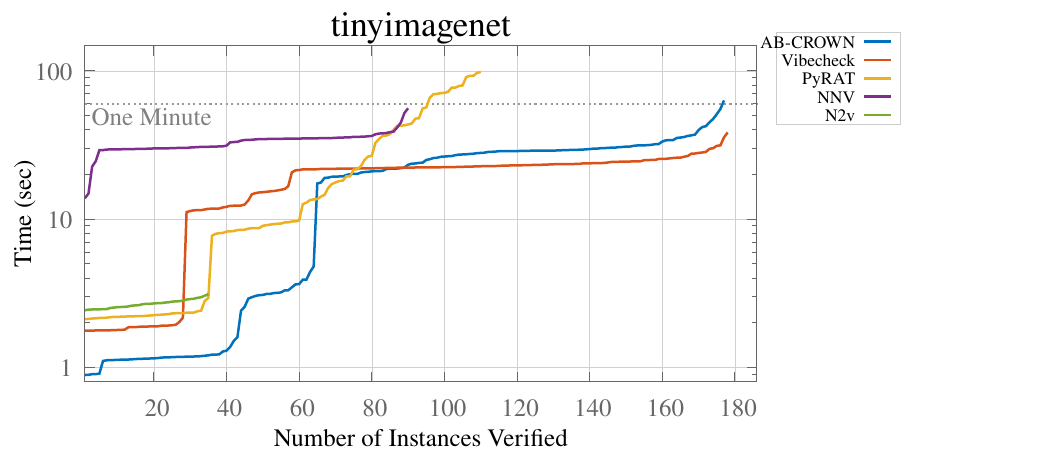}}
\caption{Cactus Plot for tinyimagenet.}
\label{fig:quantPic}
\end{figure}

\clearpage

\begin{table}[h]
\begin{center}
\caption{Benchmark \texttt{2026-tllverifybench-2023}} \label{tab:cat_2026_tllverifybench_2023}
{\setlength{\tabcolsep}{2pt}
\begin{tabular}[h]{@{}llllllrrr@{}}
\toprule
\textbf{\# ~} & \textbf{Tool} & \textbf{Verified} & \textbf{Falsified} & \textbf{Fastest} & \textbf{Penalty} & \textbf{Points} & \textbf{Score} & \textbf{Solved}\\
\midrule
1 & Vibecheck & 20 & 12 & 0 & 0 & 320 & 100.0 & 100.0\% \\
2 & SobolBox & 20 & 12 & 0 & 0 & 320 & 100.0 & 100.0\% \\
3 & PyRAT & 20 & 12 & 0 & 0 & 320 & 100.0 & 100.0\% \\
4 & NeuralSAT & 20 & 12 & 0 & 0 & 320 & 100.0 & 100.0\% \\
5 & $\alpha$-$\beta$-CROWN & 20 & 12 & 0 & 0 & 320 & 100.0 & 100.0\% \\
6 & CORA & 19 & 12 & 0 & 0 & 310 & 96.9 & 96.9\% \\
7 & Bright & 19 & 12 & 0 & 0 & 310 & 96.9 & 96.9\% \\
8 & NNV & 8 & 12 & 0 & 0 & 200 & 62.5 & 62.5\% \\
9 & Act & 13 & 7 & 0 & 0 & 200 & 62.5 & 62.5\% \\
10 & nnenum & 3 & 11 & 0 & 0 & 140 & 43.8 & 43.8\% \\
11 & N2v & 0 & 12 & 0 & 0 & 120 & 37.5 & 37.5\% \\
12 & Agate & 6 & 0 & 0 & 12 & -1740 & 0 & 18.8\% \\
\bottomrule
\end{tabular}
}
\end{center}
\end{table}

\begin{figure}[h]
\centerline{\includegraphics[width=\textwidth]{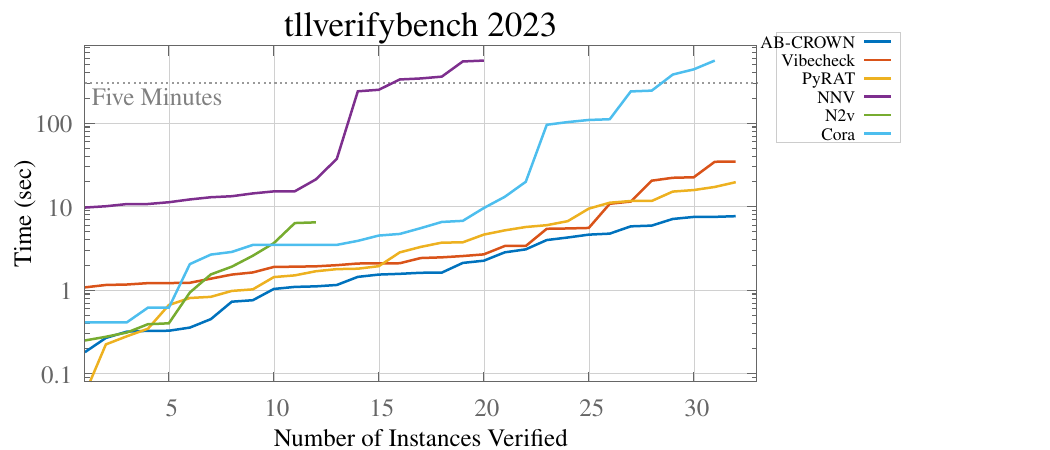}}
\caption{Cactus Plot for tllverifybench 2023.}
\label{fig:quantPic}
\end{figure}

\clearpage

\begin{table}[h]
\begin{center}
\caption{Benchmark \texttt{2026-traffic-signs-recognition-2023}} \label{tab:cat_2026_traffic_signs_recognition_2023}
{\setlength{\tabcolsep}{2pt}
\begin{tabular}[h]{@{}llllllrrr@{}}
\toprule
\textbf{\# ~} & \textbf{Tool} & \textbf{Verified} & \textbf{Falsified} & \textbf{Fastest} & \textbf{Penalty} & \textbf{Points} & \textbf{Score} & \textbf{Solved}\\
\midrule
1 & $\alpha$-$\beta$-CROWN & 0 & 36 & 0 & 0 & 360 & 100.0 & 80.0\% \\
2 & Vibecheck & 0 & 22 & 0 & 0 & 220 & 61.1 & 48.9\% \\
3 & PyRAT & 0 & 1 & 0 & 0 & 10 & 2.8 & 2.2\% \\
\bottomrule
\end{tabular}
}
\end{center}
\end{table}

\begin{figure}[h]
\centerline{\includegraphics[width=\textwidth]{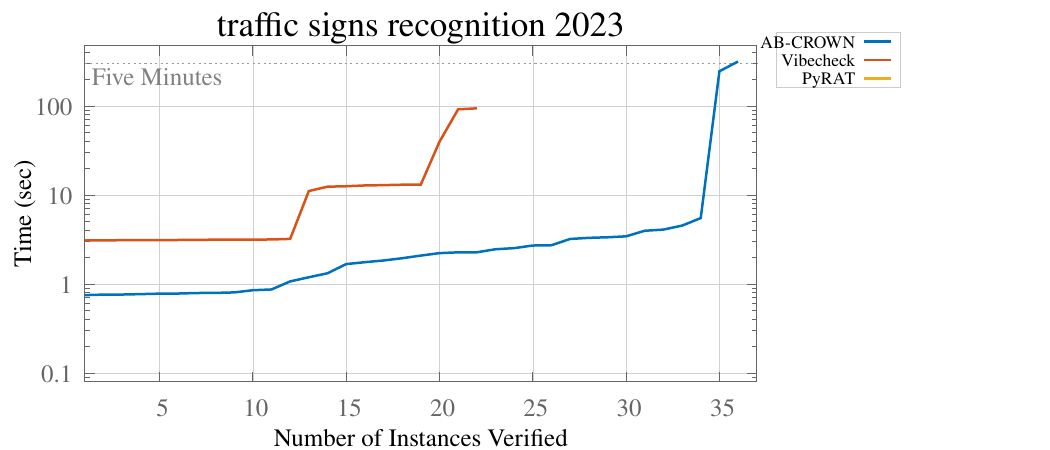}}
\caption{Cactus Plot for traffic signs recognition 2023.}
\label{fig:quantPic}
\end{figure}

\clearpage

\begin{table}[h]
\begin{center}
\caption{Benchmark \texttt{2026-vggnet16-2022}} \label{tab:cat_2026_vggnet16_2022}
{\setlength{\tabcolsep}{2pt}
\begin{tabular}[h]{@{}llllllrrr@{}}
\toprule
\textbf{\# ~} & \textbf{Tool} & \textbf{Verified} & \textbf{Falsified} & \textbf{Fastest} & \textbf{Penalty} & \textbf{Points} & \textbf{Score} & \textbf{Solved}\\
\midrule
1 & $\alpha$-$\beta$-CROWN & 18 & 0 & 0 & 0 & 180 & 100.0 & 100.0\% \\
2 & Vibecheck & 16 & 0 & 0 & 0 & 160 & 88.9 & 88.9\% \\
3 & PyRAT & 15 & 0 & 0 & 0 & 150 & 83.3 & 83.3\% \\
4 & NeuralSAT & 15 & 0 & 0 & 0 & 150 & 83.3 & 83.3\% \\
\bottomrule
\end{tabular}
}
\end{center}
\end{table}

\begin{figure}[h]
\centerline{\includegraphics[width=\textwidth]{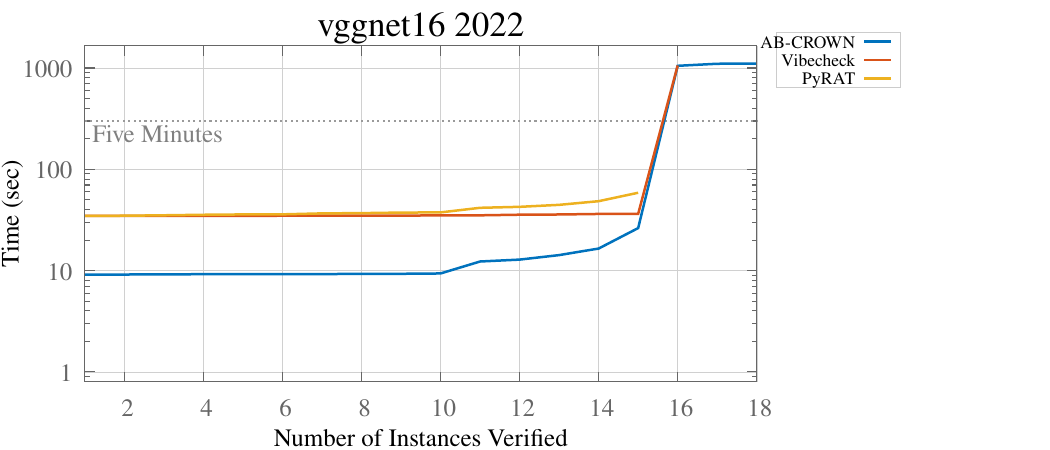}}
\caption{Cactus Plot for vggnet16 2022.}
\label{fig:quantPic}
\end{figure}

\clearpage

\begin{table}[h]
\begin{center}
\caption{Benchmark \texttt{2026-vit-2023}} \label{tab:cat_2026_vit_2023}
{\setlength{\tabcolsep}{2pt}
\begin{tabular}[h]{@{}llllllrrr@{}}
\toprule
\textbf{\# ~} & \textbf{Tool} & \textbf{Verified} & \textbf{Falsified} & \textbf{Fastest} & \textbf{Penalty} & \textbf{Points} & \textbf{Score} & \textbf{Solved}\\
\midrule
1 & Act & 200 & 0 & 0 & 0 & 2000 & 100.0 & 100.0\% \\
2 & $\alpha$-$\beta$-CROWN & 123 & 0 & 0 & 0 & 1230 & 61.5 & 61.5\% \\
3 & Vibecheck & 111 & 0 & 0 & 0 & 1110 & 55.5 & 55.5\% \\
4 & PyRAT & 70 & 0 & 0 & 0 & 700 & 35.0 & 35.0\% \\
5 & NeuralSAT & 63 & 0 & 0 & 0 & 630 & 31.5 & 31.5\% \\
6 & N2v & 20 & 0 & 0 & 0 & 200 & 10.0 & 10.0\% \\
\bottomrule
\end{tabular}
}
\end{center}
\end{table}

\begin{figure}[h]
\centerline{\includegraphics[width=\textwidth]{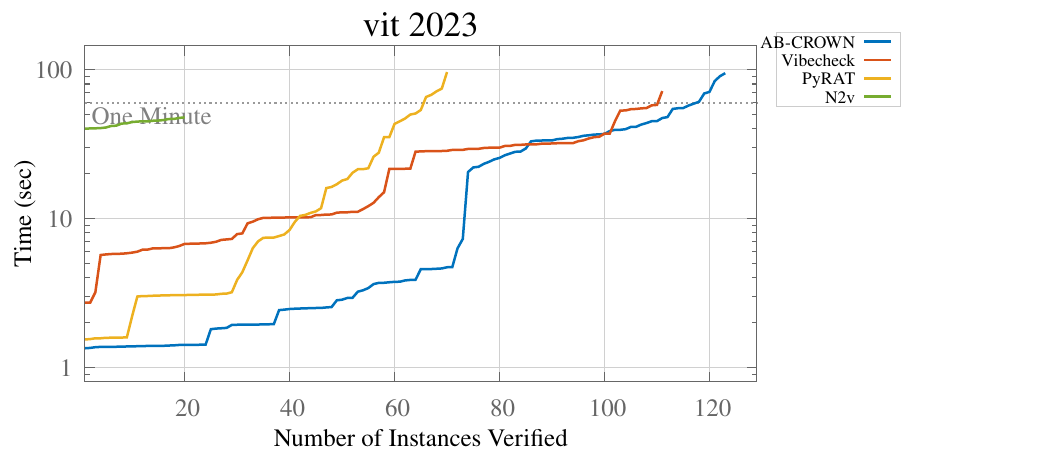}}
\caption{Cactus Plot for vit 2023.}
\label{fig:quantPic}
\end{figure}

\clearpage

\begin{table}[h]
\begin{center}
\caption{Benchmark \texttt{2026-yolo-2023}} \label{tab:cat_2026_yolo_2023}
{\setlength{\tabcolsep}{2pt}
\begin{tabular}[h]{@{}llllllrrr@{}}
\toprule
\textbf{\# ~} & \textbf{Tool} & \textbf{Verified} & \textbf{Falsified} & \textbf{Fastest} & \textbf{Penalty} & \textbf{Points} & \textbf{Score} & \textbf{Solved}\\
\midrule
1 & $\alpha$-$\beta$-CROWN & 66 & 0 & 0 & 0 & 660 & 100.0 & 91.7\% \\
2 & Vibecheck & 64 & 0 & 0 & 0 & 640 & 97.0 & 88.9\% \\
3 & NeuralSAT & 64 & 0 & 0 & 0 & 640 & 97.0 & 88.9\% \\
4 & PyRAT & 50 & 0 & 0 & 0 & 500 & 75.8 & 69.4\% \\
5 & SuperSimple & 8 & 0 & 0 & 0 & 80 & 12.1 & 11.1\% \\
\bottomrule
\end{tabular}
}
\end{center}
\end{table}

\begin{figure}[h]
\centerline{\includegraphics[width=\textwidth]{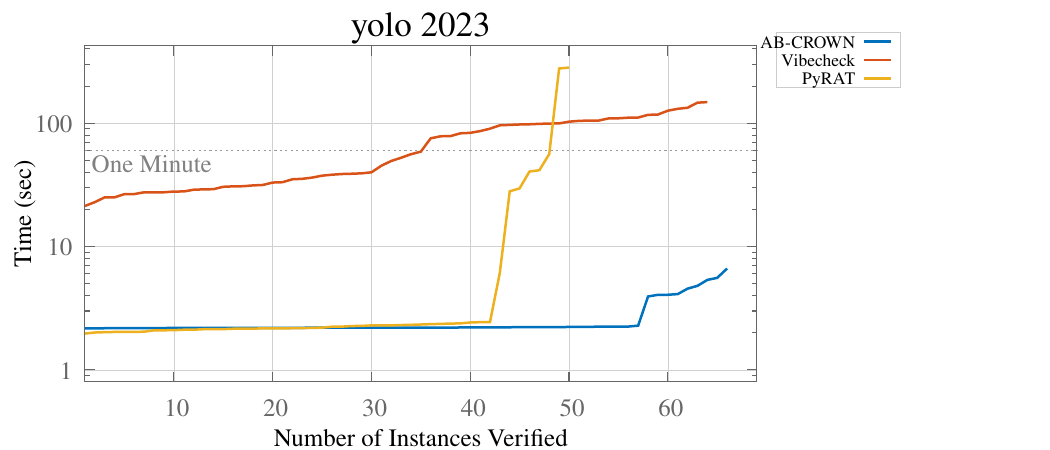}}
\caption{Cactus Plot for yolo 2023.}
\label{fig:quantPic}
\end{figure}

\clearpage
\subsection{Extended Track}
\label{sec:benchmark_results_extended}

\clearpage

\begin{table}[h]
\begin{center}
\caption{Benchmark \texttt{2026-adaptive-cruise-control-non-linear-2026}} \label{tab:cat_2026_adaptive_cruise_control_non_linear_2026}
{\setlength{\tabcolsep}{2pt}
\begin{tabular}[h]{@{}llllllrrr@{}}
\toprule
\textbf{\# ~} & \textbf{Tool} & \textbf{Verified} & \textbf{Falsified} & \textbf{Fastest} & \textbf{Penalty} & \textbf{Points} & \textbf{Score} & \textbf{Solved}\\
\midrule
1 & $\alpha$-$\beta$-CROWN & 34 & 16 & 0 & 0 & 500 & 100.0 & 100.0\% \\
2 & Vibecheck & 32 & 17 & 0 & 0 & 490 & 98.0 & 98.0\% \\
3 & PyRAT & 32 & 17 & 0 & 0 & 490 & 98.0 & 98.0\% \\
4 & N2v & 23 & 16 & 0 & 0 & 390 & 78.0 & 78.0\% \\
5 & NNV & 18 & 1 & 0 & 0 & 190 & 38.0 & 38.0\% \\
\bottomrule
\end{tabular}
}
\end{center}
\end{table}

\begin{figure}[h]
\centerline{\includegraphics[width=\textwidth]{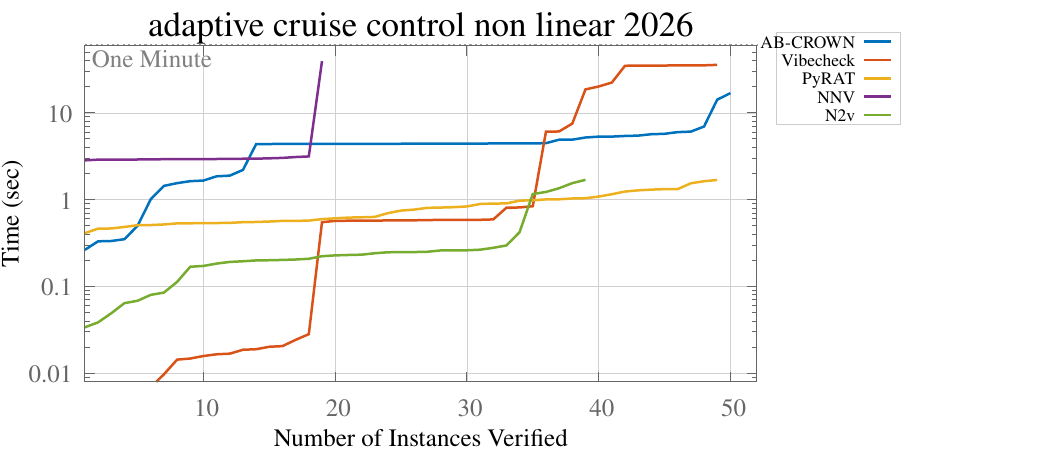}}
\caption{Cactus Plot for adaptive cruise control non linear 2026.}
\label{fig:quantPic}
\end{figure}

\clearpage

\begin{table}[h]
\begin{center}
\caption{Benchmark \texttt{2026-cctsdb-yolo-2023}} \label{tab:cat_2026_cctsdb_yolo_2023}
{\setlength{\tabcolsep}{2pt}
\begin{tabular}[h]{@{}llllllrrr@{}}
\toprule
\textbf{\# ~} & \textbf{Tool} & \textbf{Verified} & \textbf{Falsified} & \textbf{Fastest} & \textbf{Penalty} & \textbf{Points} & \textbf{Score} & \textbf{Solved}\\
\midrule
1 & Vibecheck & 11 & 28 & 0 & 0 & 390 & 100.0 & 100.0\% \\
2 & PyRAT & 11 & 28 & 0 & 0 & 390 & 100.0 & 100.0\% \\
3 & NNV & 11 & 28 & 0 & 0 & 390 & 100.0 & 100.0\% \\
4 & $\alpha$-$\beta$-CROWN & 11 & 28 & 0 & 0 & 390 & 100.0 & 100.0\% \\
\bottomrule
\end{tabular}
}
\end{center}
\end{table}

\begin{figure}[h]
\centerline{\includegraphics[width=\textwidth]{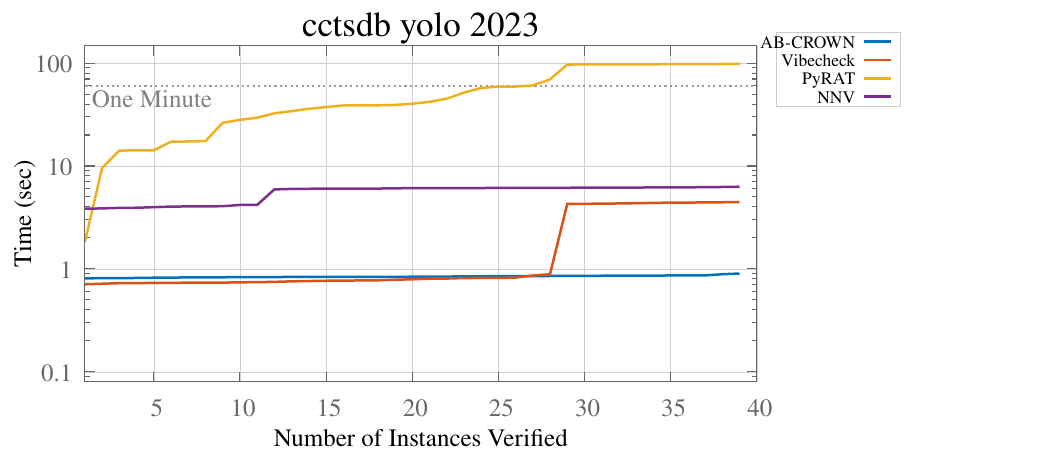}}
\caption{Cactus Plot for cctsdb yolo 2023.}
\label{fig:quantPic}
\end{figure}

\clearpage

\begin{table}[h]
\begin{center}
\caption{Benchmark \texttt{2026-collins-aerospace-benchmark}} \label{tab:cat_2026_collins_aerospace_benchmark}
{\setlength{\tabcolsep}{2pt}
\begin{tabular}[h]{@{}llllllrrr@{}}
\toprule
\textbf{\# ~} & \textbf{Tool} & \textbf{Verified} & \textbf{Falsified} & \textbf{Fastest} & \textbf{Penalty} & \textbf{Points} & \textbf{Score} & \textbf{Solved}\\
\midrule
1 & Vibecheck & 0 & 6 & 0 & 0 & 60 & 100.0 & 100.0\% \\
2 & PyRAT & 0 & 6 & 0 & 0 & 60 & 100.0 & 100.0\% \\
3 & N2v & 0 & 6 & 0 & 0 & 60 & 100.0 & 100.0\% \\
4 & $\alpha$-$\beta$-CROWN & 0 & 6 & 0 & 0 & 60 & 100.0 & 100.0\% \\
5 & NNV & 0 & 2 & 0 & 0 & 20 & 33.3 & 33.3\% \\
\bottomrule
\end{tabular}
}
\end{center}
\end{table}

\begin{figure}[h]
\centerline{\includegraphics[width=\textwidth]{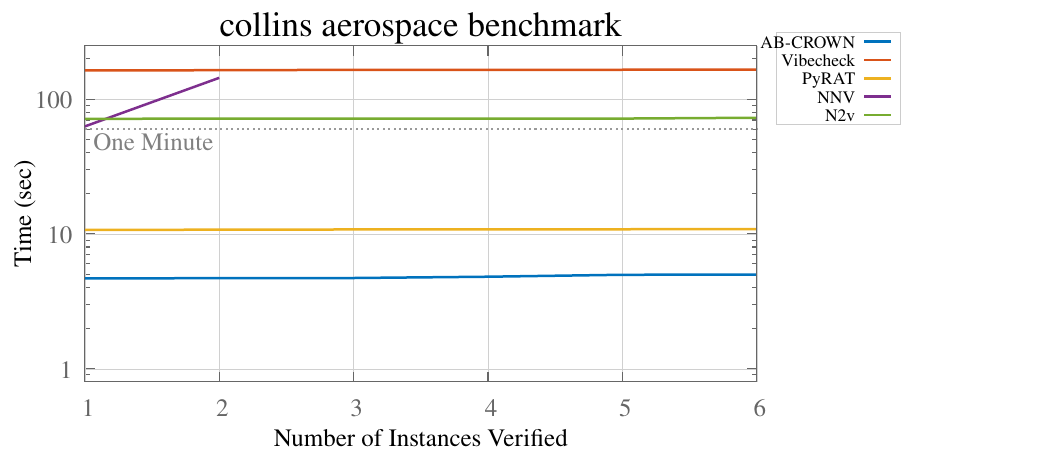}}
\caption{Cactus Plot for collins aerospace benchmark.}
\label{fig:quantPic}
\end{figure}

\clearpage

\begin{table}[h]
\begin{center}
\caption{Benchmark \texttt{2026-isomorphic-acasxu-2026}} \label{tab:cat_2026_isomorphic_acasxu_2026}
{\setlength{\tabcolsep}{2pt}
\begin{tabular}[h]{@{}llllllrrr@{}}
\toprule
\textbf{\# ~} & \textbf{Tool} & \textbf{Verified} & \textbf{Falsified} & \textbf{Fastest} & \textbf{Penalty} & \textbf{Points} & \textbf{Score} & \textbf{Solved}\\
\midrule
1 & $\alpha$-$\beta$-CROWN & 46 & 4 & 0 & 0 & 500 & 100.0 & 100.0\% \\
2 & PyRAT & 40 & 0 & 0 & 0 & 400 & 80.0 & 80.0\% \\
3 & Vibecheck & 34 & 4 & 0 & 0 & 380 & 76.0 & 76.0\% \\
4 & CORA & 28 & 4 & 0 & 0 & 320 & 64.0 & 64.0\% \\
\bottomrule
\end{tabular}
}
\end{center}
\end{table}

\begin{figure}[h]
\centerline{\includegraphics[width=\textwidth]{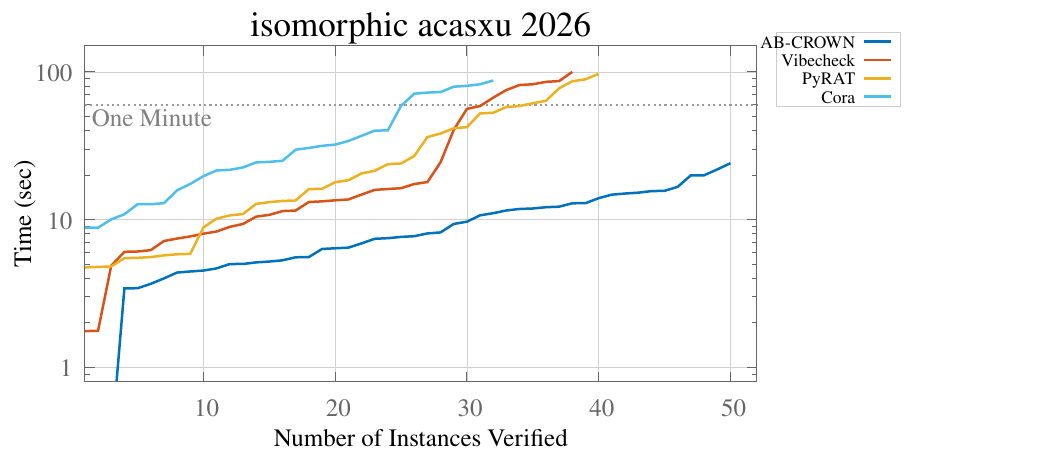}}
\caption{Cactus Plot for isomorphic acasxu 2026.}
\label{fig:quantPic}
\end{figure}

\clearpage

\begin{table}[h]
\begin{center}
\caption{Benchmark \texttt{2026-monotonic-acasxu-2026}} \label{tab:cat_2026_monotonic_acasxu_2026}
{\setlength{\tabcolsep}{2pt}
\begin{tabular}[h]{@{}llllllrrr@{}}
\toprule
\textbf{\# ~} & \textbf{Tool} & \textbf{Verified} & \textbf{Falsified} & \textbf{Fastest} & \textbf{Penalty} & \textbf{Points} & \textbf{Score} & \textbf{Solved}\\
\midrule
1 & Vibecheck & 0 & 50 & 0 & 0 & 500 & 100.0 & 100.0\% \\
2 & PyRAT & 0 & 50 & 0 & 0 & 500 & 100.0 & 100.0\% \\
3 & NNV & 0 & 50 & 0 & 0 & 500 & 100.0 & 100.0\% \\
4 & $\alpha$-$\beta$-CROWN & 0 & 50 & 0 & 0 & 500 & 100.0 & 100.0\% \\
5 & CORA & 0 & 47 & 0 & 0 & 470 & 94.0 & 94.0\% \\
\bottomrule
\end{tabular}
}
\end{center}
\end{table}

\begin{figure}[h]
\centerline{\includegraphics[width=\textwidth]{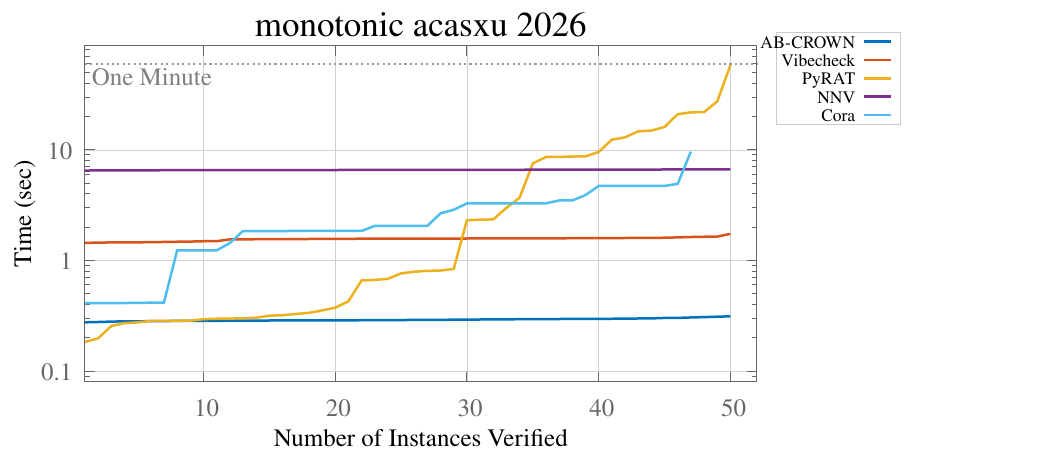}}
\caption{Cactus Plot for monotonic acasxu 2026.}
\label{fig:quantPic}
\end{figure}

\clearpage

\begin{table}[h]
\begin{center}
\caption{Benchmark \texttt{2026-smart-turn-multimodal-2026}} \label{tab:cat_2026_smart_turn_multimodal_2026}
{\setlength{\tabcolsep}{2pt}
\begin{tabular}[h]{@{}llllllrrr@{}}
\toprule
\textbf{\# ~} & \textbf{Tool} & \textbf{Verified} & \textbf{Falsified} & \textbf{Fastest} & \textbf{Penalty} & \textbf{Points} & \textbf{Score} & \textbf{Solved}\\
\midrule
1 & Vibecheck & 0 & 50 & 0 & 0 & 500 & 100.0 & 100.0\% \\
2 & $\alpha$-$\beta$-CROWN & 0 & 50 & 0 & 0 & 500 & 100.0 & 100.0\% \\
\bottomrule
\end{tabular}
}
\end{center}
\end{table}

\begin{figure}[h]
\centerline{\includegraphics[width=\textwidth]{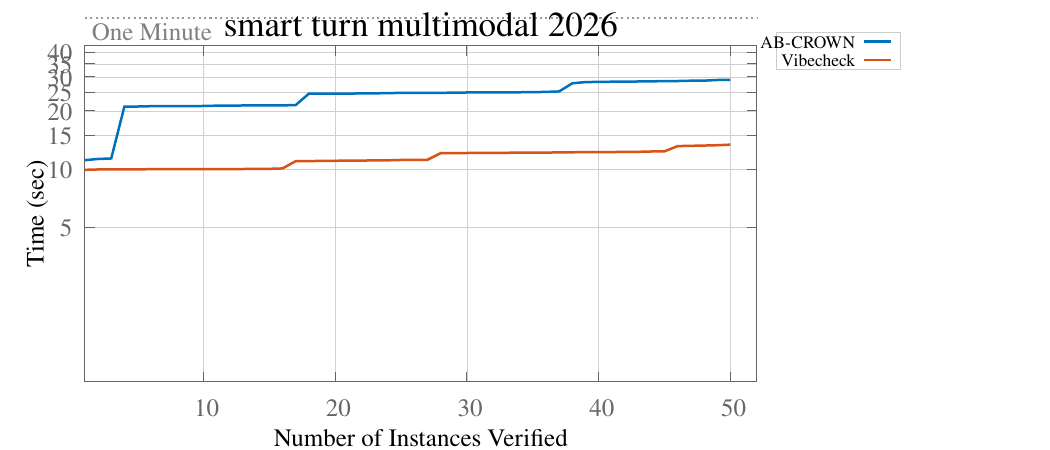}}
\caption{Cactus Plot for smart turn multimodal 2026.}
\label{fig:quantPic}
\end{figure}

\clearpage
\begin{landscape}
\section{Instance Results}
\label{sec:results_detailed}

\begin{center}
{\setlength{\tabcolsep}{1pt}
\scriptsize
\begin{longtable}{@{}llllllllllllllll@{}}
\caption{\footnotesize Instance Runtimes. Fastest times are \textcolor{blue}{blue}. Second fastest are \textcolor{second}{green}. Penalties are red crosses (\textbf{\textcolor{red}{\ding{55}}}).} \label{tab:all_results} \\
\toprule
\textbf{Category} & \textbf{Id} & \textbf{Result} & \textbf{$\alpha$-$\beta$-C} & \textbf{vibecheck} & \textbf{PyRAT} & \textbf{NSAT} & \textbf{NNV} & \textbf{CORA} & \textbf{bright} & \textbf{n2v} & \textbf{SSimp} & \textbf{act} & \textbf{NNen} & \textbf{SB} & \textbf{agate} \\
\midrule
\endhead
2026 Acasxu 2023 & 0 & ~\textsc{unsat} & \textcolor{darkgray}{2.37} & \textcolor{darkgray}{3.62} & \textcolor{darkgray}{2.06} & \textcolor{darkgray}{11.3} & \textcolor{darkgray}{98.3} & \textcolor{darkgray}{7.18} & \textcolor{darkgray}{13.9} & - & - & \textcolor{darkgray}{1.36} & \textcolor{darkgray}{0.99} & \textcolor{darkgray}{69.1} & - \\
2026 Acasxu 2023 & 1 & ~\textsc{unsat} & \textcolor{darkgray}{1.68} & \textcolor{darkgray}{3.57} & \textcolor{darkgray}{1.30} & \textcolor{darkgray}{11.3} & \textcolor{darkgray}{19.3} & \textcolor{darkgray}{3.49} & \textcolor{darkgray}{12.2} & - & - & \textcolor{darkgray}{1.37} & \textcolor{darkgray}{1.35} & \textcolor{darkgray}{66.8} & - \\
2026 Acasxu 2023 & 2 & ~\textsc{unsat} & \textcolor{darkgray}{1.83} & \textcolor{darkgray}{3.92} & \textcolor{darkgray}{1.58} & \textcolor{darkgray}{11.5} & \textcolor{darkgray}{19.7} & \textcolor{darkgray}{3.89} & \textcolor{darkgray}{13.1} & - & - & \textcolor{darkgray}{2.22} & \textcolor{darkgray}{2.24} & \textcolor{darkgray}{65.7} & - \\
2026 Acasxu 2023 & 3 & ~\textsc{unsat} & \textcolor{darkgray}{1.82} & \textcolor{darkgray}{3.80} & \textcolor{darkgray}{1.32} & \textcolor{darkgray}{11.4} & \textcolor{darkgray}{19.4} & \textcolor{darkgray}{3.48} & \textcolor{darkgray}{12.3} & - & - & \textcolor{darkgray}{1.95} & \textcolor{darkgray}{2.42} & \textcolor{darkgray}{68.0} & - \\
2026 Acasxu 2023 & 4 & ~\textsc{unsat} & \textcolor{darkgray}{1.68} & \textcolor{darkgray}{3.32} & \textcolor{darkgray}{1.34} & \textcolor{darkgray}{11.3} & \textcolor{darkgray}{19.3} & \textcolor{darkgray}{3.28} & \textcolor{darkgray}{12.1} & - & - & \textcolor{darkgray}{1.28} & \textcolor{darkgray}{1.36} & \textcolor{darkgray}{66.4} & - \\
\bottomrule
\end{longtable}
}
\end{center}
\end{landscape}


\end{document}